%% file: PrivacyProber_main.tex
\documentclass[10pt,journal,compsoc]{IEEEtran}
\usepackage{amsmath,amssymb,amsfonts}
\usepackage{algorithmic}
\usepackage{graphicx}
\usepackage{textcomp}
\usepackage{tikz}
\usepackage{pifont}
\newcommand{\cmark}{\ding{51}}

\usepackage{caption}
\usepackage{makecell}
\usepackage{booktabs}
\usepackage{multirow}
\usepackage{subcaption}
\usepackage{caption}
\usepackage{multirow}
\usepackage{multicol}
\usepackage{xcolor}
\usepackage{graphicx}
\usepackage{rotating}
\usepackage{enumitem}
\usepackage{colortbl}
\usepackage{arydshln}
\usepackage{lipsum}
\usepackage[ruled,vlined]{algorithm2e}

\definecolor{mygray}{gray}{0.6}
\newcommand{\gmark}{\color{mygray} \ding{55}}
\usepackage{url}
\usepackage[pagebackref=false,breaklinks=true,colorlinks,bookmarks=false]{hyperref}

\ifCLASSOPTIONcompsoc
  \usepackage[nocompress]{cite}
\else
  \usepackage{cite}
\fi
\ifCLASSINFOpdf
\else
\fi
\newcommand{\frameworkname}{PrivacyProber}

\begin{document}
%
\title{\frameworkname: Assessment and Detection of Soft--Biometric Privacy--Enhancing Techniques}
%
%
%
%

\author{Peter~Rot,~\IEEEmembership{Member,~IEEE,} Klemen~Grm,~\IEEEmembership{Member,~IEEE,}
        Peter~Peer,~\IEEEmembership{Senior~Member,~IEEE,}
        and~Vitomir~\v{S}truc,~\IEEEmembership{Senior~Member,~IEEE}
\IEEEcompsocitemizethanks{\IEEEcompsocthanksitem P. Rot is with the Faculty of Electrical Engineering as well as with the Faculty of Computer and information Science, University of Ljubljana, Ljubljana, Slovenia, Corresponding author e-mail: peter.rot@fe.uni-lj.si\protect\\
\IEEEcompsocthanksitem P. Peer is with the Faculty of Computer and information Science, University of Ljubljana, Ve\v{c}na Pot 113, 1000 Ljubljana, Slovenia.\protect\\
\IEEEcompsocthanksitem K. Grm and V. \v{S}truc are with the Faculty of Electrical Engineering, University of Ljubljana, Tr\v{z}a\v{s}ka cesta 25, 1000 Ljubljana, Slovenia.\protect\\
}
\thanks{
}}

\markboth{Journal of \LaTeX\ Class Files}
{Shell \MakeLowercase{\textit{et al.}}: Bare Demo of IEEEtran.cls for Computer Society Journals}
%



\IEEEtitleabstractindextext{%
\begin{abstract}
Soft--biometric privacy--enhancing techniques represent machine learning methods that aim to: (i) mitigate privacy concerns associated with face recognition technology by suppressing selected soft--biometric attributes in facial images (e.g., gender, age, ethnicity) and (ii) make  unsolicited extraction of 
sensitive personal information infeasible. Because such techniques are increasingly used in real--world applications, it is imperative to understand to what extent the privacy enhancement can be inverted and how much attribute information can be recovered from privacy--enhanced images. While these aspects are critical, they have not been investigated in the literature so far. 
In this paper, we, therefore, \textit{study the robustness} of several state--of--the--art soft--biometric privacy--enhancing techniques to attribute recovery attempts. We propose PrivacyProber, a high--level framework for restoring soft--biometric information from privacy--enhanced facial images, and apply it for attribute recovery 
in comprehensive experiments on three public face datasets, i.e., LFW, MUCT and Adience. Our experiments show that the proposed framework is able to restore a considerable amount of suppressed information, regardless of the  privacy--enhancing technique used, but also that there are significant differences between the considered privacy models. These results point to the need for novel mechanisms that can improve the robustness of existing privacy--enhancing techniques and secure them against potential adversaries trying to restore suppressed information. 
Additionally, we demostrate that PrivacyProber can also be used to detect privacy--enhancement in facial images (under black--box assumptions) with high accuracy. Specifically, we show that a detection procedure can be developed around the proposed framework that is \textit{learning free} and, therefore, generalizes well across different data characteristics and privacy--enhancing techniques.
\end{abstract}

}

\maketitle

\IEEEdisplaynontitleabstractindextext

%
\IEEEpeerreviewmaketitle

\IEEEraisesectionheading{\section{Introduction}\label{sec:introduction}}

\IEEEPARstart{F}{acial} images represent a rich source of information, from which a multitude of attributes can be extracted automatically using contemporary machine learning models, including gender, age, ethnicity, affective state, or even body mass indices~\cite{dantcheva2015else, ranjan2017hyperface,pan2018mean,niu2016ordinal,zavan2019benchmarking,berthouze2020emopain,li2019face,mao2020deep}. 
However, with the rapid proliferation of automatic recognition techniques in facial analytics, privacy--related concerns have also emerged and now represent a {\em key challenge with regard to the trustworthiness} of the technology ~\cite{jobin2019global,singh2020trustworthy}. 

To address these concerns, researchers are increasingly looking into \textit{privacy--enhancing} mechanisms capable of ensuring a trade--off between the utility of the data for facial analytics, on the one hand, and privacy protection, on the other~\cite{mirjalili2019flowsan,chhabra2018anonymizing, meden2021Survey, mirjalili2020privacynet, PEMIU_Access2020, BortolatoFG2020}. \textcolor{black}{Examples of such mechanisms include: (i) \textit{deidentification techniques} that try to conceal the identity of the subjects in the data, while retaining other useful attributes and information~\cite{shawn2020fawkes,meden2017face,meden2021Survey}, but also} (ii) so-called \textit{soft--biometric privacy--enhancing techniques}, which aim to perturb (transform, modify) facial images in a way that makes it difficult (or impossible) for automatic machine learning techniques to infer sensitive attributes (e.g., gender, age, ethnicity), while making only minimal changes to the visual appearance of the images. \textcolor{black}{While a considerable amount of research has been done on deidentification technology, soft--biometric privacy--enhancing techniques represents a relatively new research topic that has so far been underexplored.}  

\textcolor{black}{The importance of soft--biometric privacy--enhancing techniques is highlighted through various  application scenarios where concealing soft-biometrics attributes, such as demographics, is critical. For example, when sharing personal images online (e.g., on social media), soft--biometric privacy--enhancement can help to avoid automatic user profiling, attribute extraction and other unsolicited forms of processing. Additionally, in face verification systems, where only the identity information is needed, such techniques can conceal information on other attributes that are not required for verification purposes.} Such privacy--enhancing mechanisms represent pivotal tools for addressing societal expectations about the appropriate use of personal data, but also for meeting safeguards and standards defined in data--protection legislation and privacy acts, such as, GDPR~\cite{Voigt:2017:EGD:3152676}, CCPA~\cite{10.2307/j.ctvjghvnn}, BIPA~\cite{bipa} and others~\cite{meden2021Survey}. \textcolor{black}{As noted in recent surveys on biometric privacy \cite{meden2021Survey,melzi2022overview}, other types of techniques exist that deal with different aspects of privacy (e.g., encryption, model training, synthetic data, etc.). However, the focus of this paper is exclusively on soft--biometric privacy--enhancement.}

Several powerful solutions have been proposed in the literature for ensuring soft--biometric privacy and preventing automatic extraction of facial attributes, including synthesis--based techniques~\cite{othman2014privacy}, auto--encoder based models~\cite{mirjalili2018semi,mirjalili2019flowsan} or adversarial perturbations~\cite{rozsa2019facial,chhabra2018anonymizing}. While these solutions have been shown to successfully obscure 
soft--biometric information for selected attribute classifiers, experimental performance evaluations 
are typically limited to vanilla (or zero--effort) evaluation scenarios, where no attempt is made to reconstruct the obscured attributes. 
Due to this practice, it is not clear if the  privacy levels reported in the literature generalize to real--world applications, where a potential adversary may exploit additional knowledge and invest considerable effort and resources to recover the concealed information. 
Such zero--effort evaluations, hence, \textit{raise questions about the reliability} of existing privacy models\footnote{The term privacy model is used as a synonym for biometric privacy--enhancing technique in this paper for brevity.} and their sensitivity to attribute recovery attempts. Comprehensive reliability studies are, therefore, critical for better understanding the capabilities of contemporary privacy--enhancing techniques and have implications for their deployment in practice. However, to the best of our knowledge, such studies are largely missing from the literature.

\begin{figure}[!t]
	\includegraphics[width=\columnwidth, trim = 3mm 2mm 0 0, clip]{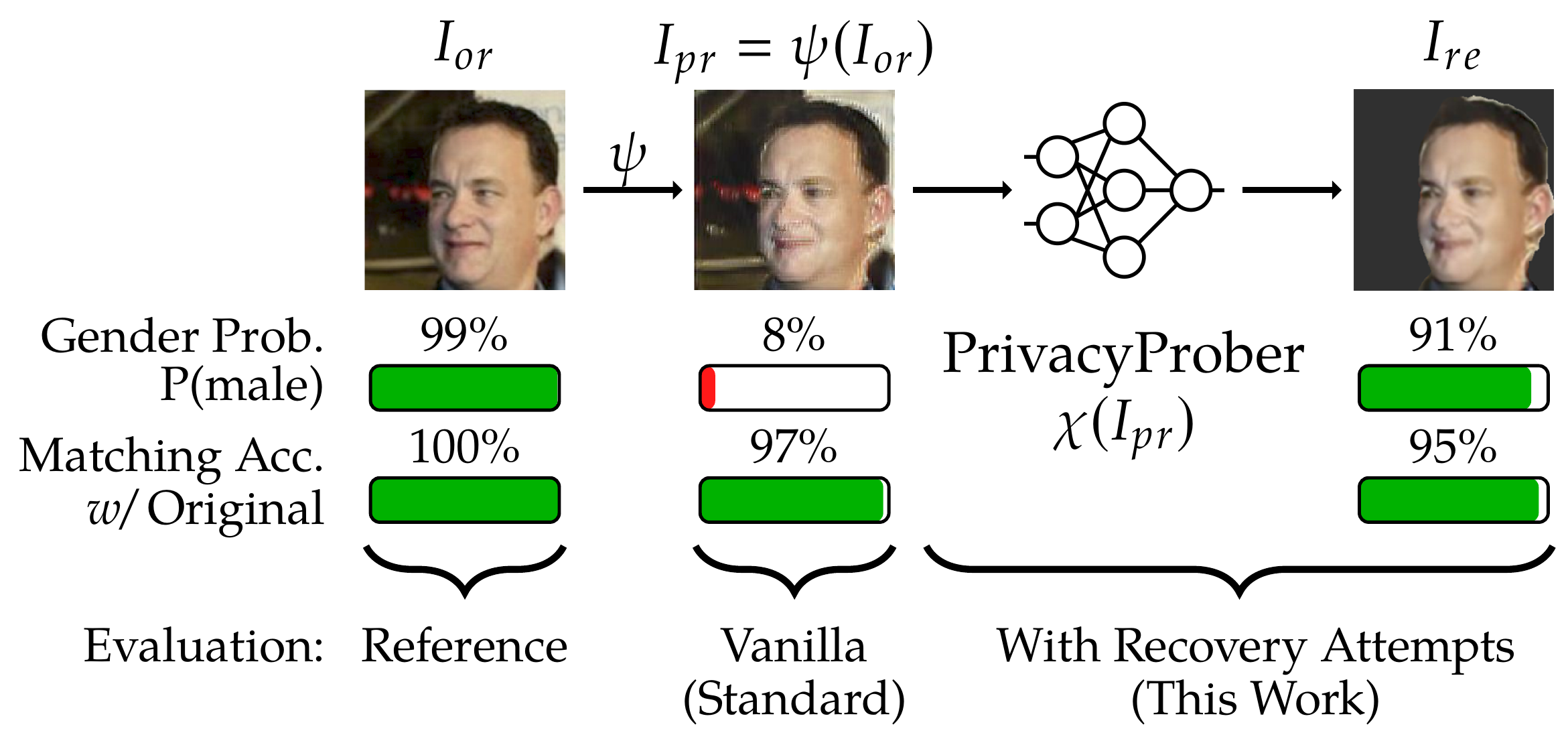}
	\caption[width=\columnwidth]
	{The performance of soft--biometric privacy models, $\psi$, is commonly evaluated by comparing attribute classification performance over the original (reference) face images, $I_{or}$, and their privacy--enhanced counterparts, $I_{pr}$. In this paper we study the robustness of existing privacy models beyond such vanilla evaluation scenarios and propose an attribute recovery framework, called PrivacyProber, that allows for more comprehensive evaluations using images with reconstructed soft--biometric information, $I_{re}$. The bottom part of the figure visualizes the idea through illustrative gender predictions (i.e., probabilities of being male) and the match scores between $I_{or}$ and the transformed images, $I_{pr}$ and $I_{re}$. Note the difference in gender predictions for the different tasks with and without attribute recovery.
	}
	\label{fig:overview}
\end{figure}

In this paper, we address this gap and explore possibilities for recovering obscured (concealed, suppressed)  information from privacy--enhanced facial images with the goal of assessing the reliability and robustness of existing soft--biometric privacy models. 
To facilitate the study, we develop an attribute recovery framework, called {\em \frameworkname}, which uses various types of image transformations (learned over clean, non--tampered images) to reconstruct suppressed information from privacy--enhanced facial images. 
The proposed framework is based on minimal (black box) assumptions and requires no examples of privacy--enhanced images to restore soft--biometric information, which makes it applicable to a wide variety (of conceptually different) \textcolor{black}{soft-biometric} privacy models. To demonstrate the feasibility of our framework, we conduct extensive experiments with multiple state--of--the--art privacy models and over three publicly available face datasets. The results of our experiments suggest that PrivacyProber is able to recover a considerable amount of suppressed attribute information and that sensitivity to reconstruction attacks is still a considerable issue with existing privacy--enhancing techniques. 
As an additional contribution, we also show that the proposed framework can be used to {\em detect privacy--enhancement} in facial images, pointing to another threat vector with respect to existing privacy models that allows for flagging tampered images and treating them differently from non--tampered data, e.g., using manual screening. 

As part of our reliability study (illustrated in Fig.~\ref{fig:overview}), we make the following main contributions: 
\begin{itemize}
    \item We conduct, to the best of our knowledge, the {\em first comprehensive investigation} into the robustness of soft--biometric privacy--enhancing techniques with respect to attribute recovery attempts. We show that despite recent progress in this area, a considerable amount of concealed information can still be recovered from (most) privacy--enhanced images, but also that there are significant differences in terms of robustness between the tested privacy models.  
    \item We propose {\em PrivacyProber}, a high-level framework for attribute recovery from privacy--enhanced facial images. The framework relies on dedicated reconstruction schemes build around inpainting, denoising, and face parsing models \textcolor{black}{as well as self-supervised adversarial defenses,} and requires no access to the evaluated privacy--enhancing techniques or prior knowledge about their inner workings.
    \item We present {\em novel methodology} to evaluate the robustness of soft--biometric privacy models. 
    Specifically, we introduce an \textit{attribute--recovery robustness (ARR)} score that reflects the robustness of privacy--enhancing techniques by comparing attribute--classification accuracy over reference and attribute--recovered images. We  demonstrate the value of ARR scores  through extensive experimentation with multiple privacy models and datasets. 
    \item \textcolor{black}{We show that~\frameworkname~can be used to detect privacy--enhancement in facial images and propose an original detection approach, called {\em APEND (Evidence \textbf{A}ggregation for \textbf{P}rivacy-\textbf{En}hancement \textbf{D}etection}), that consolidates differences between the probability predictions of an attribute classifier applied to facial images before and after a series of attribute recovery attempts. We show that APEND facilitates the detection of privacy--enhanced images, while ensuring highly encouraging performance when compared to related solutions from the literature. }
\end{itemize}

\section{Background and Related work}
\label{sec:RelatedWork}

In this section, we provide background information on soft--biometric privacy-enhancing techniques and survey the most relevant prior work. For an in-depth coverage of the broader topic of visual privacy and privacy protection of facial images, we refer the reader to some of the excellent and comprehensive recent surveys in this area, e.g.,~\cite{meden2021Survey,Winkler2014_networks_survey,PadillaLopez2015_survey,garfinkel2015identification,Ribaric2016_survey,melzi2022overview}. 

\subsection{Problem Definition and Model Taxonomy\label{SubSec:Problam_definition}}

\textcolor{black}{Assume an original bona-fide face image, $I_{or}\in\mathbb{R}^{w\times h}$, and an attribute classifier $\xi_{a}: \mathbb{R}^{w\times h} \mapsto \{a_1,a_2,\ldots,a_N\}$, with the attribute labels $\{a_i\}_{i=1}^N$ corresponding to  classes $\{C_1, C_2, \ldots, C_N$\}. Soft--biometric privacy--enhancement, $\psi$, aims to produce privacy--enhanced images, $I_{pr}=\psi({I_{or}})$, from which the class labels $a_i$ cannot be correctly predicted by $\xi_a$ with high confidence. In general, the goal of $\psi$ is to obscure attribute information from machine learning models, but also to ensure that the appearance of the perturbed image is as close to the bone-fide one as possible, so that the visual content appears similar for human observers, i.e., $\min||I_{pr}- I_{or}||_{L_p}$, where $||\cdot||_{L_p}$ is an $L_p$ norm}.  
These characteristics are illustrated on the left part of of Fig.~\ref{fig:overview}. 

\textcolor{black}{As suggested in~\cite{rot2022detecting}, existing soft--biometric privacy models can in general be  grouped into: (i) techniques that try to induce incorrect attribute predictions, and (ii) techniques that try to generate approximately equal class probabilities for all attributes. Solutions from the first group aim to enhance privacy by inducing misclassifications, i.e., $\xi_{a}(I_{pr})\neq\xi_{a}(I_{or})$, through various mechanism, e.g., adversarial perturbations and related strategies. With these techniques, an \textit{incorrect} attribute label is typically predicted from $I_{pr}$ with high probability.} Techniques from the second group, on the other hand, commonly rely on (input--conditioned) synthesis methods that enhance privacy by altering image characteristics is a way that makes attribute predictions unreliable, i.e., $p(C_1|I_{pr})\approx p(C_1|I_{pr})\approx\cdots\approx p(C_N|I_{pr})$, where $p(C_i|I_{pr})$ represents the posteriors of the attribute classes given the privacy--enhanced image $I_{pr}$ generated by $\xi_a$. \textcolor{black}{Recent techniques from both groups are discussed below}. 

\subsection{Soft--biometric Privacy Models}\label{SubSec: OverviewSoftPrivacy}

A significant amount of research has been presented in the literature recently that addresses different problems related  to soft--biometric privacy~\cite{mirjalili2020privacynet, mirjalili2019flowsan, rozsa2019facial,terhorst2019Unsupervised,BortolatoFG2020,terhorst2020evaluation,morales2020sensitivenets, terhorst2019suppressing}.  Mirjalili and Ross~\cite{mirjalili2017soft}, for example, were among the first to explore privacy--enhancing techniques that perturb gender information in facial images. The solution, developed by the authors, first performs Delaunay triangulation (over face feature points) to decompose faces into a set of triangles that can be manipulated with the goal of privacy protection. Next, the texture within the triangles is optimized, such that a selected gender classifier generates unreliable predictions, while the input image is perturbed ever so slightly. The authors showed that their solution results in image manipulations that obscure gender information well, while having only a minimal impact on facial appearance and, consequently, on verification performance.

Another notable technique for soft--biometric gender privacy was introduced by Mirjalili \textit{et al.} in~\cite{mirjalili2018semi}. This work described \textit{so-called} Semi--Adversarial Networks (SANs), i.e., machine learning models that rely on \textit{conditional image synthesis} to conceal gender information in facial images. Conceptually, SANs represent convolutional auto--encoders that are paired with two distinct discriminators that steer the synthesis process -- one for enforcing gender privacy and the second for retaining verification accuracy (i.e., appearance similarities). The proposed SAN models were demonstrated to be capable of efficiently suppressing gender information, while retaining the data utility for identity verification tasks. 
To improve the generalization capabilities of SANs to unseen attribute classification models, FlowSAN models~\cite{mirjalili2019flowsan} were introduced in the follow-up work of the same authors. The main idea behind the FlowSANs was to employ multiple SAN transformations one after the other, with the goal of making the privacy enhancement applicable with arbitrary gender classifiers. The FlowSAN models were demonstrated to generalize better than their predecessors, SANs, while providing a good trade--off between privacy enhancement and utility preservation.

In~\cite{mirjalili2020privacynet}, Marialli \textit{et al.} introduced PrivacyNet~\cite{mirjalili2020privacynet}, an extended SAN model build around Generative Adversarial Networks (GANs). Unlike competing solutions, PrivacyNet was shown to be capable of soft--biometric privacy enhancement with respect to different (non-binary, continuous) facial attributes, including race, age and gender. The model, hence, generalized the fundametal idea behind SAN models to arbitrary soft--biometric attributes (as well as their combinations). We note at this point that the SAN--based family of algorithms is not based on adversarial perturbations, but relies on \textit{facial synthesis} facilitated by auto--encoders driven by a number of competing discriminators. 

An adversarial approach for privacy enhancement of $k$ facial attributes ($k$-AAP) was described in~\cite{chhabra2018anonymizing}. The proposed approach tries to infuse facial images with adversarial noise with the goal of obscuring selected soft-biometric attributes, while preserving others. 
$k$--AAP relies on the established Carlini Wagner $L_2$ attack~\cite{carlini2017towards} and was demonstrated to achieve competitive  results with attribute classifiers considered during the construction of the adversarial noise.  Similarly as related adversarial methods, $k$--AAP results in image manipulation that effectively conceal attribute information for machine learning models, while being barely detectable by human observers. Much like the original SAN models, $k$--AAP performs best with a known attribute classifiers, but struggles somewhat with unseen classification models. A conceptually similar idea involving adversarial noise was also investigated in~\cite{rozsa2019facial} where the Fast Gradient Sign Method (FGSM) method~\cite{goodfellow2014explaining} was utilized to explore the robustness of facial feature perturbations.

\subsection{Evaluation of Soft--biometric Privacy Models}


 Quantifying the level of privacy enhancement is a challenging task and requires well--defined evaluation methodologies and corresponding performance scores that provide insight into the characteristics of the tested privacy models. To quantify performance, the majority of prior work in this area exploits automatic recognition techniques trained for extracting various facial attributes. Recognition experiments are then performed on the original and privacy--enhanced images, and differences in the observed classification accuracies are used for performance reporting~\cite{morales2020sensitivenets, terhorst2019suppressing, mirjalili2019flowsan,mirjalili2020privacynet,BortolatoFG2020}. 
 
 Additionally, several scalar performance scores have also been proposed in the literature. Othman and Ross~\cite{othman2014privacy}, for example, introduced \textit{gender suppression levels}, a performance score for measuring the success of a synthesis--based privacy--enhancing technique with respect to the induced misclassification of gender. Dhar \textit{et al.}~\cite{dhar2019attributes} studied how state--of--the--art recognition models process soft--biometric information and explored how much sensitive information is encoded in different layers of a deep face recognition model. They proposed \textit{expressivity} as a measure  of  how  much information a given representation carries about a selected face attribute. Terhörst \textit{et al.} proposed the \textit{privacy-gain identity--loss} coefficient (PIC), which measures the gain of privacy with respect to a chosen facial attribute (e.g. gender) but takes the retained verification utility into account as well. The same authors also proposed ~\cite{terhorst2019suppressing} \textit{correct overall/female/male classification rates} (COCR/CFCR/CMCR) to score the success of privacy--enhancing algorithms. More recently, the authors of~\cite{terhorst2020evaluation} proposed a set of evaluation protocols with associated performance measures to enable 
 reproducible research on soft--biometric privacy.

While the evaluation methodologies reviewed above provide initial estimates about the performance of biometric privacy--enhancing techniques, they only assume zero--effort evaluation scenarios. In this study, we built on the presented work and \textit{propose a more comprehensive evaluation methodology} that also considers attribute reconstruction attempts when scoring privacy models. We introduce a novel performance measure that captures the robustness of the models and offers insight into the difficulty of recovering concealed attribute information.   


\subsection{Detection of Privacy Enhancement}

Soft--biometric privacy-enhancing techniques introduce changes to the visual characteristics of facial images and can, hence, be seen as a form of image tampering. While a significant amount of work has been described in the literature to detect such tampering, e.g.~\cite{zheng2019survey,da2020critical,meena2019image}, such detection techniques have mainly been studied only within the information-forensics community. The problem of detecting privacy enhancement, on the other hand, is new and, \textcolor{black}{except for the PREM work in \cite{rot2022detecting}, where a detection model based on face super-resolution and prediction divergence was presented, has not been studied widely in the open literature. Different from \cite{rot2022detecting}, we demonstrate in this paper that the process of attribute recovery, facilitated by the proposed PrivacyProber, can also be exploited to develop efficient privacy-enhancement detectors and that the information aggregated through different instantiations of PrivacyProber lead to highly robust performance.}  

\textcolor{black}{Finally, we note that because some privacy models are based on adversarial perturbations, the problem of detecting soft-biometric privacy enhancements is also partially related to adversarial attack detection methods~\cite{nowroozi2020survey, bulusu2020anomalous, wang2019security,chakraborty2018adversarial,agarwal2020image}. However, because soft--biometric privacy model also include} synthesis--based methods (among others)  -- as discussed in Section~\ref{SubSec: OverviewSoftPrivacy} -- the problem of detecting such image modifications is considerably broader.

\section{\frameworkname}\label{Sec: PrivacyProber} 
In this section, we describe the proposed \frameworkname, a framework for the  
recovery of suppressed soft--biometric attribute information from privacy enhanced facial images. \textcolor{black}{We discuss multiple contributions, including novel schemes for recovering attribute information under minimal assumptions as well as a novel approach for the detection of privacy enhancement using PrivacyProber.}


\subsection{Overview of \frameworkname}
\begin{figure}[!t]
	\includegraphics[width=\columnwidth
	]{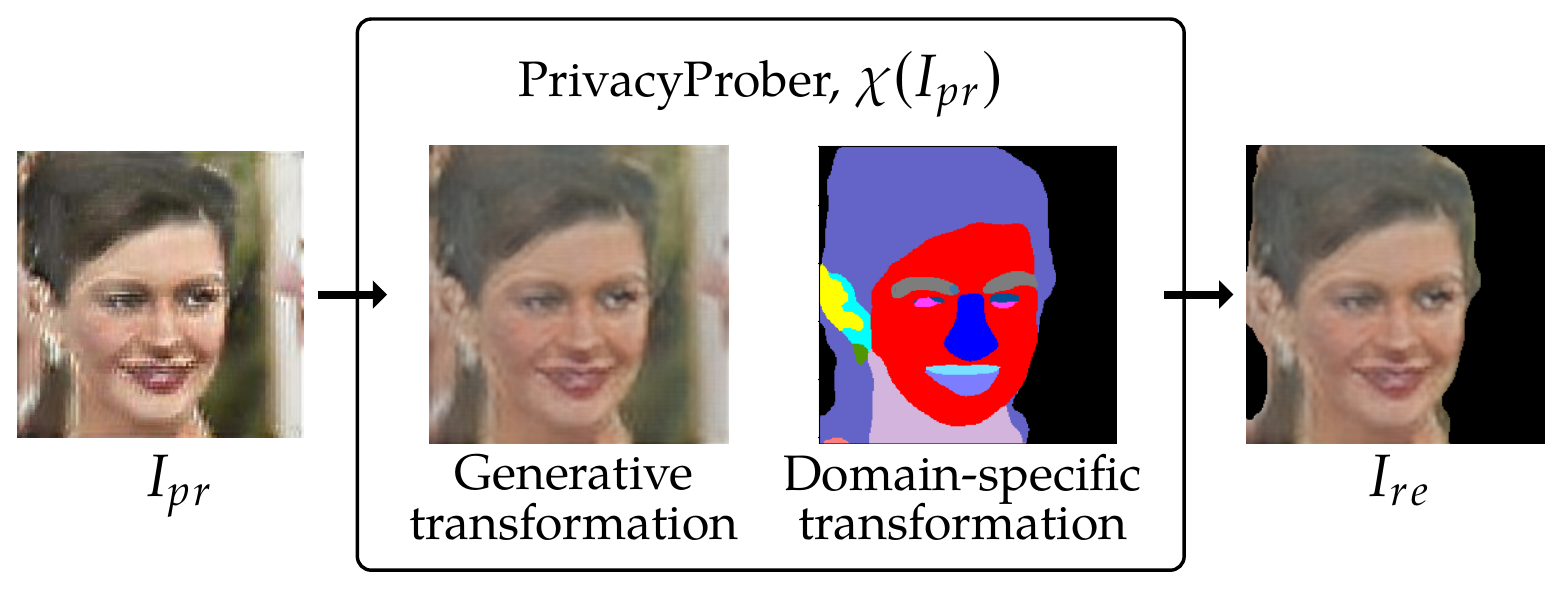}
	\caption{High--level illustration of PrivacyProber. Generative and domain--specific transformations are used (separately or in sequence) to recover attribute information concealed/suppressed by soft--biometric privacy models. \label{fig:PP_placeholder}}
\end{figure}

Existing evaluation schemes for soft--biometric privacy models typically compare the performance of an attribute\footnote{Common attributes considered in the literature include {\em gender, age, or ethnicity.}} classifier $\xi_a(\cdot)$, when applied to the original $\xi_a(I_{or})$ and privacy enhanced images $\xi_a(I_{pr})$. While the observed performance difference provides a first estimate of the level of privacy ensured by the privacy models, it also assumes that no attempt is made to recover the information initially contained in the input image $I_{or}$. As a result, the privacy levels empirically determined with such vanilla evaluation methodology may be overestimated and not representative of the actual capabilities of existing privacy models.

The proposed PrivacyProber, described below, tries to address this problem by facilitating performance assessments beyond zero--effort evaluation experiments. Specifically, PrivacyProber seeks to recover concealed image information by transforming privacy--enhanced images $I_{pr}$ in such a way that the privacy enhancement is reversed, or formally, such that the classifier $\xi_a(\cdot)$ generates the same predictions for the original, $I_{or}$,  and recovered images, $I_{re}$, i.e.:
\begin{equation}
    \xi_a(I_{or}) = \xi_a(\chi(I_{pr})),
\end{equation}
where $I_{re}=\chi(I_{pr})$ and $\chi(\cdot)$ is the transformation applied by the PrivacyProber. By estimating the level of privacy through performance differences of the attribute classifier $\xi(\cdot)$ applied to the original $\xi(I_{or})$ and attribute--recovered images $\xi_a(I_{re})$, where $I_{re}= \chi(I_{pr})$, a more informative estimate of privacy--enhancement performance (or robustness) can be obtained. The main idea is presented in Fig.~\ref{fig:overview}. 

While in general, the transformation $\chi(\cdot)$ could be learnt in a data--driven manner by defining a loss penalizing the difference in classification outputs between $\xi(I_{or})$  and $\xi(\chi(I_{pr}))$, such an approach is (i) model-specific and not universally applicable, (ii) requires access to training data for each considered privacy model, and (iii) warrants prior knowledge about the internal mechanism governing the privacy--enhancement procedure. 
In this paper we, therefore, follow a more general approach and design PrivacyProber under the following (minimal) assumptions:
\begin{itemize}
    \item \textbf{Black--box privacy models:} We assume no information about (or access to) the privacy models is available. We only exploit  the fact that the privacy models aim to suppress soft--biometrics information, while making minimal changes to the appearance of the  images.
    \item \textbf{Target domain:} We assume that privacy--enhancement is applied within a fixed image domain, i.e., on facial images, 
    and not on images of arbitrary scenes. 
\end{itemize}

Thus, we construct PrivacyProber based on a set of \textit{(i) generative} and \textit{(ii) domain-specific} transformations, which can be used separately or in sequence, as illustrated in Fig.~\ref{fig:PP_placeholder}. In the next sections we propose several possibilities for recovering suppressed soft--biometric attributes from privacy--enhanced images using such transformations.



\subsection{Generative transformations}

Soft--biometric privacy--enhancing techniques modify or partially corrupt the visual content in the original images $I_{or}$ with the goal of reducing the utility of the data for some targeted attribute classification task. The use of generative transformations\footnote{Note that we use the term \textit{generative transformation} loosely here to account for the transformation considered in this group, which include inpaining, \textcolor{black}{auto--encoder}, super--resolution and denoising.} for mitigating such image tampering is motivated by the fact that prior information about (clean, non--tampered) original face images can be incorporated efficiently into generative models (without the need for examples of  privacy--enhanced images) and exploited for attribute recovery. Similar approaches have also proven useful for protection against adversarial attacks~\cite{mustafa2019image,Gupta_2019_ICCV,mustafa2019image}. We propose three generative transformations for implementing PrivacyProber \textcolor{black}{using dedicated schemes based on {\em (i) inpainting}, {\em(ii) denoising}, and {\em(iii) image reconstruction}}.   
\subsubsection{Attribute Recovery Through Inpainting}\label{SubSubSec: Inpainting}

\begin{figure}[t]
\centering
    \begin{minipage}[b]{0.20\columnwidth}
        \includegraphics[width=0.99\columnwidth]{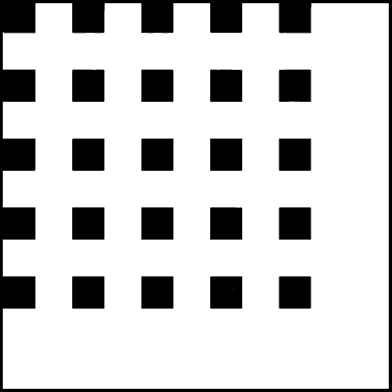}
    \end{minipage}
    \hfill
    \begin{minipage}[b]{0.20\columnwidth}
        \includegraphics[width=0.99\columnwidth]{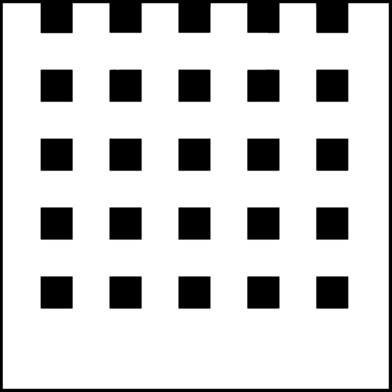}
    \end{minipage}
    \hfill
    \begin{minipage}[b]{0.20\columnwidth}
        \includegraphics[width=0.99\columnwidth]{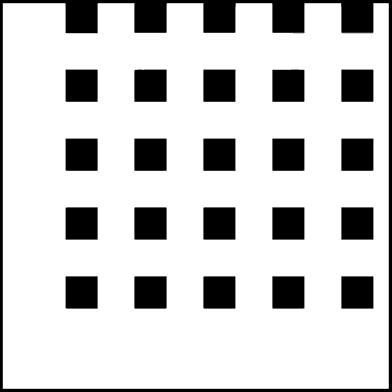}
    \end{minipage}
    \hfill
    \begin{minipage}[t][][t]{0.08\columnwidth}
    \centering
    \vspace{-10mm}
       $\cdots$
    \end{minipage}
    \hfill
    \begin{minipage}[b]{0.20\columnwidth}
        \includegraphics[width=0.99\columnwidth]{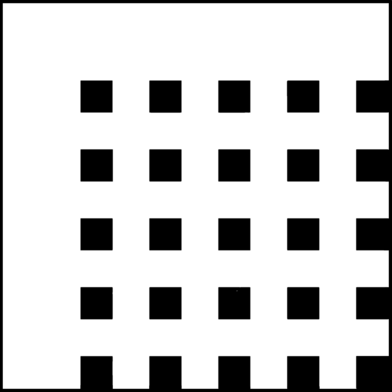}
    \end{minipage}
    \hfill\\ 
    
    \begin{minipage}[b]{0.20\columnwidth}
    	\centering
        {\footnotesize $B_1$}
    \end{minipage}
    \hfill
    \begin{minipage}[b]{0.20\columnwidth}
    	\centering
        {\footnotesize $B_2$}
    \end{minipage}
    \hfill
    \begin{minipage}[b]{0.20\columnwidth}
    	\centering
        {\footnotesize $B_3$}
    \end{minipage}
    \hfill
    \begin{minipage}[b]{0.08\columnwidth}
        \ \
    \end{minipage}
    \hfill
    \begin{minipage}[b]{0.20\columnwidth}
    	\centering
        {\footnotesize $B_N$}
    \end{minipage}\\\vspace{-0.5mm}
    \color{gray}\noindent\makebox[\linewidth]{\rule{3cm}{0.05pt}}\vspace{2mm}
    
	\centering
	\includegraphics[width=0.99\columnwidth, trim=40 50 90 0, clip]{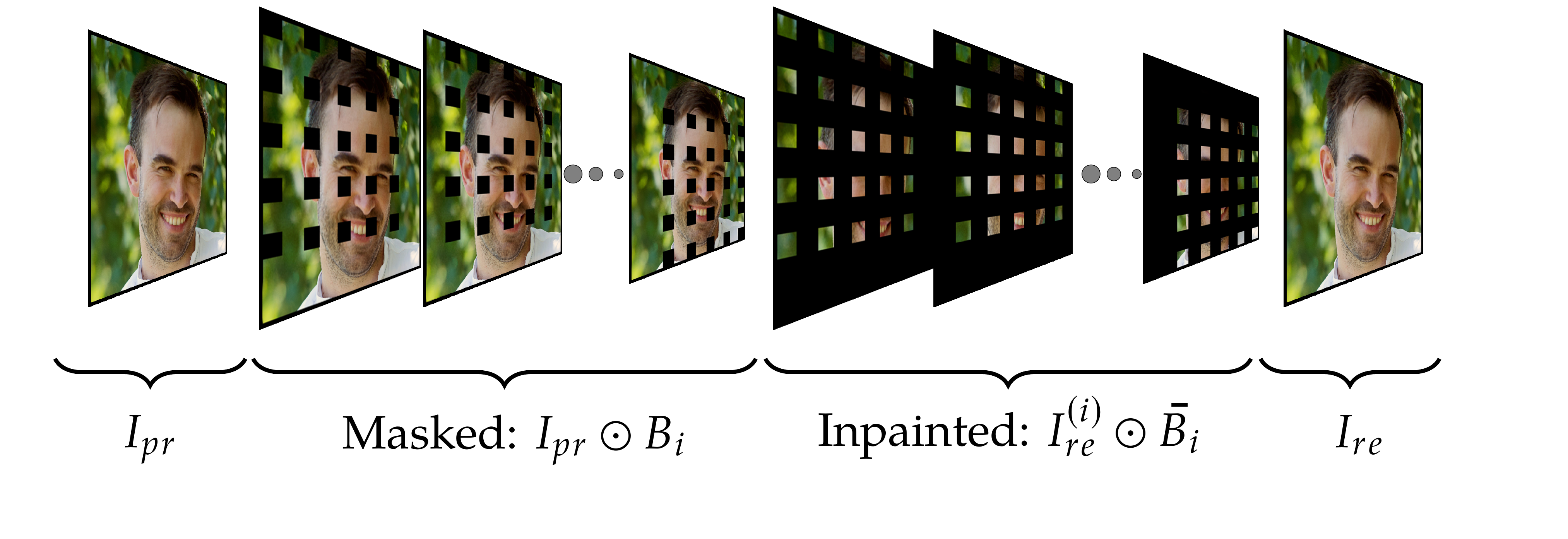}
	\caption{Illustration of the inpainting--based attribute recovery procedure, $\chi_{in}$, proposed for  PrivacyProber. The privacy--enhanced image, $I_{pr}$, is masked $N$ times with the binary masks $B_i$ -- shown in the top row. The masked regions are then inpainted based on the remaining context. Finally, the attribute--recovered image $I_{re}$ is reconstructed from the inpainted regions only.}
	\label{fig:inp}
\end{figure}

Inpainting models typically aim to fill in missing pixels in a damaged image and to restore the corrupted content. Our goal, on the other hand, is to recover a complete image $I_{re}$ from $I_{pr}$ with attribute information restored and not just a small portion of the data. We, therefore, design a \textcolor{black}{novel attribute-recovery scheme based on} inpainting for this task that sequentially inpaints small parts of the image at the time and then aggregates the results to recover the complete image. The basic assumption with this attribute--recovery strategy is that inpainting can restore (clean) non--tampered content by inferring pixels from contextual information, even if this information was tampered with by a soft--biometric privacy--enhancing model. 


The proposed recovery procedure, illustrated in Fig.~\ref{fig:inp}, starts with the privacy--enhanced image $I_{pr}\in\mathbb{R}^{w\times h}$ and a set of $N$ binary masks $B_i\in\mathbb{R}^{w\times h}$, where $w$ and $h$ again denote the width and height of the image, respectively, and $i\in\{1, 2, \ldots, N\}$.  The binary masks are initialized as matrices of all ones. Next, a chess--like pattern composed of multiple square regions (of size $d \times d$, where $d\ll\{w,h\}$) is  constructed and placed into the initialized masks. Pixels in $B_i$ corresponding to the chess--like pattern are set to zero. As illustrated in the top row of Fig~\ref{fig:inp}, the pattern is positioned in the top right corner for the first mask and then shifted $N$--times in a sliding--window manner (with overlap), such that all positions are traversed in both the horizontal and vertical direction. To facilitate inpainting, the constructed binary masks $B_i$ are then used to remove pixels from the privacy--enhanced image, i.e.:
\begin{equation}
  I_{pr}^{(i)} = I_{pr}\odot B_i, \  \text{for} \  i\in\{1,2,\ldots,N\},
\end{equation}
where $\odot$ represents the Hadamard product and $I_{pr}^{(i)}$ denotes the input image masked by $B_i$. In the next step, all pixels set to zero are reconstructed using a predefined inpainting model, which given a set of masked images $\{ I_{pr}^{(i)} \}_{i=1}^N$ generates a set of corresponding (partially) recovered images $\{I_{re}^{(i)}\}_{i=1}^N$. Finally, the complete attribute--recovered image $I_{re}$ is reconstructed from the inpainted regions only:
\begin{equation}
   I_{re} = \lfloor I_{re}^{(i)}\odot \bar{B_i}  \rfloor_{i=1}^N, 
\end{equation}
where $\bar{B_i}$ is an inverted version of $B_i$ used to exclude the original (non-inpainted) areas from $I_{re}^{(i)}$ and $\lfloor \cdot \rfloor$ represents an averaging operation over non--zero pixels. This type of averaging is needed due to the overlap in the masked regions between binary masks. We denote the presented attribute--recovery procedure as $\chi_{in}: I_{pr} \mapsto I_{re}$ hereafter and implement it using an off--the--shelf inpainting model.

\subsubsection{Attribute Recovery Through Denoising}
The second proposed approach to attribute recovery is based on denoising. Soft--biometric privacy--enhancing techniques typically aim to make only minute changes to the input images, $I_{or}$, and alter their visual characteristics as little as possible. The changes introduced can, therefore, be well accounted for by the high--frequency part of the privacy--enhanced images $I_{pr}$. We model these high--frequency alterations as noise and try to remove them using a denoising procedure. Such denoising strategies have proven useful for removal of adversarial noise~\cite{mustafa2019image}, and are, therefore, also expected to be useful for reversing soft--biometric privacy enhancement that shares characteristics with techniques based on adversarial examples. Similarly as in the previous section, we denote the denoising--based attribute--recovery procedure as $\chi_{d}: I_{pr} \mapsto I_{re}$ hereafter and again implement it using an off--the--shelf inpainting model.


\begin{figure}[t]
	\centering
	\includegraphics[width=0.99\columnwidth]{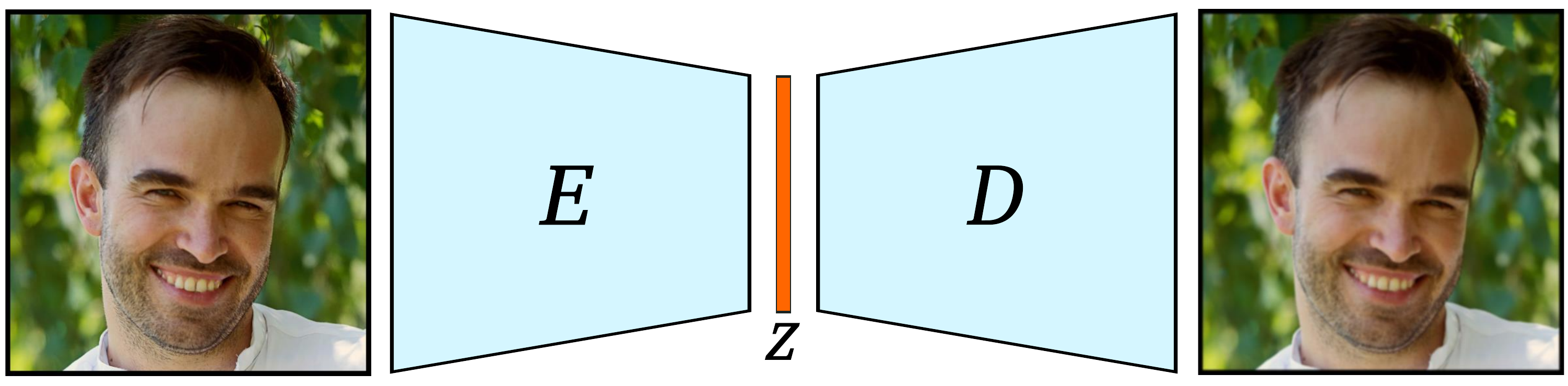}
	\\ \vspace{-1mm}
    
    	\centering
    \begin{minipage}[b]{0.24\columnwidth}
    	\centering
        {\footnotesize $I_{pr}$}
    \end{minipage}
    \hfill
    \begin{minipage}[b]{0.24\columnwidth}
    	\centering
    \end{minipage}
    \hfill
    \begin{minipage}[b]{0.24\columnwidth}
    	\centering
    \end{minipage}
    \hfill
    \begin{minipage}[b]{0.24\columnwidth}
    	\centering
        {\footnotesize $I_{re}$}
    \end{minipage}\vspace{-1mm}
	\caption{\textcolor{black}{Illustration of the attribute recovery procedure using an auto--encoder ($\chi_{a}$) for image reconstruction, proposed for PrivacyProber. The auto--encoder maps the privacy--enhanced image $I_{pr}$ to the output image $I_{re}$, such that high-frequency components, e.g., adversarial noise, are removed.}}
	\label{fig:ae}
\end{figure}

\subsubsection{\textcolor{black}{Attribute Recovery Through Adversarial Defenses\label{SubSubSec: AutoEncoder}}}

\textcolor{black}{As soft--biometric privacy--enhancing techniques often rely on adversarial noise, the robustness of privacy--enhancement can be also evaluated against existing adversarial defense methods (e.g.~\cite{folz2020adversarial}). This means that the attribute-recovery component in the proposed PrivacyProber framework (whose main goal is to evaluate the robustness of privacy--enhancing methods) can also be implemented as an arbitrary adversarial defense algorithm. Here, we consider a self-supervised approach that relies on an auto-encoder for image reconstruction, as illustrated in Fig.~\ref{fig:ae}. The auto--encoder, optimized for removal of adversarial noise from an input image, consists of an encoder $E$ and a decoder $D$. The privacy--enhanced image $I_{pr}$ is provided as an input to $E$, which compresses the image into a latent representation $z = E(I_{pr})$. The latent representation $z$ is then decoded using $D$ to produce the reconstructed image $I_{re}$, i.e. $I_{re} = D(z)$. The result of this process is, therefore, an image $I_{re}$ that is typically free of high-frequency (adversarial) noise that often serves as means for the privacy enhancement. We denote the reconstruction-based attribute recovery procedure as: $\chi_{a}: I_{pr} \mapsto I_{re}$ hereafter.} 



\subsection{Domain-specific transformation}

Instead of building on generative models, trained on clean non--tampered data, to restore facial attributes, another possibility is to base attribute recovery on specifics of the targeted image domain. With facial images, for example, automatic attribute inference should be based solely on the facial region and ignore other contextual information, e.g., background. With domain--specific transformations we, hence, aim to incorporate information on facial semantics into the attribute--recovery procedure and exclude image regions irrelevant for attribute classification from the data. 

\begin{figure}[t]
	\centering
    \begin{minipage}[b]{0.24\columnwidth}
        \includegraphics[width=0.99\columnwidth]{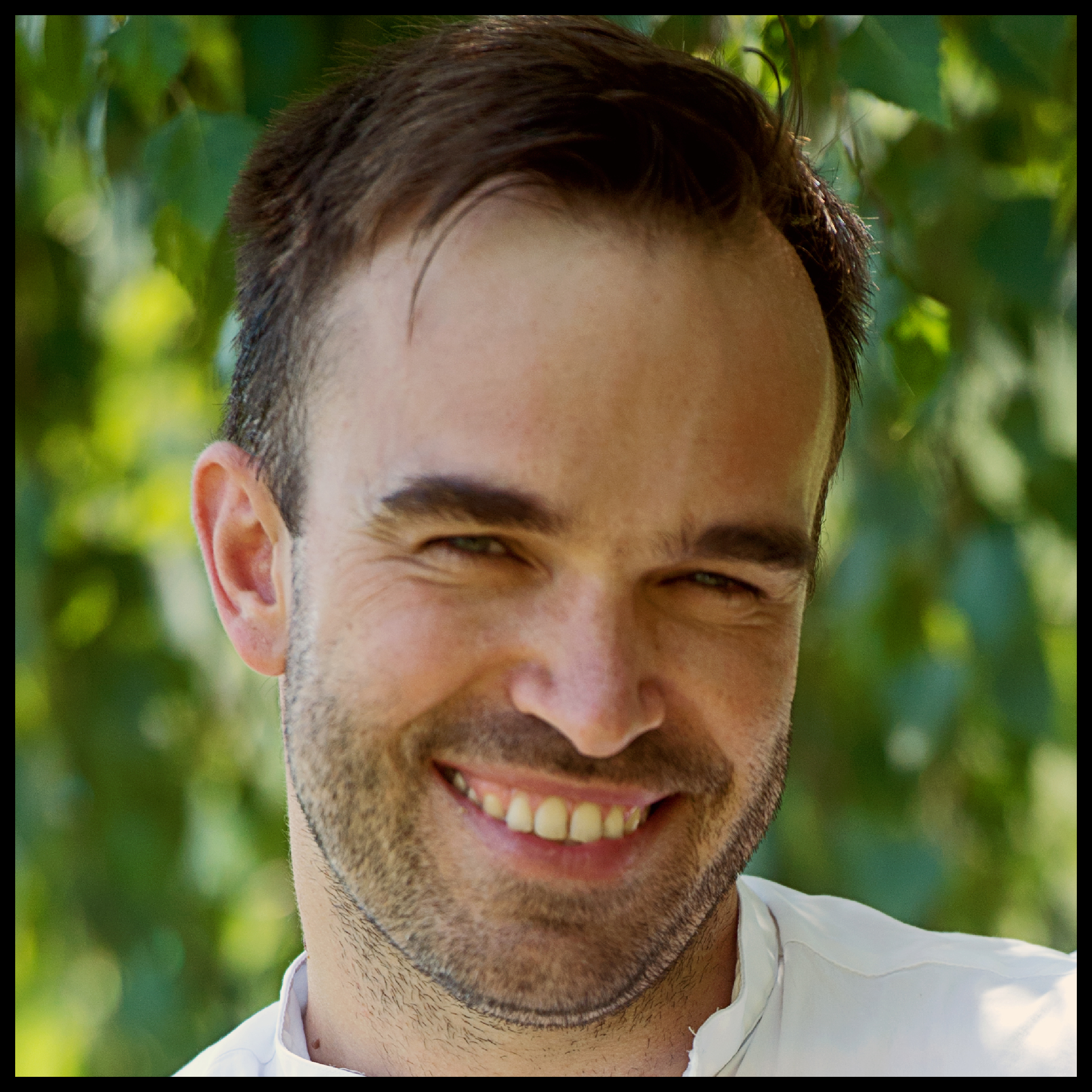}
    \end{minipage}
    \hfill
    \begin{minipage}[b]{0.24\columnwidth}
        \includegraphics[width=0.99\columnwidth]{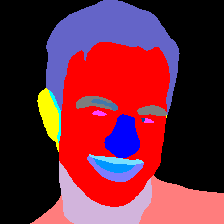}
    \end{minipage}
    \hfill
    \begin{minipage}[b]{0.24\columnwidth}
        \includegraphics[width=0.99\columnwidth]{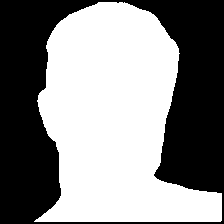}
    \end{minipage}
    \hfill
    \begin{minipage}[b]{0.24\columnwidth}
        \includegraphics[width=0.99\columnwidth]{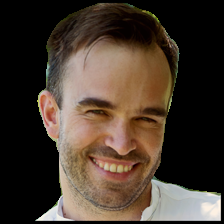}
    \end{minipage}\\ \vspace{-1mm}
    
    	\centering
    \begin{minipage}[b]{0.24\columnwidth}
    	\centering
        {\footnotesize $I_{pr}$}
    \end{minipage}
    \hfill
    \begin{minipage}[b]{0.24\columnwidth}
    	\centering
        {\footnotesize Parsing map}
    \end{minipage}
    \hfill
    \begin{minipage}[b]{0.24\columnwidth}
    	\centering
        {\footnotesize Binary mask $B$}
    \end{minipage}
    \hfill
    \begin{minipage}[b]{0.24\columnwidth}
    	\centering
        {\footnotesize $I_{re}$}
    \end{minipage}\vspace{-1mm}
	\caption{Illustration of the domain--specific attribute recovery procedure, $\chi_{br}$, proposed for  PrivacyProber. The privacy--enhanced image, $I_{pr}$, is masked with a binary mask $B$ that corresponds to the facial region. Because the background is partially affected by the privacy enhancement, focusing only on the facial area impacts the behavior of attribute classification models.}
	\label{fig:domain}
\end{figure}

We propose one domain--specific transform in this work that relies on face parsing. Specifically, given an arbitrary face parser $f_p$, we extract facial--part information from the privacy--enhanced image $I_{pr}$ and aggregate all part labels that corresponds to the facial region into a binary mask $B$. The labeled facial parts produced by the face parser and the corresponding binary mask are illustrated in Fig.~\ref{fig:domain}. Once the binary mask is constructed, it is utilized to exclude background pixels from the image with the goal of making attribute inference less susceptible to artifacts generated by the privacy enhancement:
\begin{equation}
    I_{re} = I_{pr}\odot B,
\end{equation}
where $\odot$ is gain the Hadamard product. This type of attribute recovery is denoted as $\chi_{br}: I_{pr}\mapsto I_{re}$ hereafter.

\subsection{Beyond Zero-Effort Evaluation Scenarios}

PrivacyProber, $\chi$, can in general be implemented us\-ing any combination of the transformations discussed above, i.e.,  $\{\chi_{in}, \chi_{d}, \chi_{a}, \chi_{br}\}$, 
and utilized to explore the robustness of a given soft--biometric privacy model to attribute recovery attempts. A general high--level framework for using the proposed PrivacyProber in evaluation scenarios that go beyond zero--effort recognition experiments is given in Algorithm~\ref{Algo: framework}.\vspace{1mm} 

\begin{algorithm} 
\SetAlgoLined
\SetKwInput{KwInput}{Input}                
\SetKwInput{KwOutput}{Output}              
\KwInput{Attribute classifier $\xi_a$, set of $N$ input  images $\{I_{or}\}_N$, privacy model $\psi$}
\KwOutput{Performance (robustness) estimate for $\psi$} 
\vspace{1mm}

Implement PrivacyProber, $\chi$, from $\{\chi_{in}, \chi_{d}, \textcolor{black}{\chi_{a}}, \chi_{br}\}$\; \vspace{0.5mm}

\For{Each image in $\{I_{or}\}_N$}{
Apply privacy model: $I_{pr}=\psi(I_{or})$\;

Attempt attribute recovery: $I_{re} = \chi(I_{pr})$\;

Use $\xi_a$ for classification over $I_{or}$ and $I_{re}$;     

} \vspace{0.5mm}

Calculate performance of $\xi_a$ over $\{I_{or}\}_N$ and $\{I_{re}\}_N$\;

Estimate robustness of $\psi$ based on results (see Section~\ref{Subsec: Measusres} for scoring methodology)\;
 \caption{Application of PrivacyProber \label{Algo: framework}}
\end{algorithm}

\begin{figure*}[tb]
\centering
    \includegraphics[width=0.92\textwidth]{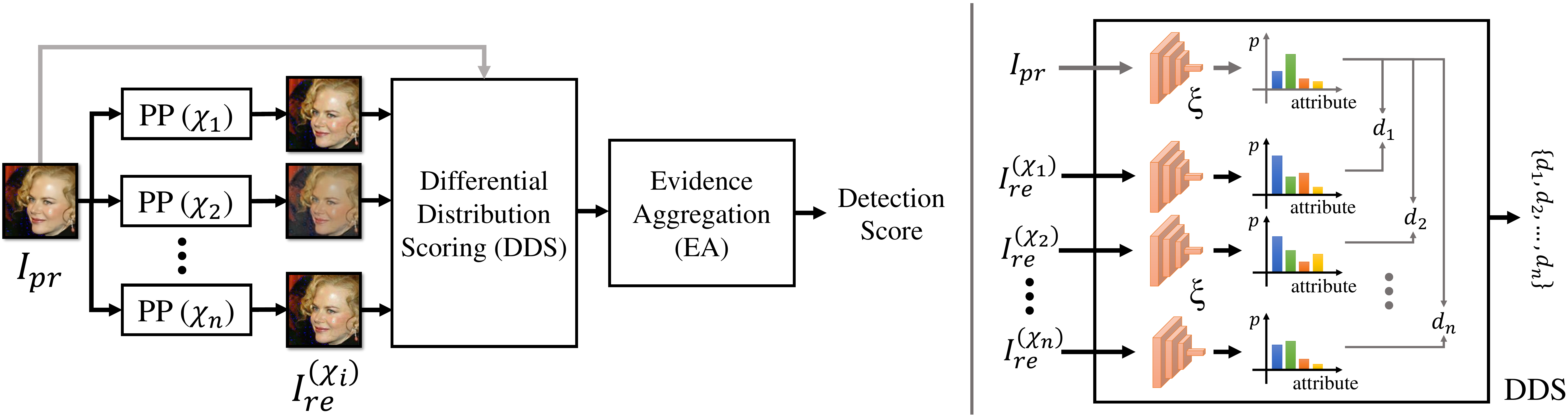}
    \caption{\textcolor{black}{High-level overview of the proposed  \textit{evidence \textbf{A}ggregation approach for \textbf{P}rivacy-\textbf{E}nhancement \textbf{D}etection} (APEND). The main idea behind the (learning-free) detection approach is to compare the posterior distributions before and after processing with different versions of PrivacyProber (PP) and exploit the (aggregated) differences in the generated distributions for privacy-enhancement detection.}}
    \label{fig:APPEND}
\end{figure*}

\subsection{Detecting (Soft--Biometric) Privacy Enhancement}


%
\textcolor{black}{The main idea behind PrivacyProber is to reconstruct facial attribute information obscured by soft-biometric privacy models. Thus, for a given privacy--enhanced input image, $I_{pr}$, a selected attribute classifier $\xi$ is expected to generate different posterior probabilities $p(C_k|I_{pr})$ than for the image processed through the proposed PrivacyProber $p(C_k|I_{re})$,  where $C_k$ denotes the $k$-th attribute class (as defined in Section~\ref{SubSec:Problam_definition}). By comparing the posteriors, it is therefore possible to determine whether an image has been tampered with or not. Based on this insight, we develop in this section a novel (proof-of-concept) approach for detecting privacy enhancements with the proposed PrivacyProber framework. }

\textcolor{black}{As illustrated in Fig.~\ref{fig:APPEND}, the proposed approach \textbf{A}ggregates evidence from multiple PrivacyProbers for \textbf{P}rivacy-\textbf{En}hancement \textbf{D}etection (APEND) and consists of the following steps:}
\begin{itemize}
    \item \textcolor{black}{\textbf{Step 1: Attribute Recovery.} In the first step, the input image $I_{pr}$ is subjected to a series of $n$ PrivacyProbers $\{\chi_i\}_{i=1}^n$ that generate a corresponding set of $n$ attribute-recovered images $\{I_{re}^{(\chi_i)}\}_{i=1}^n$. The PrivacyProbers $\{\chi_i\}_{i=1}^n$ used in this procedure are assumed to be distinct and based on different transformations.}
    \item \textcolor{black}{\textbf{Step 2: Differential Distribution Scoring (DDS).} Next, an attribute classifier $\xi$ is applied to the original input image $I_{pr}$ and the privacy-enahnced counterparts $\{I_{re}^{(\chi_i)}\}_{i=1}^n$, and posterior distributions are calculated, i.e., $p(C_k|I_{pr})$ and $\{p(C_k|I_{re}^{(\chi_i)})\}_{i=1}^n$.    To quantify the differences in the computed distributions, we use the Chi-square distance, which is an established measure for comparing histograms~\cite{liu2017local}:
    \begin{equation}
        d_i(I_{pr},I_{re}^{(\chi_i)}) = \sum_{k=1}^N\frac{\left(p(C_k|I_{pr})-p(C_k|I_{re}^{(\chi_i)})\right)^2}{\left(p(C_k|I_{pr})+p(C_k|I_{re}^{(\chi_i)})\right)},  
        \label{Eq: chi_square}
    \end{equation}
    where $i\in\{1,2,\cdots,n\}$. See the right part of Fig. \ref{fig:APPEND} for an illustration. }
    \item \textcolor{black}{\textbf{Step 3: Evidence Aggregation (EA).} Finally, we aggregate the evidence generated through the $n$ attribute recovery attempts into the final detection score $d_{fin}$ using a weighted linear combination, i.e.,
    \begin{equation}
        d_{fin} = \sum_{i=1}^Nw_id_i(I_{pr},I_{re}^{(\chi_i)}),
        \label{Eq: linear}
    \end{equation}
    where $w_i$ are balancing weights. The main motivation for the aggregation operation is to consider (complementary) recovery evidence for a more reliable detection score.} 
\end{itemize}

\textcolor{black}{It is worth noting that APEND does not require training to facilitate tampering detection. Unlike the majority of modern detection schemes, which are trained in a discriminative manner with examples of bona-fide and tampered images, APEND is \textit{training free} and \textit{knowledge-driven}, i.e., designed around the characteristics of soft-biometric privacy models.} 
\section{Evaluation of privacy models}
\label{sec:ExperimentalSetup}
In this section, we evaluate several state-of-the-art privacy enhancing techniques using standard vanilla methodology as well as the proposed PrivacyProber. The goal of these experiments is to provide insight into the performance, but more importantly robustness of biometric privacy models, and to demonstrate the importance of experimental evaluations that go beyond zero-effort recognition experiments. Furthermore, we also demonstrate how the proposed PrivacyProber can be used to detect privacy enhancement (tampering) in facial images -- something that (to the best of our knowledge) has not  been attempted \textcolor{black}{widely in the open literature so far.}

\subsection{Considered Privacy Models}

\textcolor{black}{Four} recent (soft-biometric) privacy--enhancing techniques are implemented for the experiments, i.e., the $k$-AAP method from~\cite{chhabra2018anonymizing}, the FGSM--based technique from~\cite{chatzikyriakidis2019adversarial}, the FlowSAN approach from~\cite{mirjalili2019flowsan}, and \textcolor{black}{PrivacyNet from \cite{mirjalili2020privacynet}}. 
Because FlowSAN can strike a balance between the level of privacy protection ensured and the preserved utility of the facial images, two different versions of the model are considered: \textit{i)} one with three SAN models arranged sequentially one after the other (FlowSAN--$3$ hereafter), and \textit{ii)} one with five sequential SAN models (FlowSAN--$5$). All \textcolor{black}{five} techniques are trained to obscure {\em gender information}, which is also the most frequently considered attribute in research addressing soft--biometric privacy \cite{othman2014privacy,mirjalili2019flowsan,terhorst2019suppressing,BortolatoFG2020}. The techniques are selected for the experiments because of their state--of--the--art performance and the fact that they rely on different mechanisms for ensuring soft--biometric privacy. 
As such, they serve as a representative cross-section of exiting techniques and contribute towards demonstrating the importance of evaluating biometric privacy models beyond zero--effort recognition experiments. 


\subsection{Datasets, Setup, and Performance Measures}\label{Subsec: Measusres}

To assess the performance of the considered privacy models, three publicly available face datasets are used, i.e., Labeled Faces in the Wild (LFW)~\cite{LFWTech}, MUCT~\cite{Milborrow10}, and Adience~\cite{eidinger2014adience}. The  datasets ship with the needed gender labels and contain challenging facial images captured in a wide variety of imaging conditions. 
Moreover, they represent standard datasets for assessing the performance of existing privacy--enhancing techniques (e.g,~\cite{BortolatoFG2020, mirjalili2019flowsan, mirjalili2020privacynet,PEMIU_Access2020}) and are, therefore, also used in this work.
\begin{table*}[t]
\renewcommand{\arraystretch}{1.2}
\caption{Overview of the main dataset characteristics and image splits used in the experiments. To ensure balanced experimental data  and avoid bias in the generated results, the  image splits are (approximately) gender balanced. Subjects between the training and testing part are disjoint. Testing is performed over four disjoint data splits to be able to report confidence scores on the reported results .\label{tab:databases}}
\centering
\resizebox{\textwidth}{!}{%
\begin{tabular}{lccccccccccccc}
\hline\hline
\multirow{ 2}{*}{Dataset}  &&  \multicolumn{4}{c}{Totals} && \multicolumn{2}{c}{Training} && \multicolumn{4}{c}{Testing{$^\dagger$}}\\ \cline{3-6} \cline{8-9} \cline{11-14}
&& \#Images & \#Subj. & \#AvgIm/Subj. & \#Images (Gender) && \#Images & \#Subj. && \#Images & \#Subj. & \#Mated & \#Non--mated\\
\hline
LFW~\cite{LFWTech} && $5,735$ & $1,773$& $3.39$ & $2,884$ (m),
$2,851$ (f)
  && $3,992$ & $1,425$ && $1,743$ & $348$ & $4,060$ & $9,669,923$\\
Adience~\cite{eidinger2014adience} && $10,502$ & $1,395$  & $9.22$ &  $5,238$ (m),
$5,264$ (f)  && $8,493$ & $1,006$ && $2,009$ & $389$  & $4,029$ & $9,640,738$ \\
MUCT~\cite{Milborrow10} && $3,693$ & $522$  & $13.38$ & $1,803$ (m), $1,890$ (f) && $3,119$ & $276$ && $574$ & $246$  & $1,191$ & $1,027,277$ \\
\hline\hline
\multicolumn{12}{l}{\footnotesize $^\dagger$ Totals over all 4 test data splits; \#AvgIm/Subj -- average number of images per subject; m -- male, f -- female;}
\vspace{-3mm}
\end{tabular}}
\end{table*}

All images are roughly aligned prior to the experiments such that the faces are 
cropped to exclude background pixels and then rescaled to a standard size of $224\times 224$ pixels. The preprocessed images are subjected to privacy enhancement and used in the experimental assessment to evaluate the following aspects of the privacy models:
{\begin{enumerate}[label=(\alph*),leftmargin=*,align=left]\vspace{2mm}
    \item \textbf{Level of soft--biometric privacy enhancement:} The performance of the evaluated privacy models is measured through gender ($g$) classification experiments with predefined classifiers on the original ($o$) and privacy--enhanced ($p$) images. The level of privacy enhancement achieved is determined by comparing  ROC curves generated from the two image sets. Additionally, the \textit{gender suppression rate (SR)}~\cite{terhorst2019Unsupervised} is also reported and defined in this work as:
    \begin{equation}
        SR = \max \left(\frac{AUC_{go}-f(AUC_{gp})}{AUC_{go}},0\right)\in[0,1],    
        \label{eq: suppression_rate}
    \end{equation}
    where $AUC_{go}$ and $AUC_{gp}$ denote the area under the ROC curve before  and after privacy enhancement, respectively, and the normalization function $f(\cdot)$ corresponds to:
    \begin{equation}
        f(x) =    
        \begin{cases}
            x,& \text{if } inverted \ decision\\
            \max(x-0.5,0),              & \text{otherwise}
        \end{cases}
        .
        \label{eq: truncation}
    \end{equation}
    The above definition allows us to report performance for (gender) privacy models that aim to induce misclassifications (i.e., invert classifier predictions for binary problems --  invert ROC curves) as well as for models trying to induce random gender--classification probabilities (i.e., targeting a random AUC score of $0.5$) using a \textit{single} performance measure. A SR value of $1$ indicates perfect attribute (gender) suppression, whereas a value of $0$ implies that the suppression has no 
    effect. Evaluating privacy models with SR scores and the approach presented above corresponds to zero-effort (vanilla) evaluation strategies commonly seen in the literature~\cite{mirjalili2018semi,mirjalili2019flowsan}.    
    \item \textbf{Utility preservation:} Soft-biometric privacy--enhancing techniques aim to retain as much of the original image information as possible, while 
    altering the visual appearance of the input images only slightly. In accordance with standard evaluation methodology~\cite{othman2014privacy,terhorst2019suppressing,mirjalili2019flowsan,mirjalili2020privacynet} 
    utility preservation is, therefore, assessed through verification experiments on the original and privacy--enhanced images, where minimal differences in performance are expected. Because there exists a trade--off between utility preservation and attribute suppression, a quantitative measure taking both tasks into account is reported for the experiments. Specifically, a modified version of the \textit{privacy--gain identity--loss coefficient (PIC)} is used in this work~\cite{terhorst2020evaluation,BortolatoFG2020,terhorst2019Unsupervised}, i.e.:
     \begin{equation}
        PIC = SR - IL,    
        \label{eq: PIC}
    \end{equation}
    where the {\em identity loss IL} is defined with the degradation in verification performance after the privacy enhancement:
    \begin{equation}
        IL = \max \left(\frac{AUC_{vo}-AUC_{vp}}{AUC_{vo}},0\right).
        \label{eq: identity_loss}
    \end{equation}
    In the above equations $AUC_{vo}$ and $AUC_{vp}$ are the $AUC$ scores of the verification experiments ($v$) generated with the original ($o$) and privacy--enhanced ($p$) images. PIC is bounded to $[-1,1]$, with a PIC score of $1$ implying an ideal trade--off, i.e., no loss in verification performance and perfect attribute suppression. 
    \item \textbf{Robustness:} While attribute suppression and utility preservation are standard aspects of privacy models typically evaluated in zero--effort evaluation scenarios, robustness of the models to attribute recovery attempts has \textit{so far seen limited attention in the literature}. Here, we use the proposed PrivacyProber to recover information suppressed by the studied privacy models. 
    Identity verification and gender classification experiments are then conducted after attribute recovery ($r$) 
    and the generated ROC curves are analyzed to assess robustness. A scalar robustness measure is derived from the ROC curves in this paper, i.e., the \textit{attribute--recovery robustness (ARR)}, and reported in the experiments, i.e.:
     \begin{equation}
        ARR = g(AUC_{gp})\cdot\frac{|AUC_{go}-AUC_{gr}|}{AUC_{go}}
        \label{eq: recovery_rate}
    \end{equation}
    where the subscripts suggest that the AUC score was computed from the recovered ($r$), privacy--enhanced ($p$) or original ($o$) images and $g(x)$, i.e.,
     \begin{equation}
        g(x) =    
        \begin{cases}
            1, & \text{if } x\leq0.5; inverted \ decision\\
            2,              & \text{otherwise}
        \end{cases}
        \label{eq: adjustment factor}
    \end{equation}
    ensures that robustness is measured for all privacy models on the scale $[0,1]$. Thus, ARR scores  serve as a measure of robustness to attribute recovery attempts and take a value of $0$ if after the recovery the same performance is achieved as with the original images and a value close to $1$ if no information can be inferred from the privacy--enhanced images. 
\end{enumerate}}

To facilitate the evaluation described above, the experimental datasets are split into training and testing parts. The training parts are used to train gender classifiers for each dataset and matchers for the verification experiments. Because the datasets are unbalanced with respect to gender, the number of male and female subjects in the training and testing sets is (approximately) balanced by randomly excluding images of the more represented gender. It is also made sure that at least two images per identity are present in the testing part for the verification experiments. 
To ensure a consistent evaluation setup across all datasets and be able to report confidence scores for the experiments, the test images are partitioned into $4$ experimental splits. Gender recognition experiments are performed for every test image, while a fixed number of mated and non-mated comparison is performed for the verification experiments. Details on the experimental setup are provided in Table~\ref{tab:databases}.
\begin{figure*}[t]
\begin{minipage}{0.015\textwidth}
\begin{turn}{90}
  \scriptsize{ \hspace{0mm} Adience \hspace{17mm} MUCT \hspace{17mm} LFW}
  \end{turn}
\end{minipage}
\begin{minipage}{0.985\textwidth}
\begin{subfloat}
    \centering
    \includegraphics[width=0.19\textwidth,
    trim=4mm 0mm 4mm 3mm,
    clip]{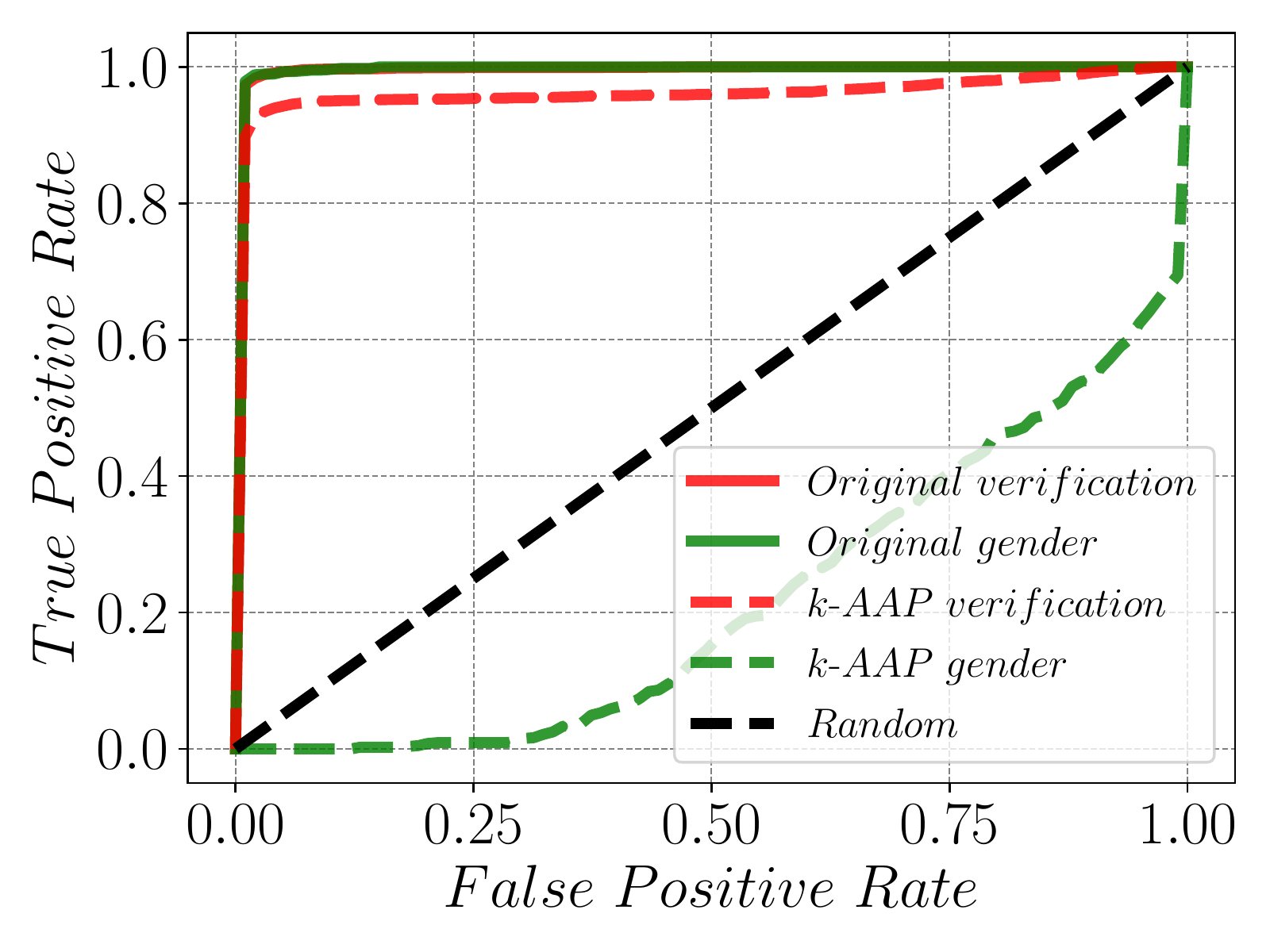}
\end{subfloat}%
\hfill
\begin{subfloat}
    \centering
    \includegraphics[width=0.19\textwidth,
    trim=4mm 0mm 4mm 3mm,
    clip]{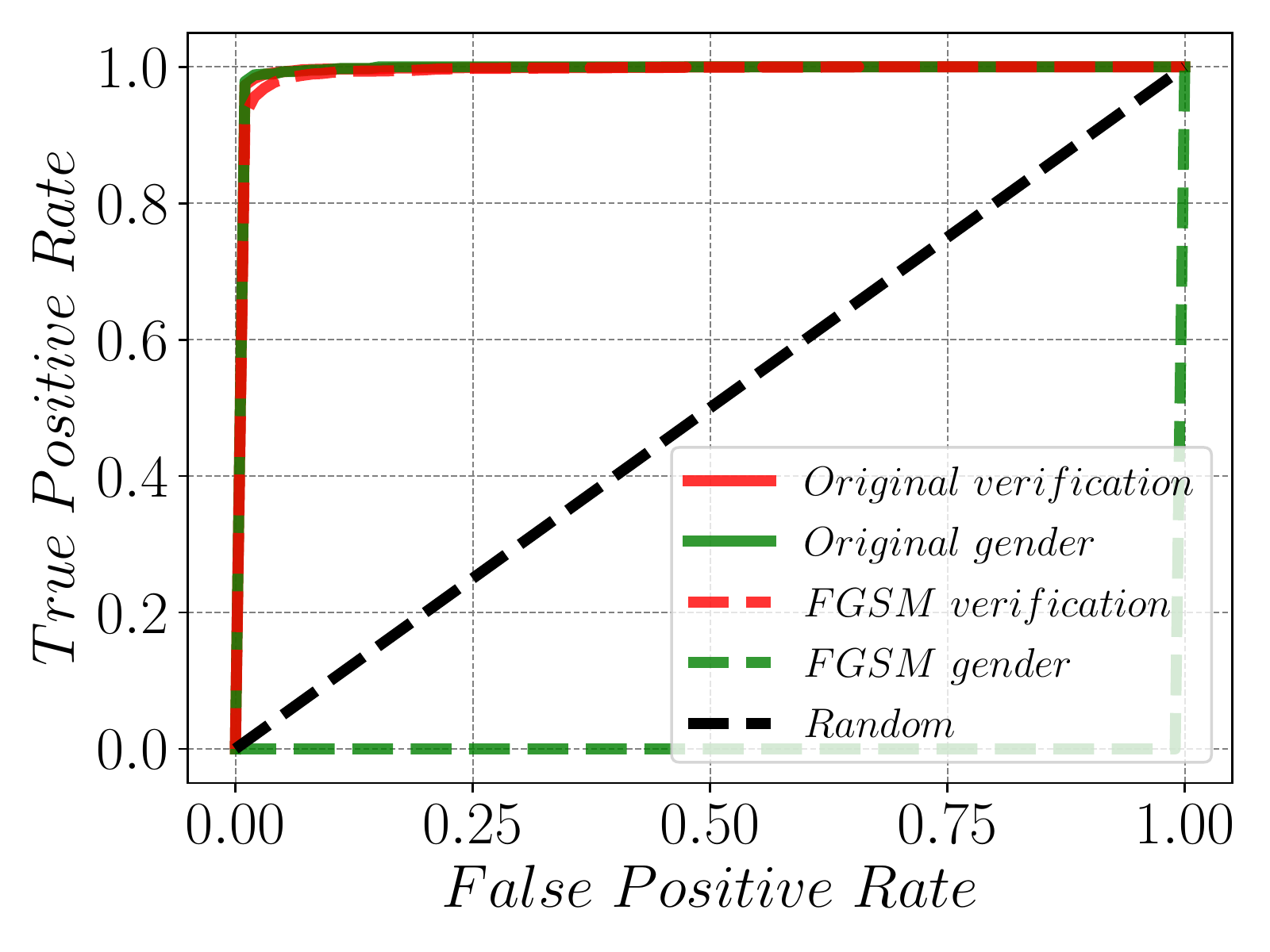}
\end{subfloat}%
\hfill
\begin{subfloat}
    \centering
    \includegraphics[width=0.19\textwidth,
    trim=4mm 0mm 4mm 3mm,
    clip]{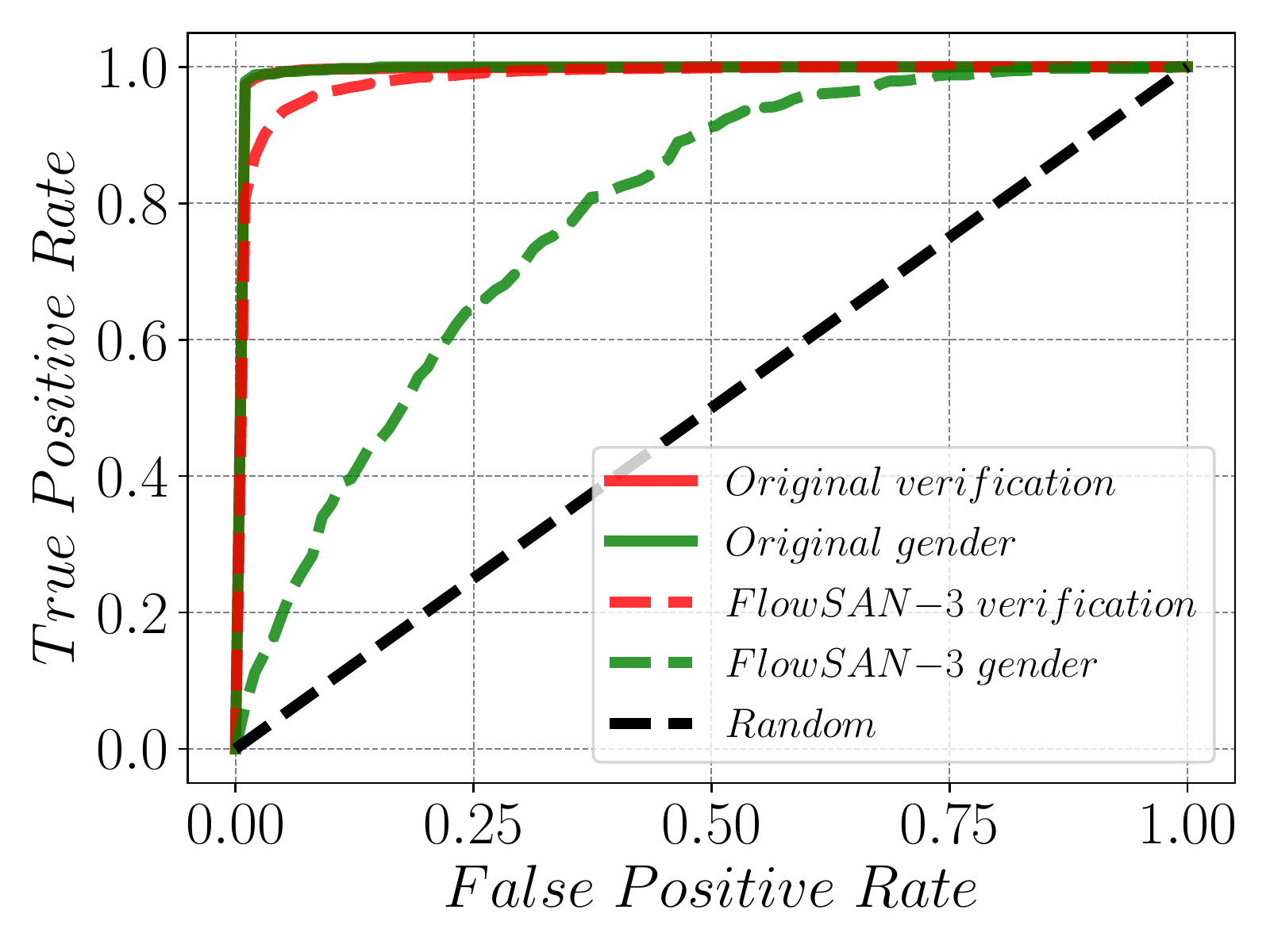}
\end{subfloat}%
\hfill
\begin{subfloat}
    \centering
    \includegraphics[width=0.19\textwidth,
    trim=4mm 0mm 4mm 3mm,
    clip]{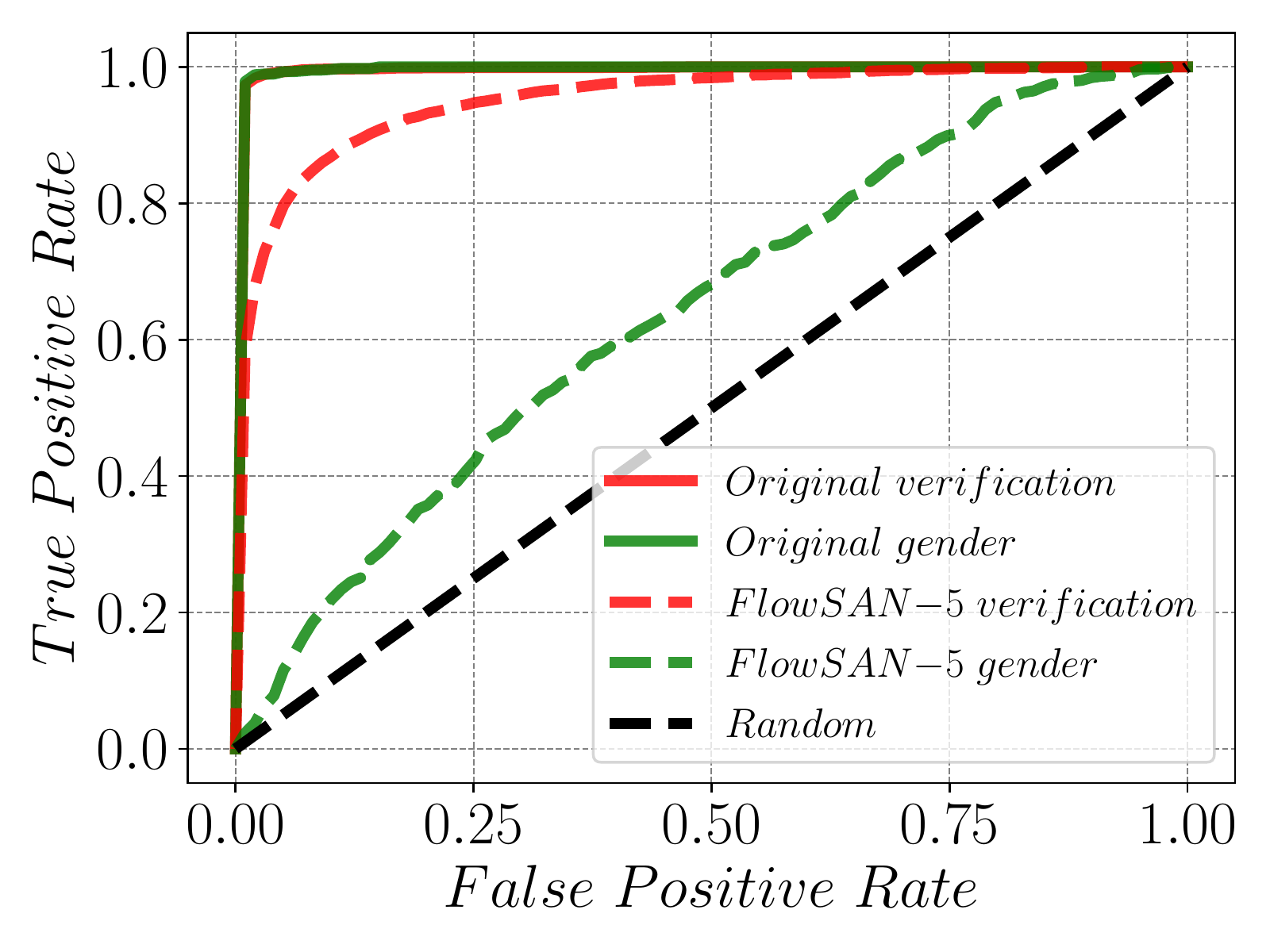}
\end{subfloat}
\hfill
\begin{subfloat}
    \centering
    \includegraphics[width=0.19\textwidth,
    trim=4mm 0mm 4mm 3mm,
    clip]{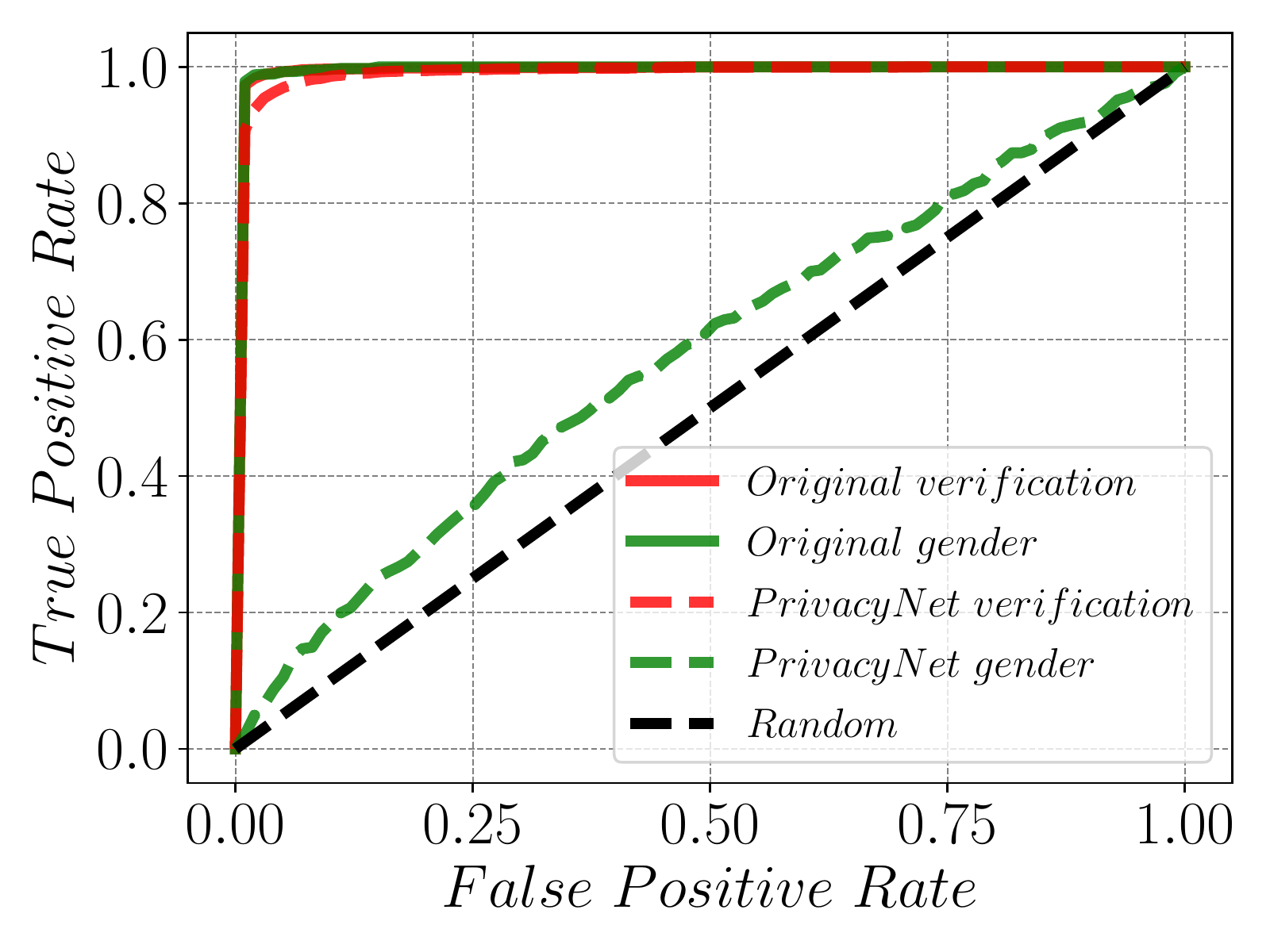}
\end{subfloat} \\
\begin{subfloat}
    \centering
    \includegraphics[width=0.19\textwidth,
    trim=4mm 0mm 4mm 3mm,
    clip]{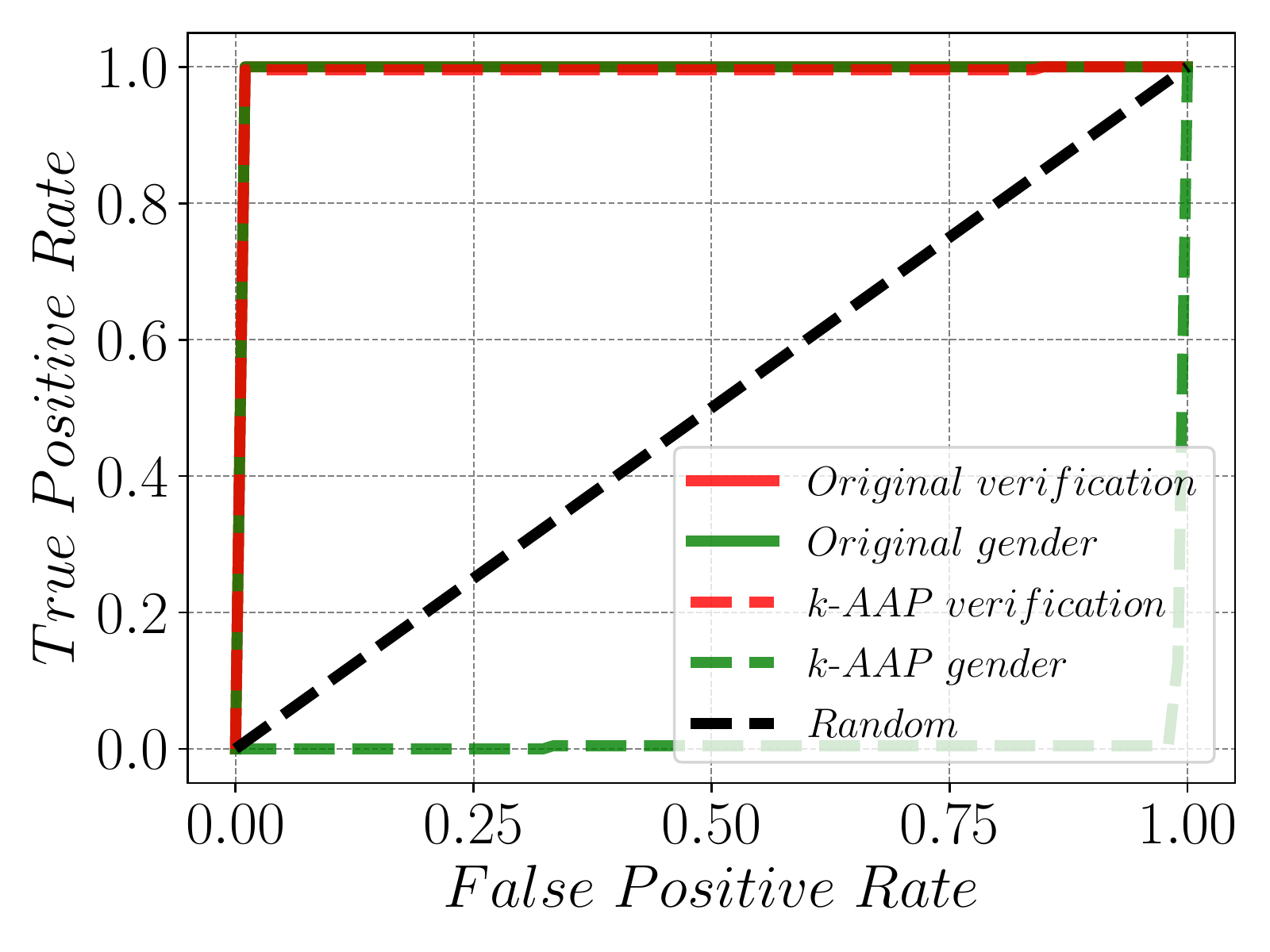}
\end{subfloat}%
\hfill
\begin{subfloat}
    \centering
    \includegraphics[width=0.19\textwidth,
    trim=4mm 0mm 4mm 3mm,
    clip]{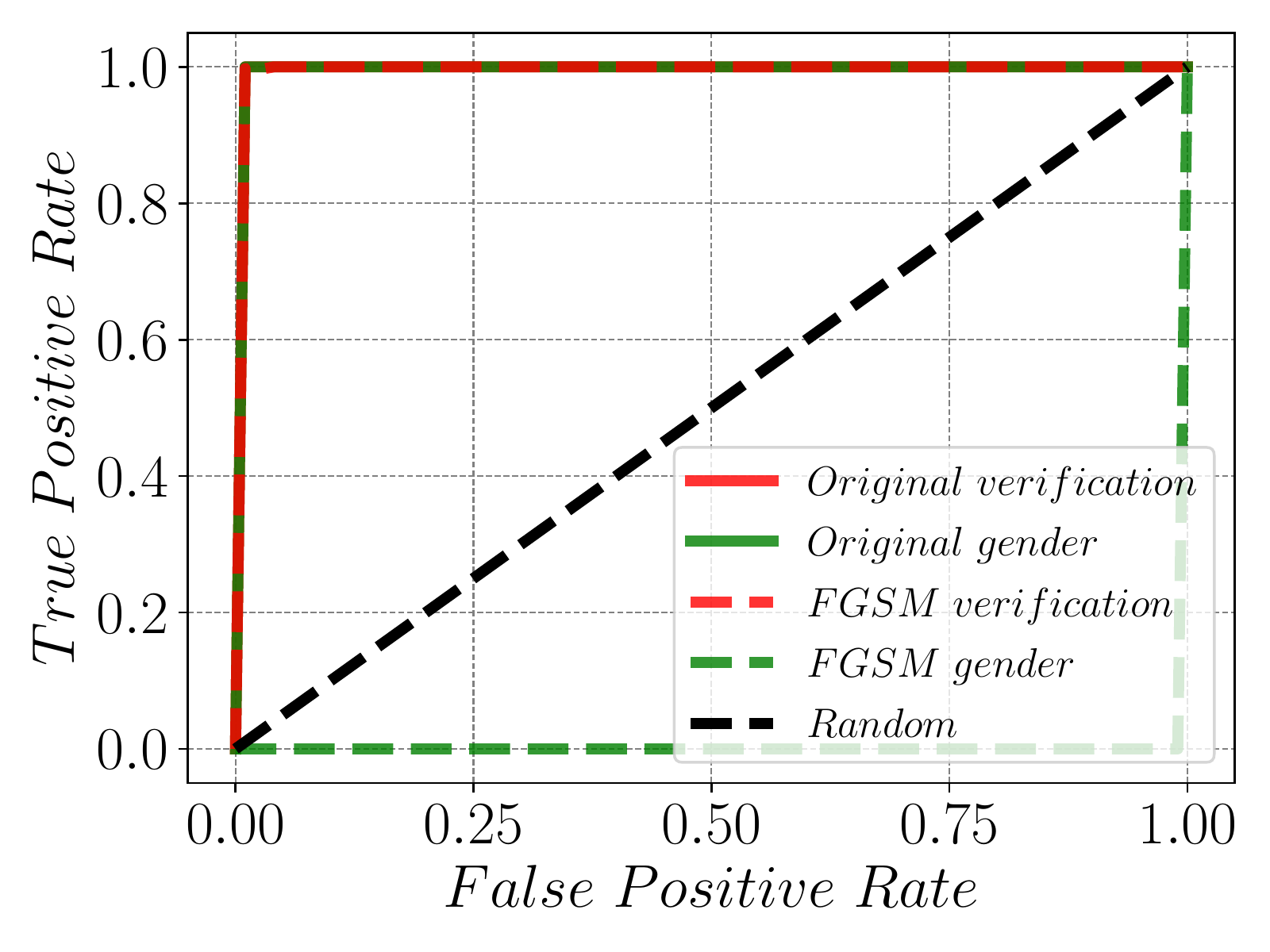}
\end{subfloat}%
\hfill
\begin{subfloat}
    \centering
    \includegraphics[width=0.19\textwidth,
    trim=4mm 0mm 4mm 3mm,
    clip]{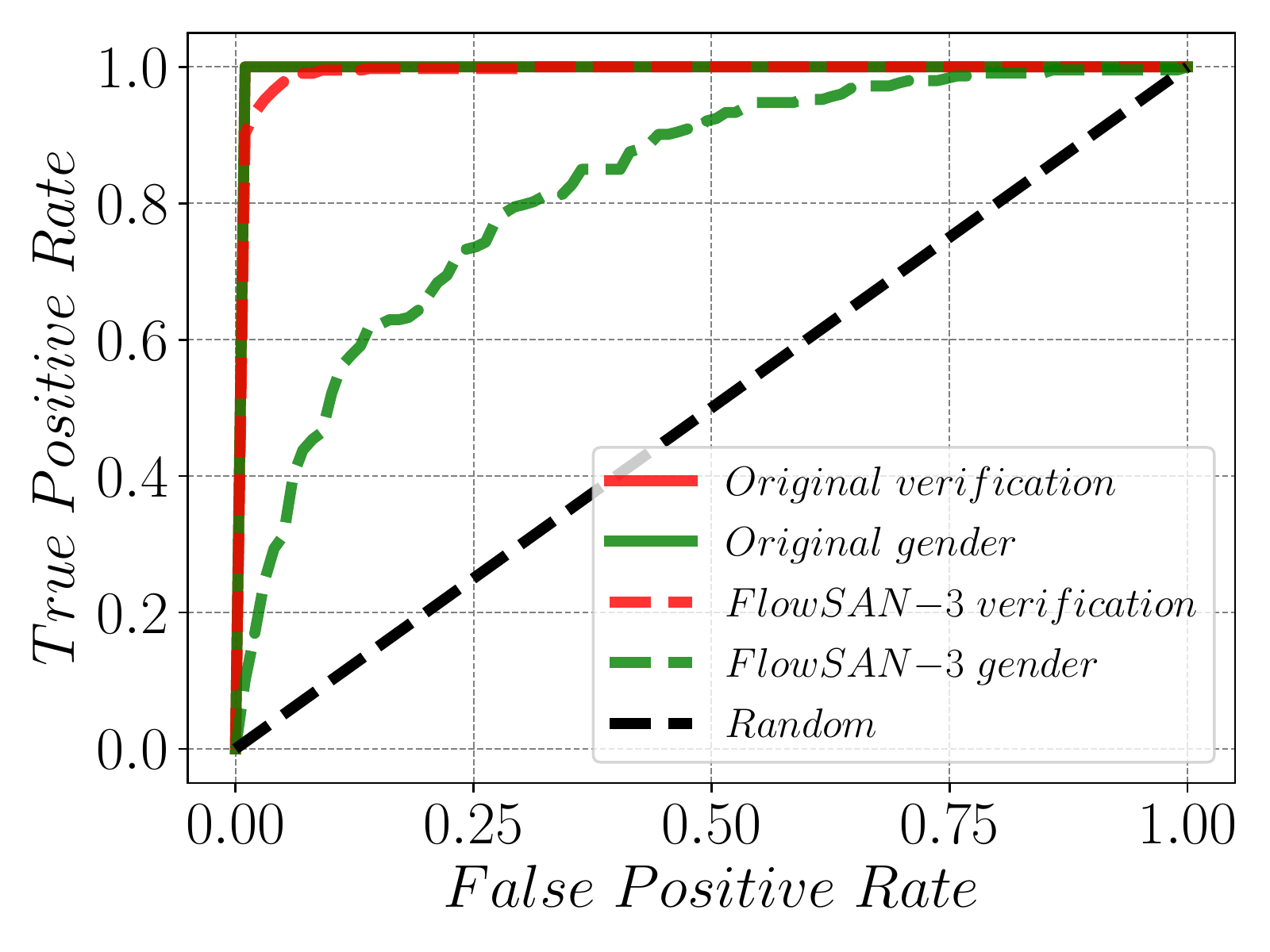}
\end{subfloat}%
\hfill
\begin{subfloat}
    \centering
    \includegraphics[width=0.19\textwidth,
    trim=4mm 0mm 4mm 3mm,
    clip]{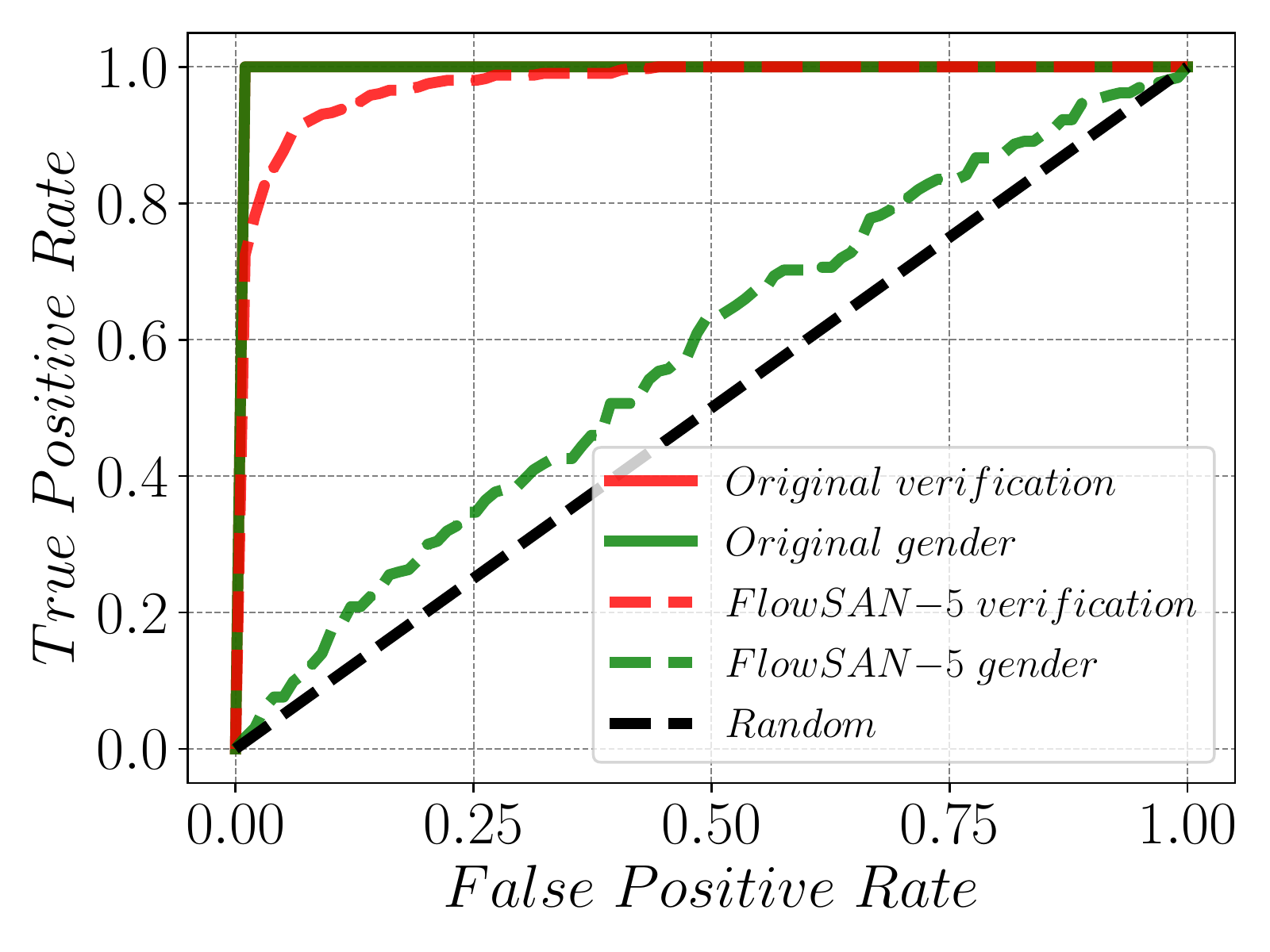}
\end{subfloat}%
\hfill
\begin{subfloat}
    \centering
    \includegraphics[width=0.19\textwidth,
    trim=4mm 0mm 4mm 3mm,
    clip]{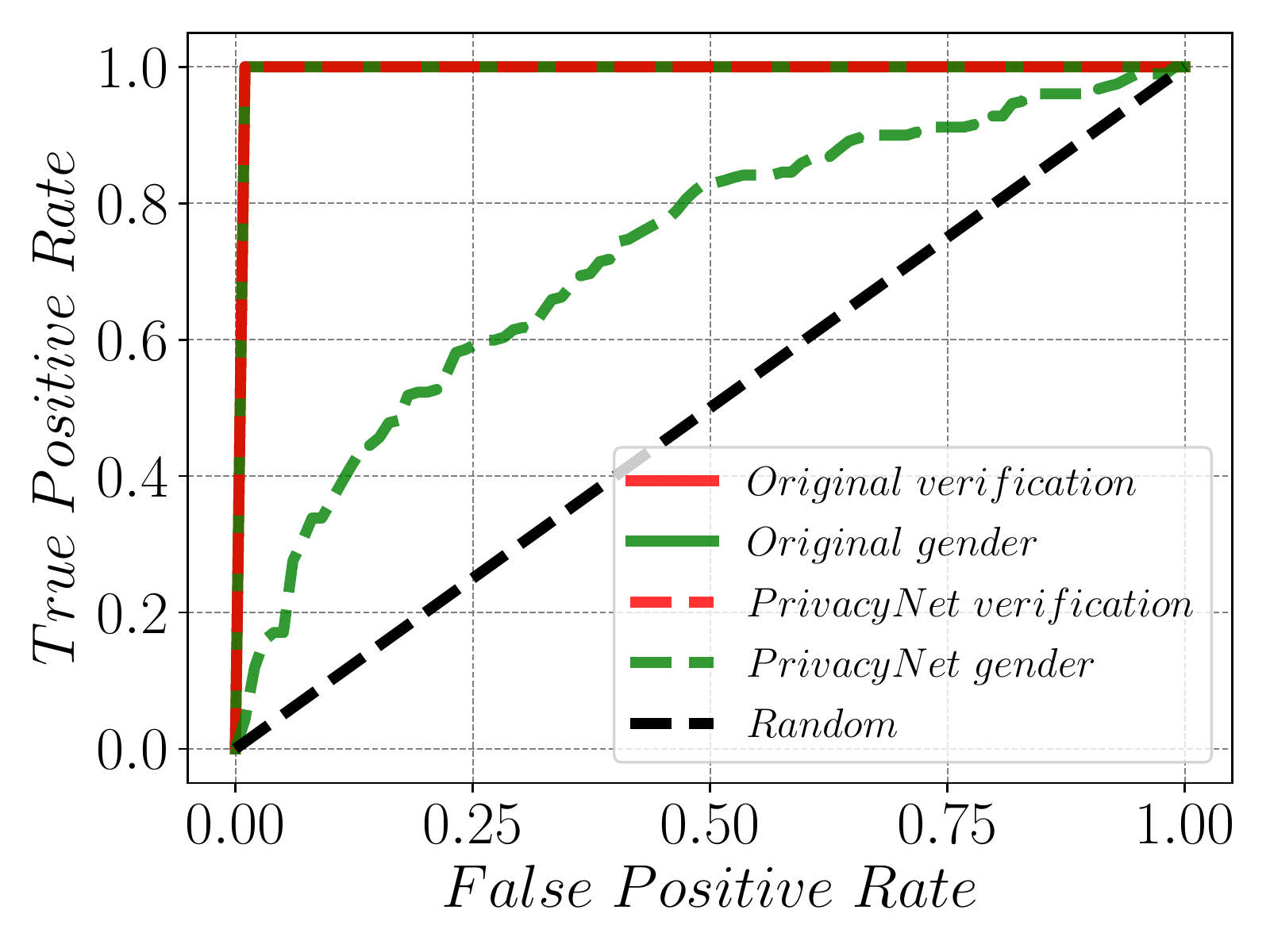}
\end{subfloat} \\
\begin{subfloat}
    \centering
    \includegraphics[width=0.19\textwidth,
    trim=4mm 0mm 4mm 3mm,
    clip]{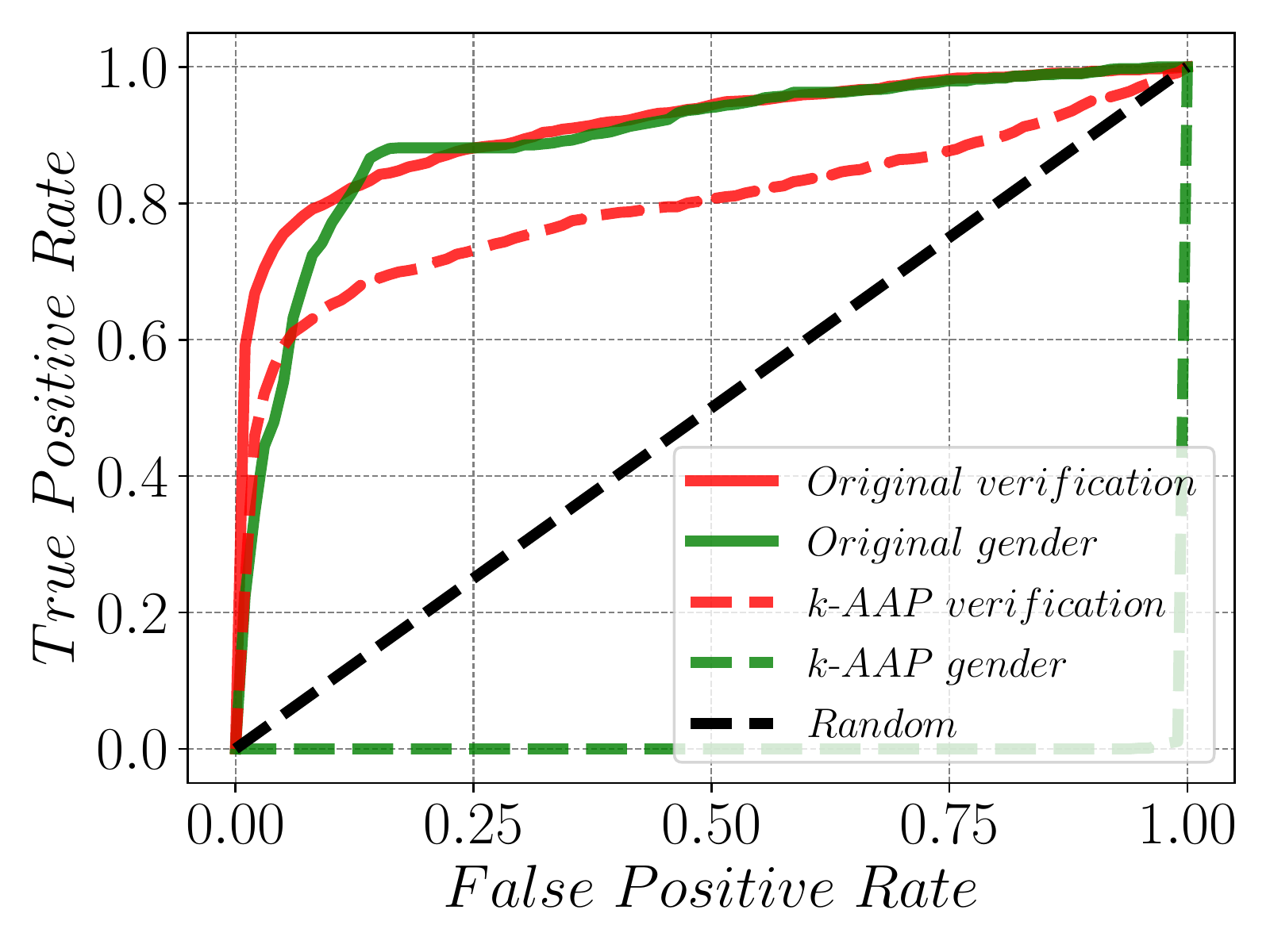}
\end{subfloat}%
\hfill
\begin{subfloat}
    \centering
    \includegraphics[width=0.19\textwidth,
     trim=4mm 0mm 4mm 3mm,
    clip]{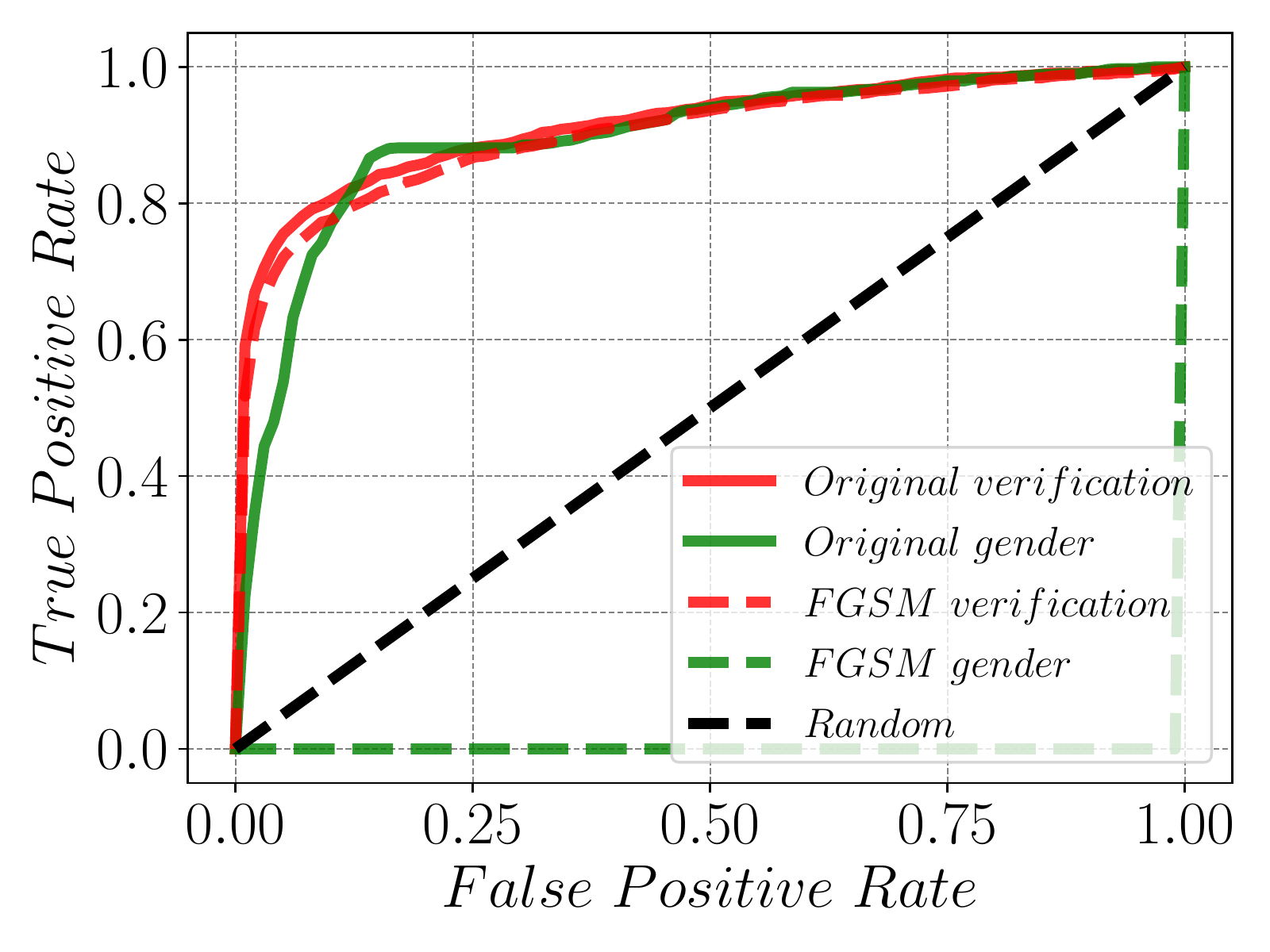}
\end{subfloat}%
\hfill
\begin{subfloat}
    \centering
    \includegraphics[width=0.19\textwidth,
    trim=4mm 0mm 4mm 3mm,
     clip]{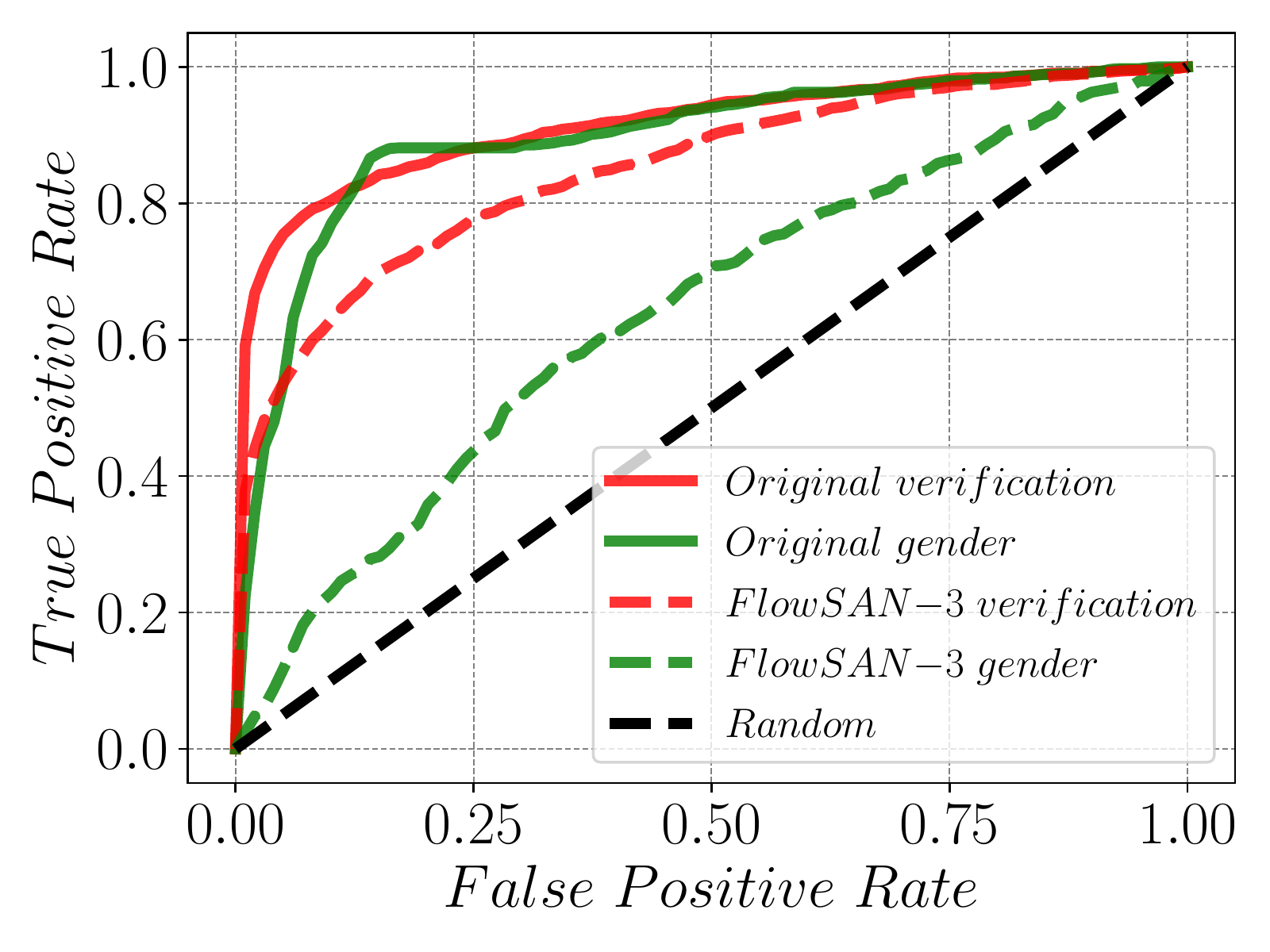}
\end{subfloat}%
\hfill
\begin{subfloat}
    \centering
    \includegraphics[width=0.19\textwidth,
     trim=4mm 0mm 4mm 3mm,
     clip]{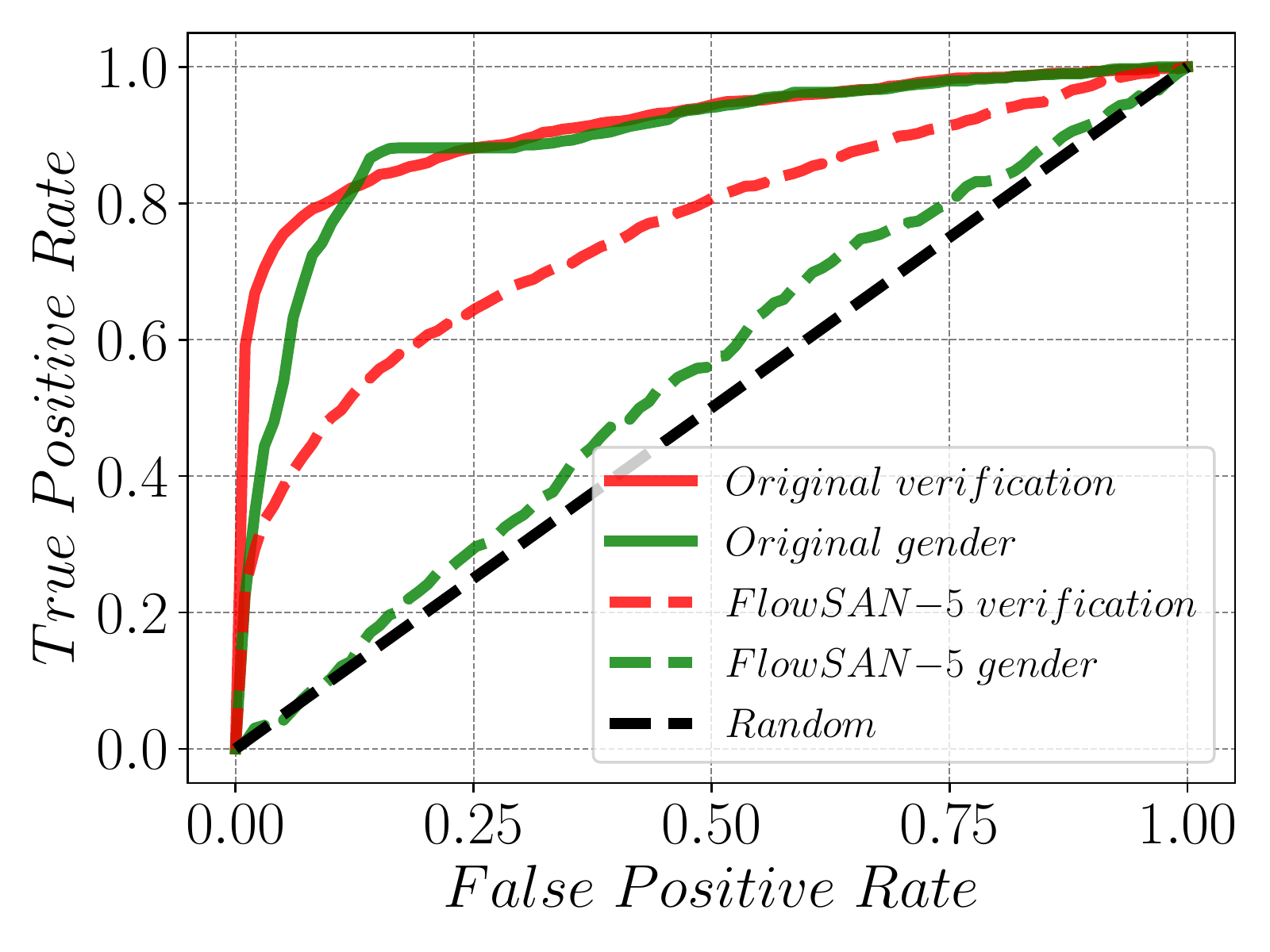}
\end{subfloat}%
\hfill
\begin{subfloat}
    \centering
    \includegraphics[width=0.19\textwidth,
     trim=4mm 0mm 4mm 3mm,
     clip]{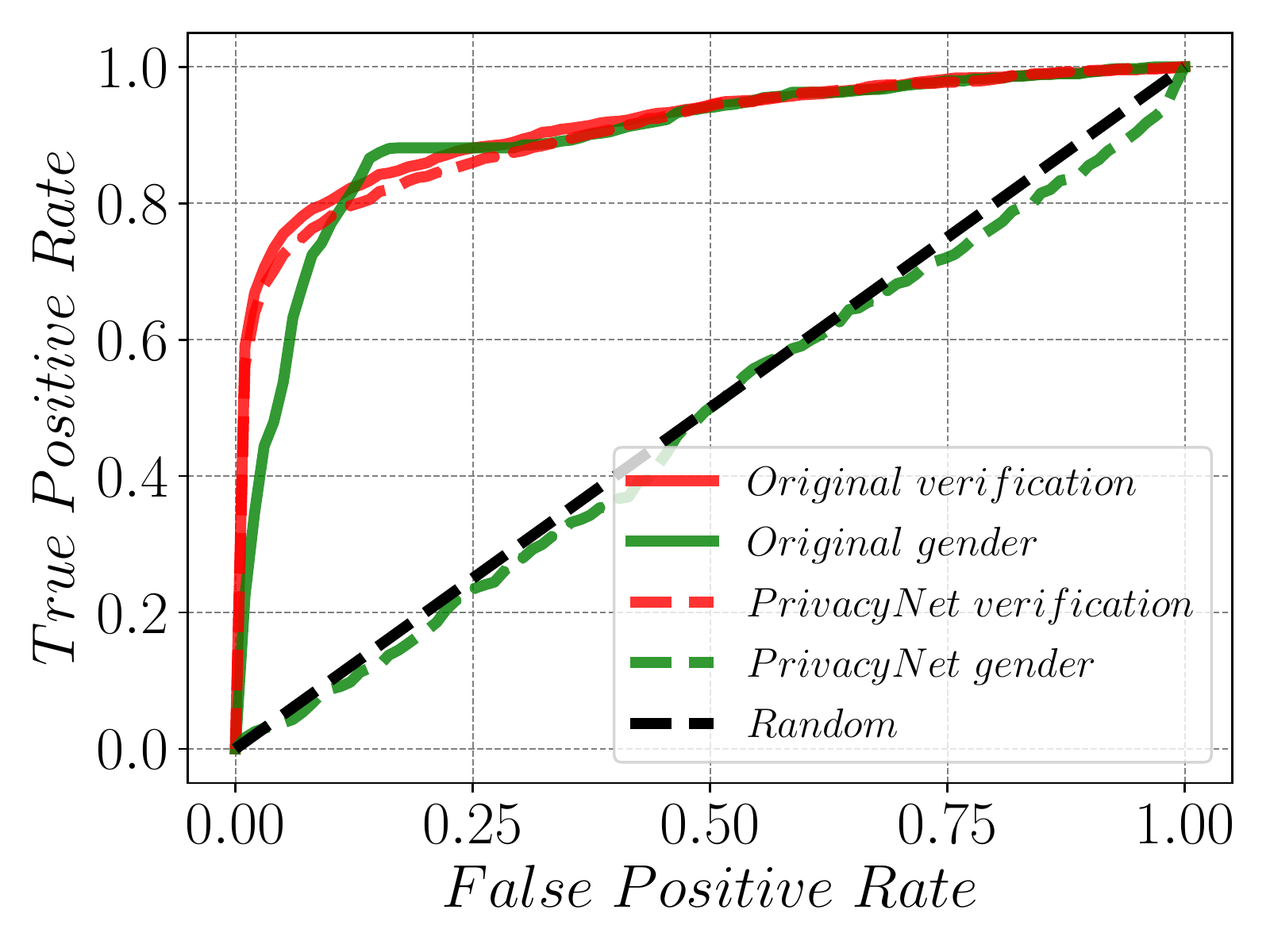}
\end{subfloat}
\end{minipage}\\
\begin{minipage}{0.19\textwidth}
\centering
\vspace{1mm}
\text{\small \ \ \ \hspace{9mm}(a) $k$--AAP} 
\end{minipage}
\hfill
\begin{minipage}{0.19\textwidth}
\centering
\vspace{1mm}
\text{\small \ \ \ \hspace{7mm}(b) FGSM}
\end{minipage}
\hfill
\begin{minipage}{0.19\textwidth}
\centering
\vspace{1mm}
\text{\small \ \ \ \hspace{6mm}(c) FlowSAN--3}
\end{minipage}
\hfill
\begin{minipage}{0.19\textwidth}
\centering
\vspace{1mm}
\text{\small \ \ \ \hspace{5mm}(d) FlowSAN--5}
\end{minipage}
\begin{minipage}{0.19\textwidth}
\centering
\vspace{1mm}
\text{\small \ \ \ \hspace{5mm}(e) PrivacyNet}
\end{minipage}
\caption{ROC curves displaying the privacy vs. utility trade--off ensured by the evaluated privacy models. Results for different datasets are shown in rows and for different privacy models in columns. All evaluated models are in general able to retain a signifant portion of the verification performance (red curves), while resulting in different gender suppression rates (green curves). Note again that $k$--AAP and FGSM aim to induce misclassifications (i.e., invert classifier predictions for binary problems), whereas the FlowSAN models aim to produce random gender classification results. Best viewed in color. \label{fig: ROC curves}}
\end{figure*}


\subsection{Baseline Evaluation of Privacy Models}

The first series of experiments studies the performance of the considered privacy models using standard vanilla evaluation methodology. The goal of these experiments is to establish the baseline performance of the models and explore their characteristics. 

\subsubsection{Privacy vs. Utility}

From an operational point of view, a critical aspect of (soft--biometric) privacy enhancement 
is the trade--off between privacy protection and utility preservation the models ensure. The first part of our evaluation, therefore, looks at this 
trade--off through a series of recognition experiments. 

\textbf{1) Baseline performance.} To establish the baseline performance of the privacy models in terms of attribute suppression rates, a gender classifier $\xi_g$ (a VGG16--based model with a two--class softmax at the top) is trained for each of the datasets and used to steer the privacy enhancement\footnote{For the FlowSAN models several such classifiers are trained using different training data configurations similarly to~\cite{Mirjalili2018GenderPA}.}. In accordance with standard evaluation methodology \cite{chhabra2018anonymizing,rozsa2019facial}, the same classifier is also used to evaluate gender recognition performance with the enhanced images on each dataset. Similarly, a ResNet--50 face recognition model~\cite{cao2018vggface2} is learned for the utility preservation experiments on images from the VGGFace2 dataset. Here, the output of the last fully connected layer of the learned model is utilized as the feature representation of the input face images. The computed representations are then matched with the cosine similarity measure in verification experiments.

\begin{figure*}[!th]
\begin{subfloat}
    \centering
    \includegraphics[width=0.33\textwidth,
     trim=2mm 3mm 5mm 3mm,
     clip]{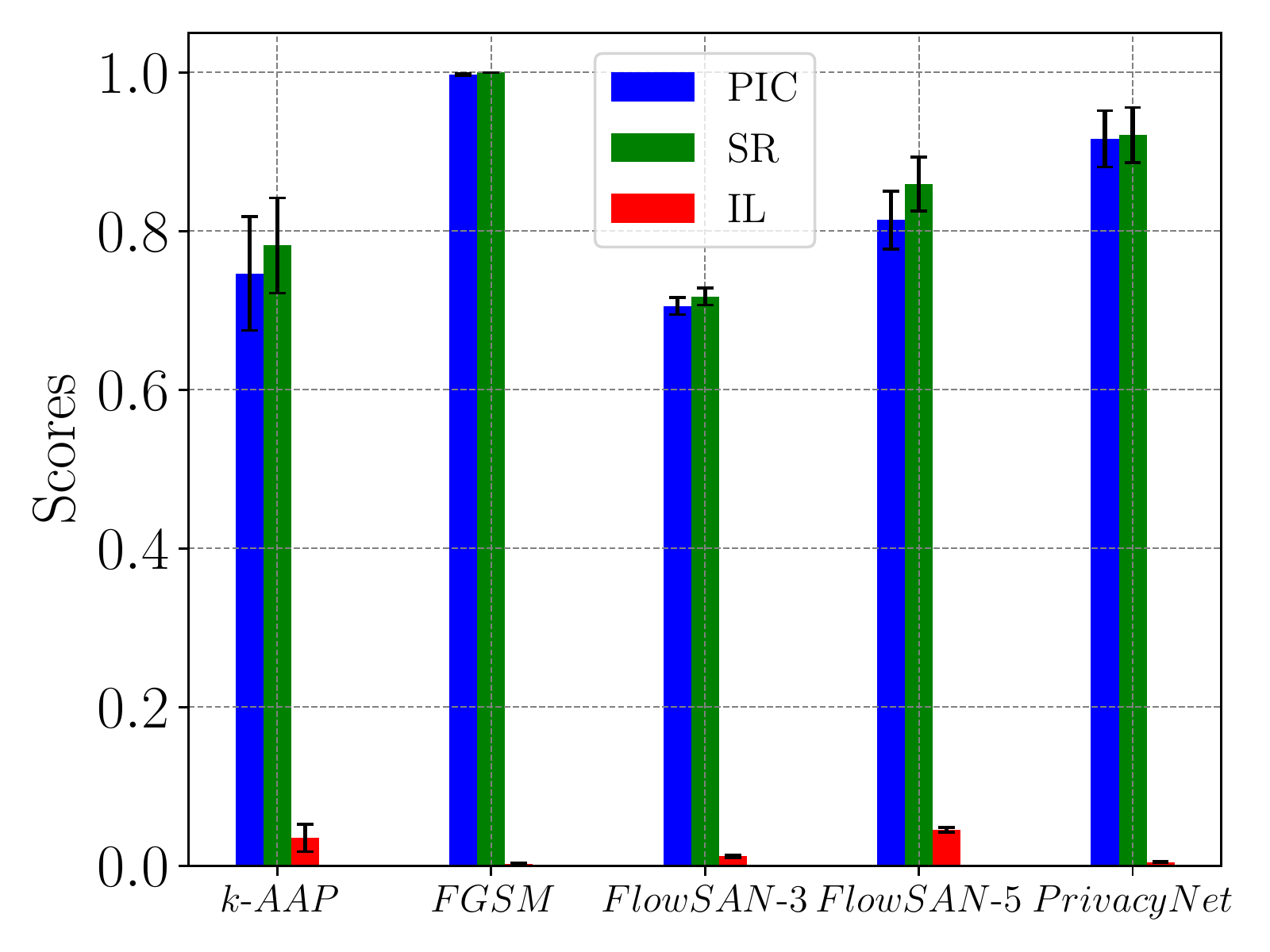}
\end{subfloat}
\begin{subfloat}
    \centering
    \includegraphics[width=0.33\textwidth,
     trim=2mm 3mm 5mm 3mm,
     clip]{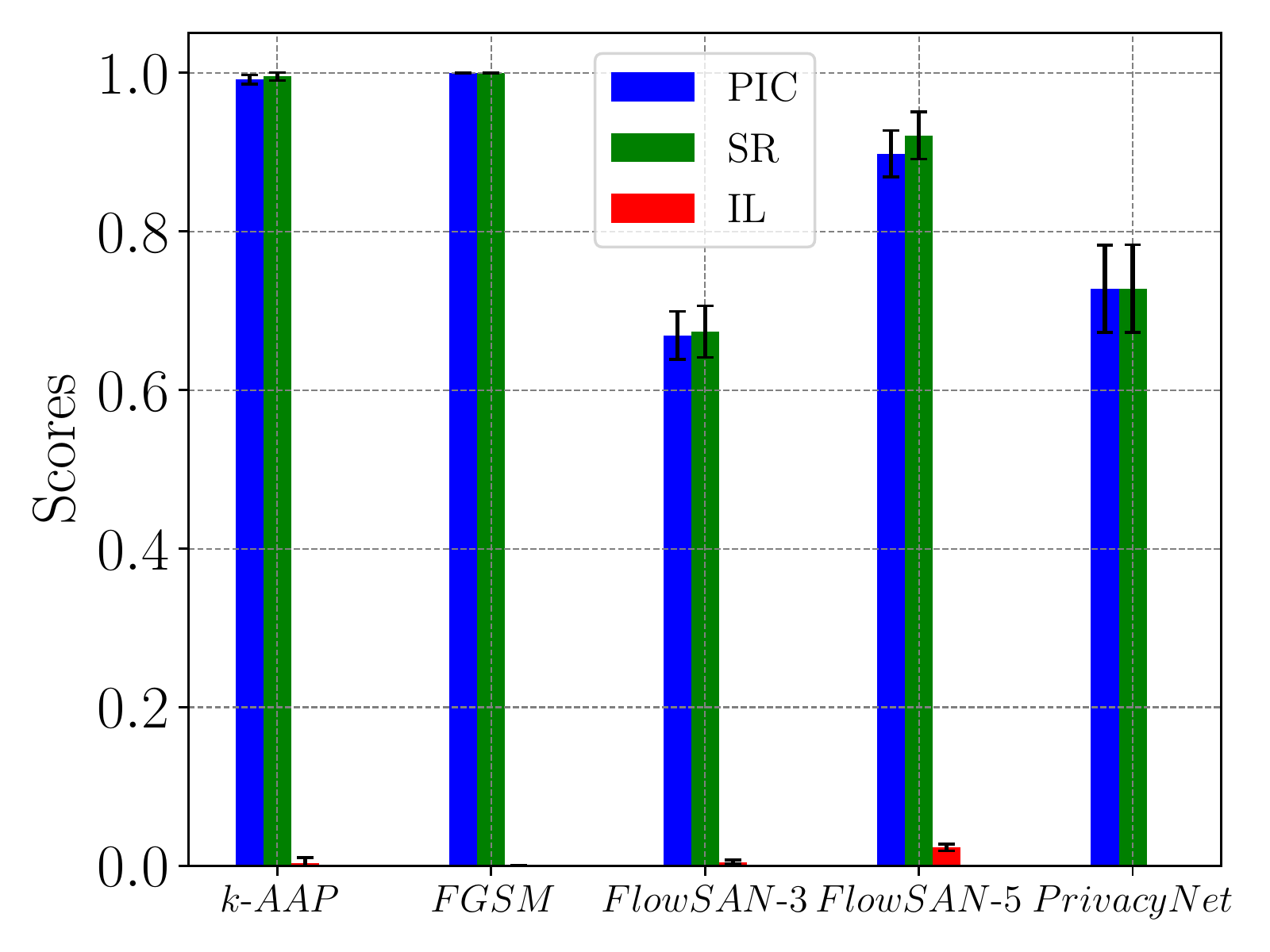}
\end{subfloat}
\begin{subfloat}
    \centering
    \includegraphics[width=0.33\textwidth,
     trim=2mm 3mm 5mm 3mm,
     clip]{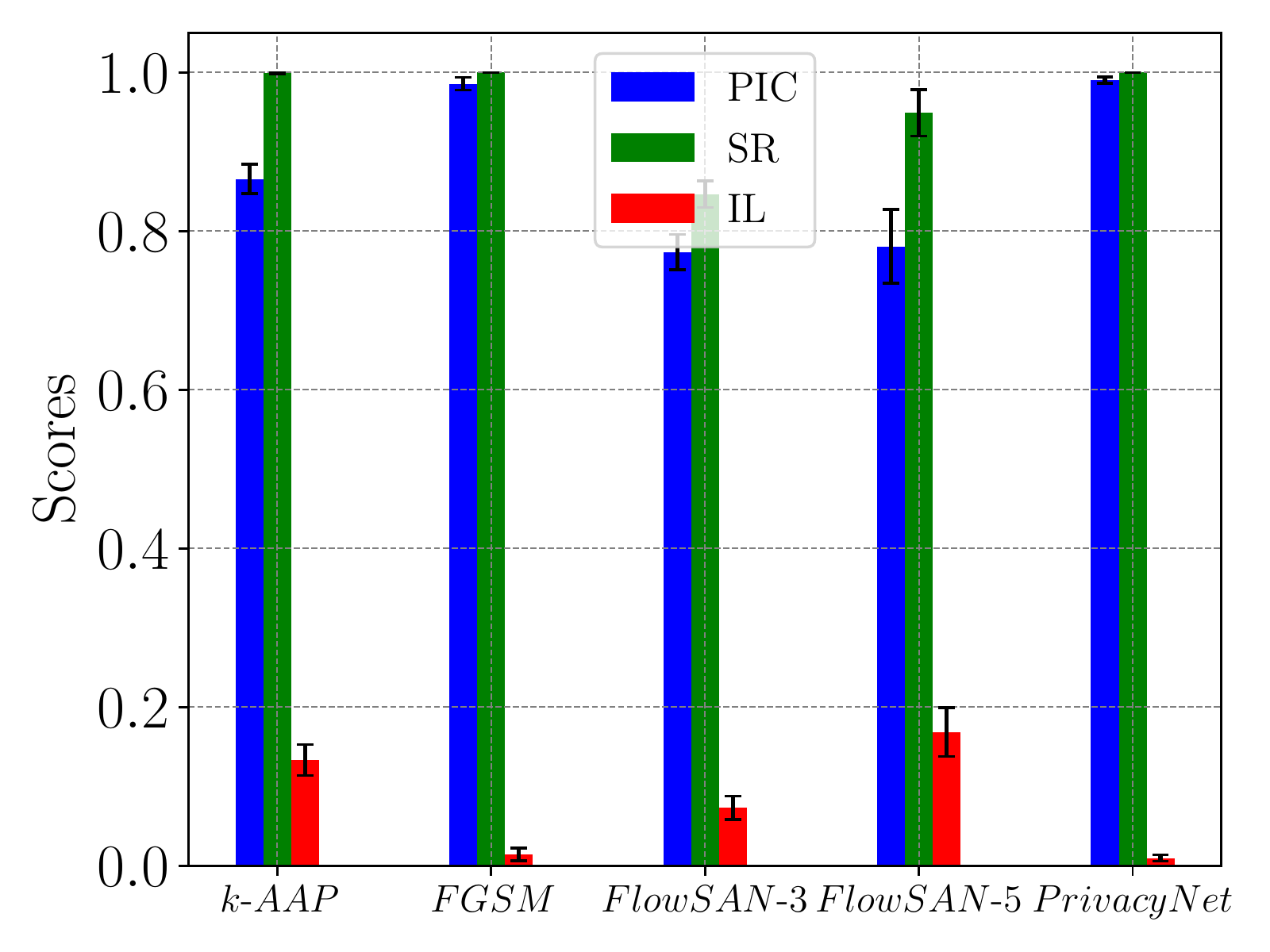}
\end{subfloat}\\
\begin{minipage}{0.33\textwidth}
\centering
\vspace{1mm}
\text{\small \ \ \ \hspace{9mm}(a) LFW} 
\end{minipage}
\hfill
\begin{minipage}{0.33\textwidth}
\centering
\vspace{1mm}
\text{\small \ \ \ \hspace{7mm}(b) MUCT}
\end{minipage}
\hfill
\begin{minipage}{0.33\textwidth}
\centering
\vspace{1mm}
\text{\small \ \ \ \hspace{6mm}(c) Adience}
\end{minipage}
\caption{Scalar performance indicators for the privacy vs. utility trade--off of the privacy models. Results are reported for three datasets and in the form of mean PIC (higher is better), SR (higher is better) and IL (lower is better) scores with corresponding standard errors computed over the $4$ test splits - see Table~\ref{tab:databases} for the experimental setup.\label{fig: scalar scores}}
\end{figure*}

Fig.~\ref{fig: ROC curves} shows the ROC curves of the 
experiments on the three experimental datasets before and after privacy enhancement. Note that confidence intervals are not included to keep the plots uncluttered. Instead, standard errors are reported for the (scalar) performance scores in Fig.~\ref{fig: scalar scores}. As can  be seen, the two privacy models that aim to induce misclassifications, $k$--AAP and FGSM, result in close to ideal gender suppression rates (SP) of $1$ on all datasets, except for LFW, where the Carlini--Wagner attack used with $k$--AAP is not successful on several test images using our optimization parameters. The suppression rates for the FlowSAN models (which aim to produce random gender classification probabilities for each input image), on the other hand, depend on the number of SAN models used in the sequence. FlowSAN--3, for example, generates lower SR scores than FlowSAN--5, but is, therefore, retaining more identity information, as evidenced by the lower identity losses (ILs) in Fig~\ref{fig: scalar scores}. \textcolor{black}{PrivacyNet improves on both Flow-SAN models both in terms of gender suppression and identity loss, and overall ensures the best overall privacy-utility trade-off among the evaluated synthesis-based privacy models.} When comparing $k$--AAP and FGSM to the FlowSAN and \textcolor{black}{PrivacyNet} models, we observe that the former two models ensure higher overall PIC scores on most dataset. 
However, this is a consequence of the fact that the FlowSAN and PrivacyNet models do not simply invert classifier probabilities and, therefore, target a more challenging problem, which makes these models applicable to a wider range of application domains. Especially successful in terms of the privacy--utility trade--off is FGSM, which achieves PIC scores of close to $1$ on all three datasets.   

While most of the findings discussed above hold for all considered datasets, there are slight differences with the Adience dataset. Here, the verification as well as gender recognition performance with the original images is lower compared to the other two datasets. This is due the characteristics of the dataset, which features real--world images with extreme appearance variations that differ significantly from those present in LFW and MUCT. As a result, even minor (additional) image degradations lead to significant performance drops, which is reflected in the relatively larger IL scores for all methods on this dataset.  
\begin{table*}[!tb]
\centering
\caption{Generalization ability of the privacy models to an unseen gender classifier $\xi_g^{u}$. The reported results are computed with a gender classifiers that differs from the one used for privacy enhancement in terms of training data and model topology. Results are reported (in terms of $\mu\pm\sigma$ computed over $4$ test data splits) for gender suppression (SR), identity loss (IL) and the the combined PIC score. The colored numbers show the relative change $\Delta$ of the mean score compared to the baseline performance, the arrows and colors indicate whether the score increased (up, blue) or decreased (down, red).}
\renewcommand{\arraystretch}{1.15}
\label{tab: generalization classifiers}
\begin{tabular}{lrrrrrr} \hline\hline
Privacy Model & Dataset   & PIC  & $\Delta$PIC (in \%) & SR & $\Delta$SR  (in \%) &  IL  \\\hline
\multirow{ 2}{*}{$k$--AAP}  & MUCT  & $0.664\pm 0.028$ & {\color{red}$32.8\%$ $\downarrow$} & $0.668\pm 0.030$ & {\color{red}$32.7\%$ $\downarrow$} & $0.004\pm 0.006$ \\
                            & Adience   & $0.688\pm 0.039$ & {\color{red}$17.8\%$ $\downarrow$} & $0.822\pm 0.033$ & {\color{red}$17.7\%$ $\downarrow$} & $0.133\pm 0.019$
\\
                            \arrayrulecolor{gray}\hline
\multirow{ 2}{*}{FGSM} & MUCT  & $0.766\pm 0.031$ & {\color{red}$23.4\%$ $\downarrow$} & $0.766\pm 0.031$ & {\color{red}$23.4\%$ $\downarrow$} & $0.000\pm 0.000$\\
                            & Adience   & $0.698\pm 0.030$ & {\color{red}$28.8\%$ $\downarrow$} & $0.712\pm 0.032$ & {\color{red}$20.6\%$ $\downarrow$} & $0.014\pm 0.008$ \\
\arrayrulecolor{gray}\hline

\multirow{ 2}{*}{FlowSAN--3} & MUCT  & $0.702\pm 0.071$ & {\color{blue}$3.3\%$ $\uparrow$}  & $0.707\pm 0.073$ & {\color{blue}$3.3\%$ $\uparrow$} & $0.005\pm 0.003$\\
                            & Adience   & $0.721\pm 0.281$ & {\color{red}$5.3\%$ $\downarrow$} & $0.794\pm 0.278$ & {\color{red}$5.3\%$ $\downarrow$} & $0.073\pm 0.015$ \\
\arrayrulecolor{gray}\hline
\multirow{ 2}{*}{FlowSAN--5} & MUCT & $0.709\pm 0.027$ & {\color{red}$18.9\%$ $\downarrow$} & $0.732\pm 0.025$ & {\color{red}$18.9\%$ $\downarrow$} & $0.023\pm 0.004$ \\
                            & Adience & $0.643\pm 0.271$ & {\color{red}$13.8\%$ $\downarrow$}& $0.812\pm 0.265$ & {\color{red}$13.7\%$ $\downarrow$} & $0.169\pm 0.031$ \\
                            
\arrayrulecolor{gray}\hline
\multirow{ 2}{*}{PrivacyNet} & MUCT & $0.711\pm 0.030$ & {\color{red}$6.1\%\downarrow$} & $ 0.711 \pm 0.030 $ &{\color{red}$6.1\%\downarrow$} & $0.000\pm 0.000$ \\
                            & Adience & $0.949\pm 0.039$ & {\color{red}$0.5\%\downarrow$} & $0.959\pm 0.039$ & {\color{red}$0.5\%\downarrow$} & $0.010\pm 0.004$ \\                            
\arrayrulecolor{black}
\hline \hline
\end{tabular}
\end{table*}

\textbf{2) Generalization to unseen classifiers.} The results discussed above were generated with the same gender classifier, $\xi_g$, that was also used for privacy enhancement on each dataset. To evaluate the generalization ability of the privacy models to unseen classification models, 
a ResNet--50 gender classifier, $\xi_g^{u}$, is trained on LFW in the next series of experiments and deployed on the remaining two datasets, i.e., MUCT and Adience. Thus, the model used for scoring gender recognition accuracy differs in topology and training data from the model utilized for privacy enhancement.

Table~\ref{tab: generalization classifiers} provides a summary of the PIC, SR and IL scores generated for this experiment. Here, the colored numbers show the relative change (marked as $\Delta$) in the computed scores when compared to the baseline performance 
from Fig.~\ref{fig: scalar scores}. As expected, all privacy models degrade in performance, both in terms PIC as well as SR score compared to the baseline experiments. 
The relative drop is quite severe for the adversarial techniques. The PIC scores drop by $17.8\%$ on Adience and by $32.8\%$ on MUCT for $k$--AAP, and by $23.4\%$ on MUCT and $28.8\%$ on Adience for FGSM. FlowSAN--3, on the other hand, achieves an increase of $3.3\%$ in terms of PIC on  MUCT and a drop of $5.3\%$ on Adience. FlowSAN--5 results in larger performance degradations, but still less so than $k$--AAP or FGSM. \textcolor{black}{Finally, PrivacyNet is observed to generalize the best overall, with PIC scores drops if $6.1\%$ on MUCT and only $0.5\%$ on Adience.} 

Overall, the results of these experiments suggest that all tested privacy models offer a certain level of robustness to unseen classifiers, but the relative drop in performance differs significantly from model to model. The FlowSAN models, which exploit several gender classifiers for (soft--biometric) privacy enhancement, appear to be more robust to changes in the classification model used and may be preferred \textcolor{black}{over k--AAP and FGSM} if no assumptions can be made regarding the target classifier used with the final application. \textcolor{black}{Even more impressive robustness was observed for the PrivacyNet model, which was trained in an adversarial setting and found to be the most robust w.r.t. unseen classifiers in the vanilla evaluation scenario.} 



\subsubsection{Feature Distribution Exploration\label{Sec:featureDistributionLFW}} 
\begin{figure}[t]
\begin{minipage}[b]{0.162\columnwidth}
    \centering
    {\scriptsize Original}
\end{minipage}%
\begin{minipage}[b]{0.162\columnwidth}
    \centering
    {\scriptsize $k$--AAP}
\end{minipage}%
\begin{minipage}[b]{0.162\columnwidth}
    \centering
    {\scriptsize FGSM}
\end{minipage}%
\begin{minipage}[b]{0.162\columnwidth}
    \centering
    {\scriptsize FlowSAN-3}
\end{minipage}%
\begin{minipage}[b]{0.162\columnwidth}
    \centering
    {\scriptsize FlowSAN-5}
\end{minipage}
\hspace{-0.2cm}
\begin{minipage}[b]{0.162\columnwidth}
    \centering
    {\scriptsize PrivacyNet}
\end{minipage}
\centering
\begin{minipage}[b]{0.162\columnwidth}
    \centering
    \includegraphics[width=1\columnwidth, trim=0 10 0 0, clip]{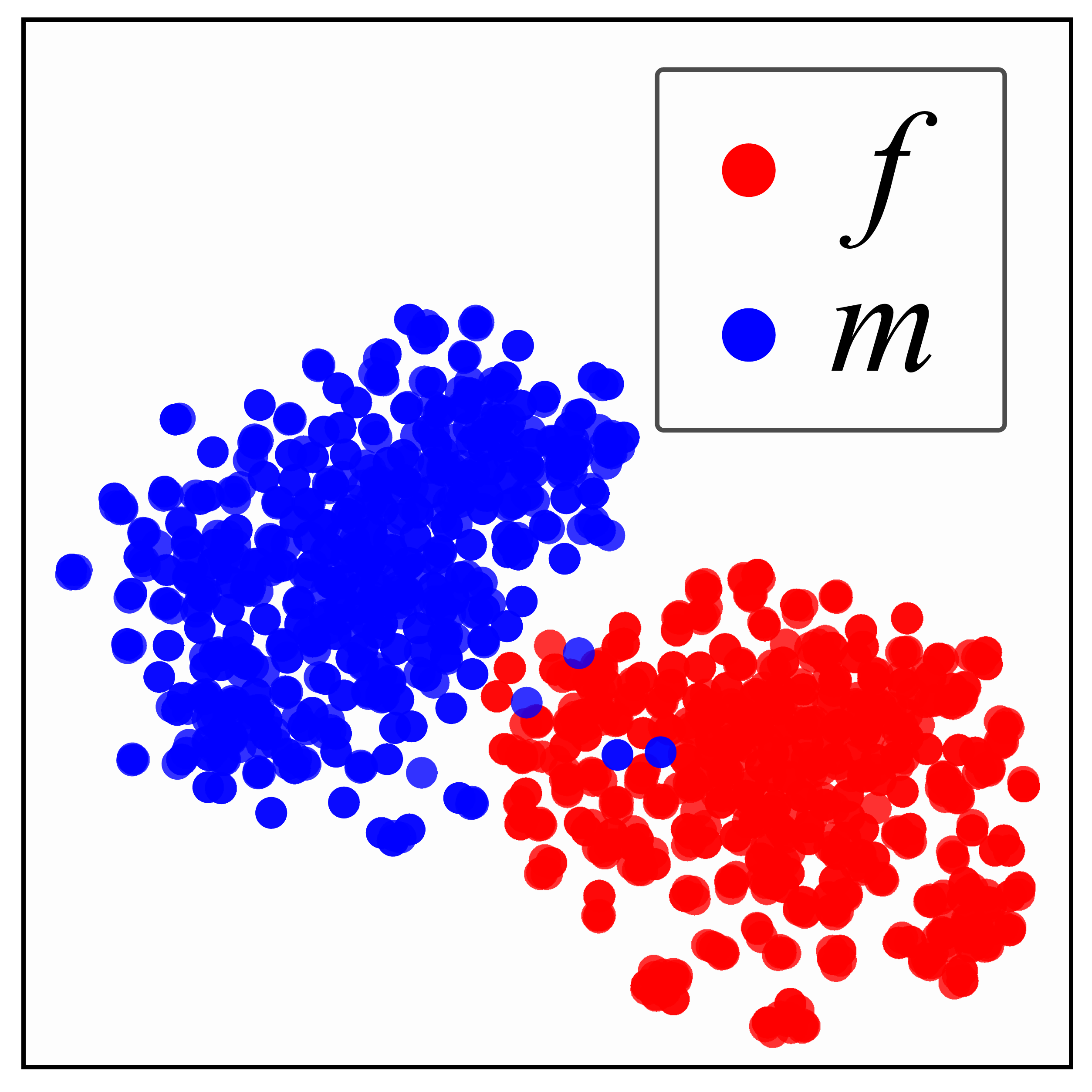}
\end{minipage}%
\begin{minipage}[b]{0.162\columnwidth}
    \centering
    \includegraphics[width=1\columnwidth, trim=0 10 0 0,
    clip]{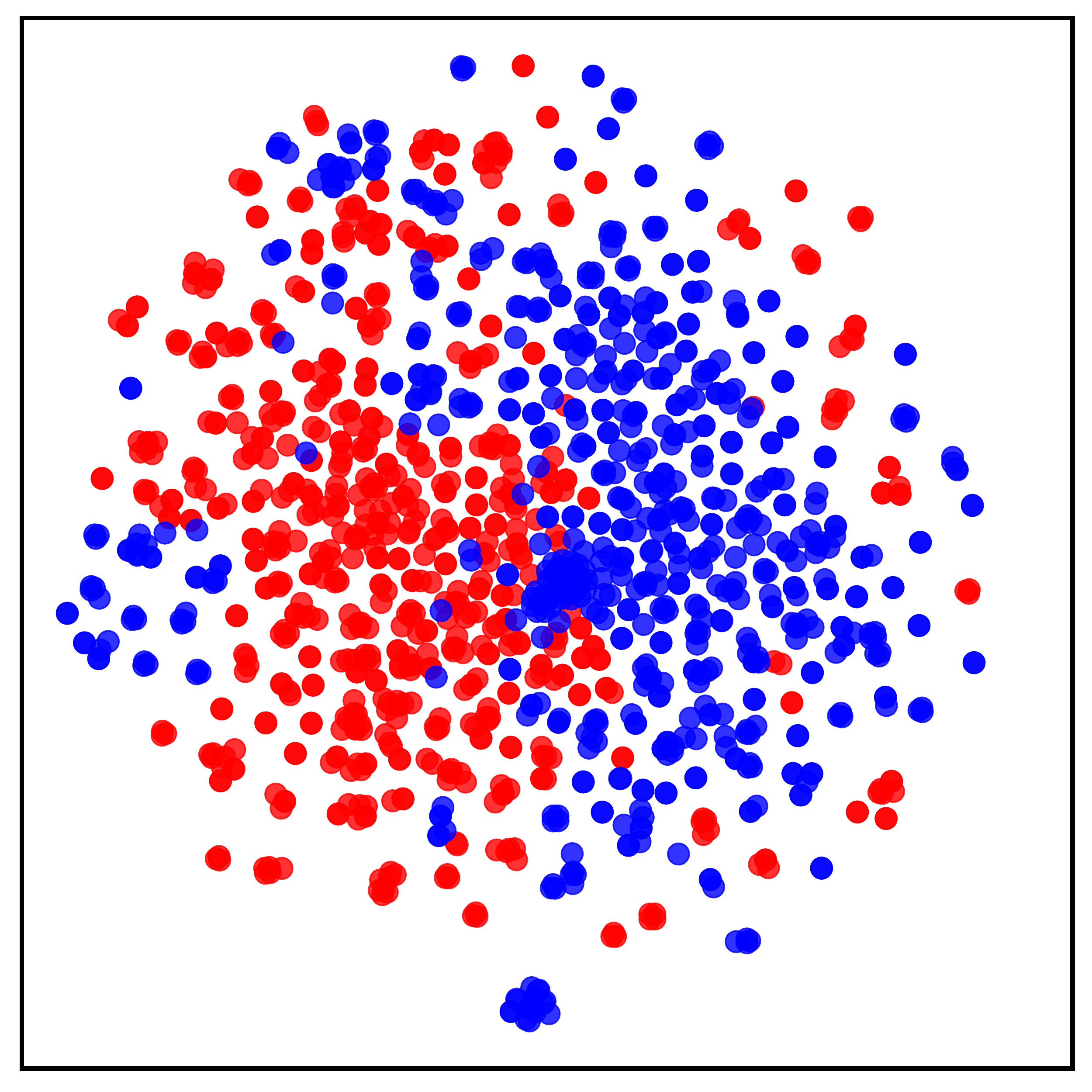}
\end{minipage}%
\begin{minipage}[b]{0.162\columnwidth}
    \centering
   \includegraphics[width=1\columnwidth, trim=0 10 0 0, clip]{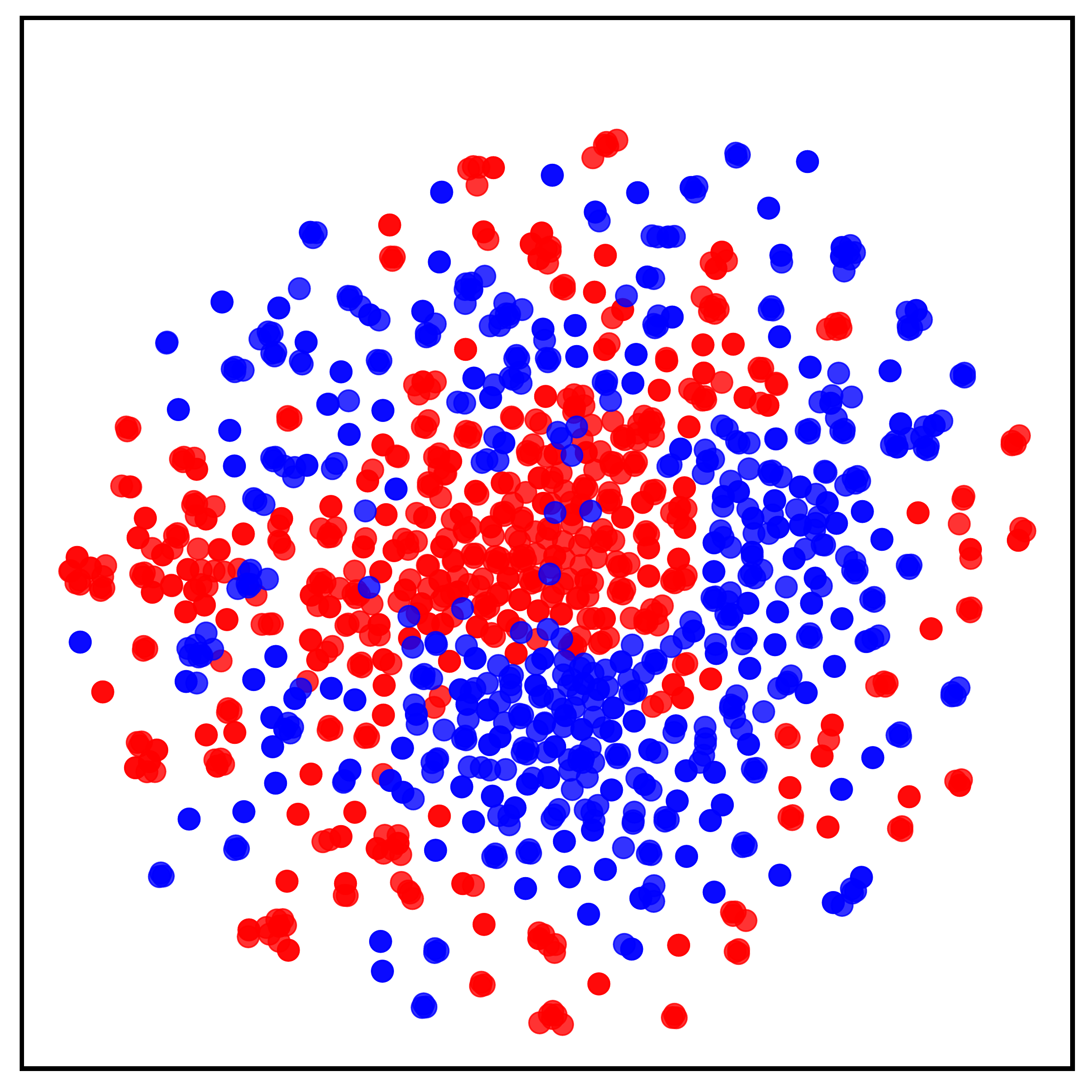}
\end{minipage}%
\begin{minipage}[b]{0.162\columnwidth}
    \centering
    \includegraphics[width=1\columnwidth, trim=0 10 0 0, clip]{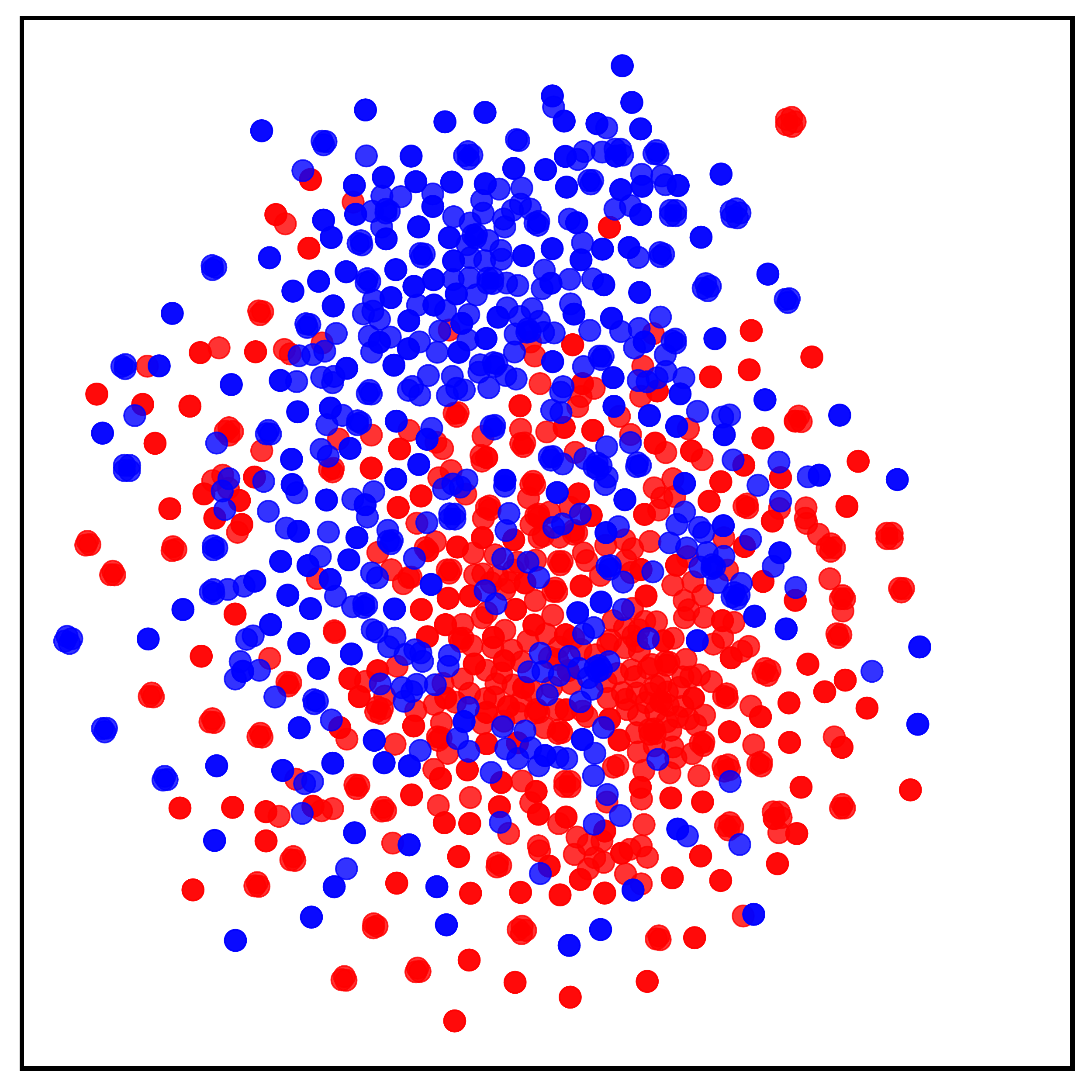}
\end{minipage}%
\begin{minipage}[b]{0.162\columnwidth}
    \centering
    \includegraphics[width=1\columnwidth,trim=0 10 0 0, clip]{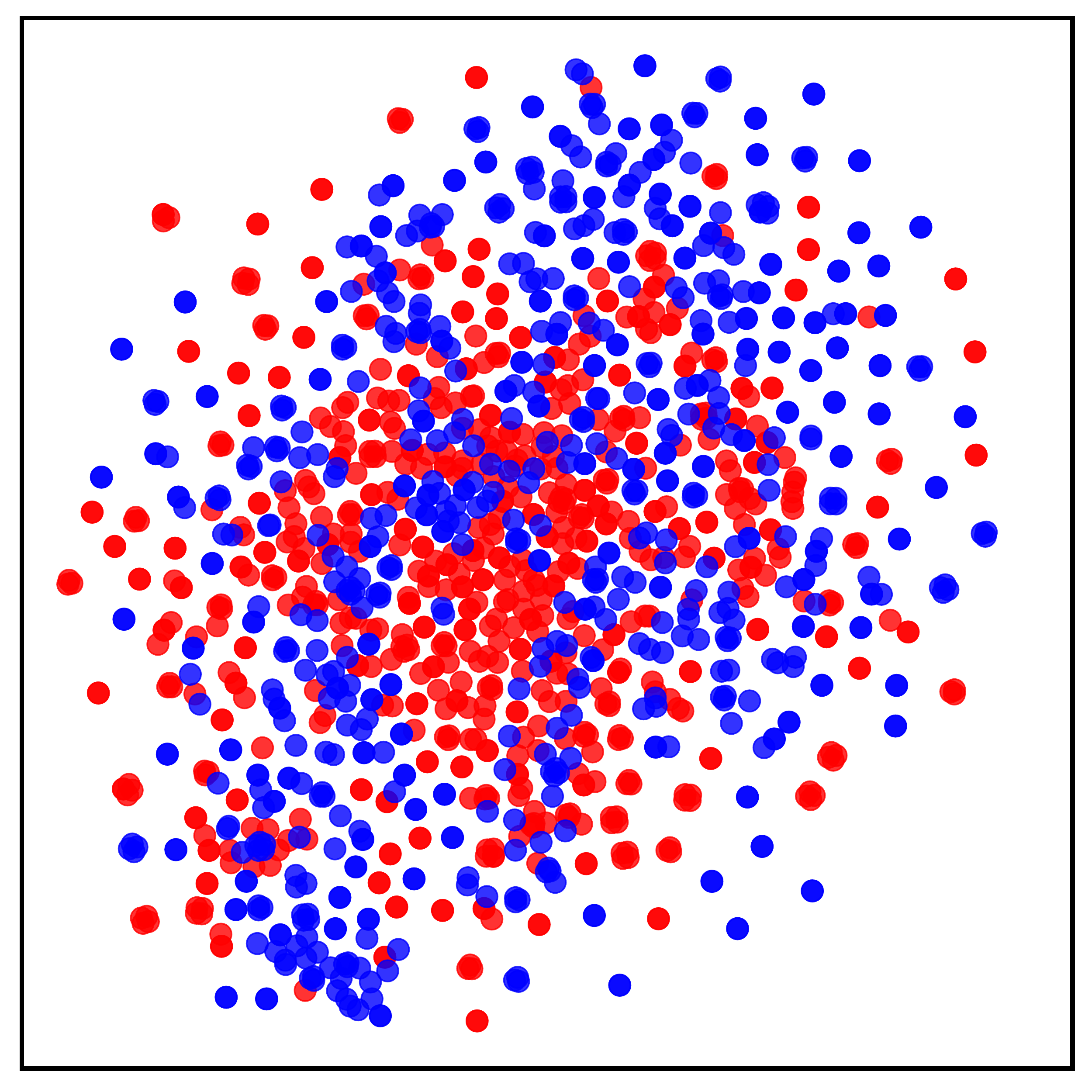}
\end{minipage}
\hspace{-0.2cm}
\begin{minipage}[b]{0.162\columnwidth}
    \centering
    \includegraphics[width=1\columnwidth,trim=0 10 0 0, clip]{
    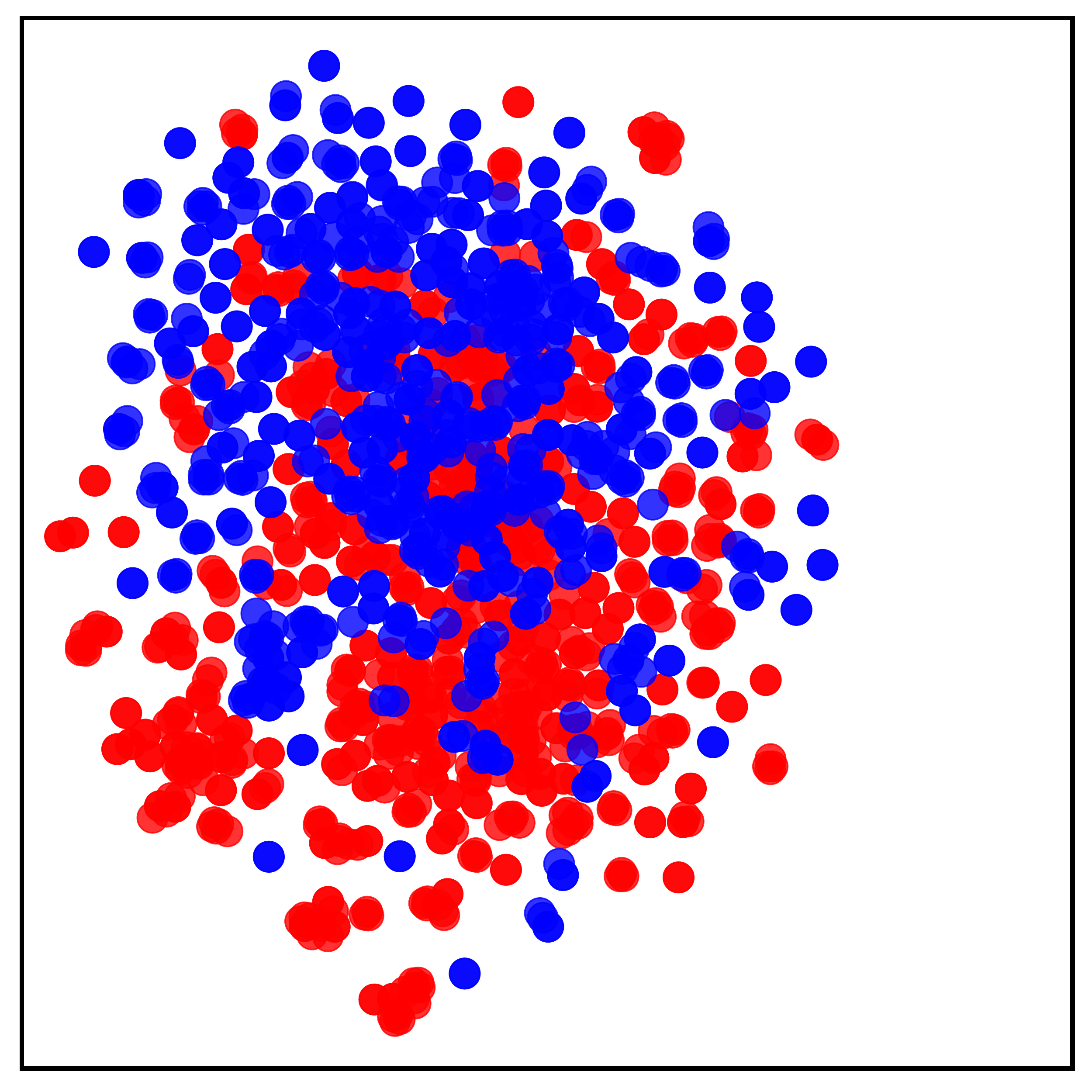}
\end{minipage}
\\
\begin{minipage}[b]{0.162\columnwidth}
    \centering
    \includegraphics[width=1\columnwidth,trim= 0 10 0 0, clip]{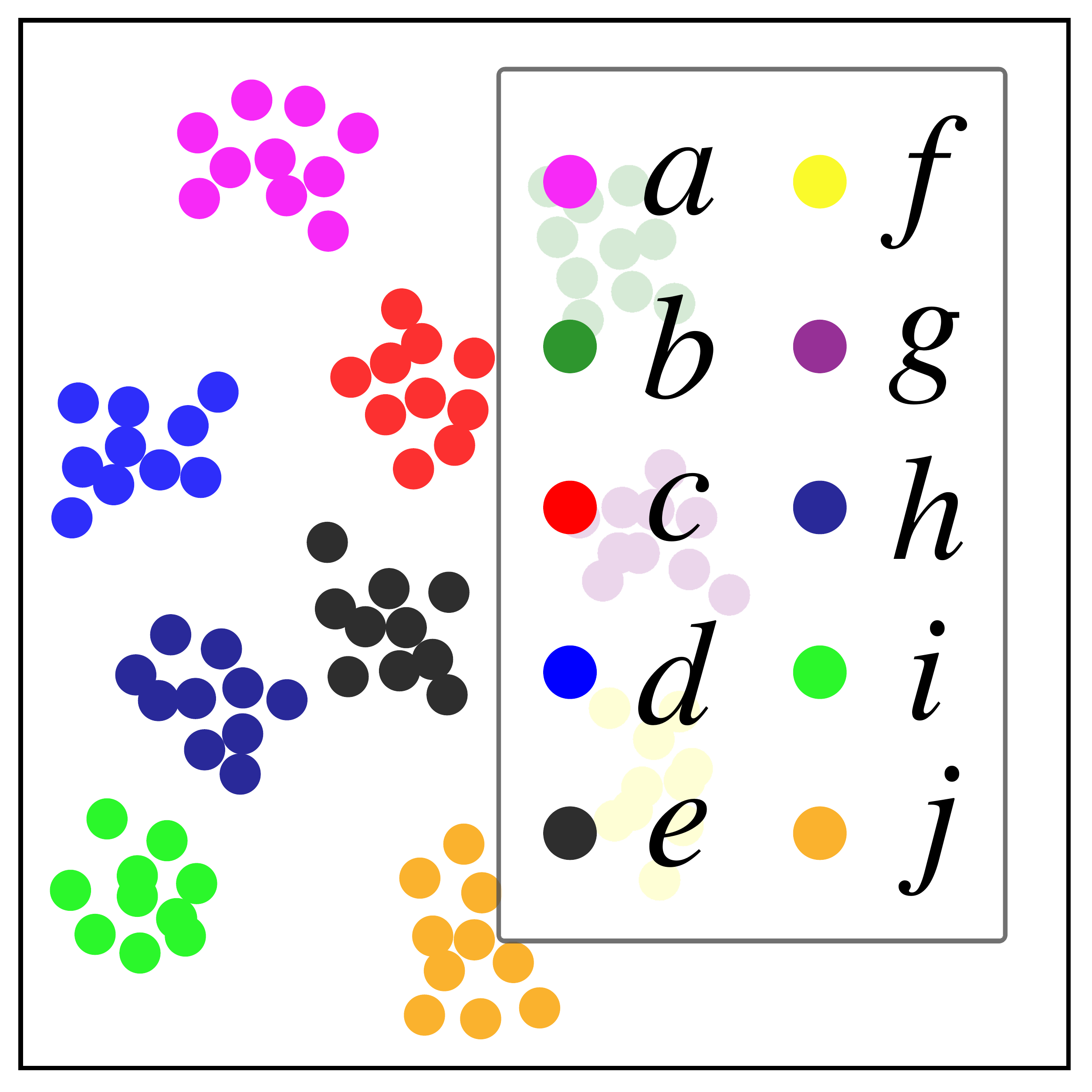}
\end{minipage}%
\begin{minipage}[b]{0.162\columnwidth}
    \centering
    \includegraphics[width=1\columnwidth,trim=0 10 0 0, clip]{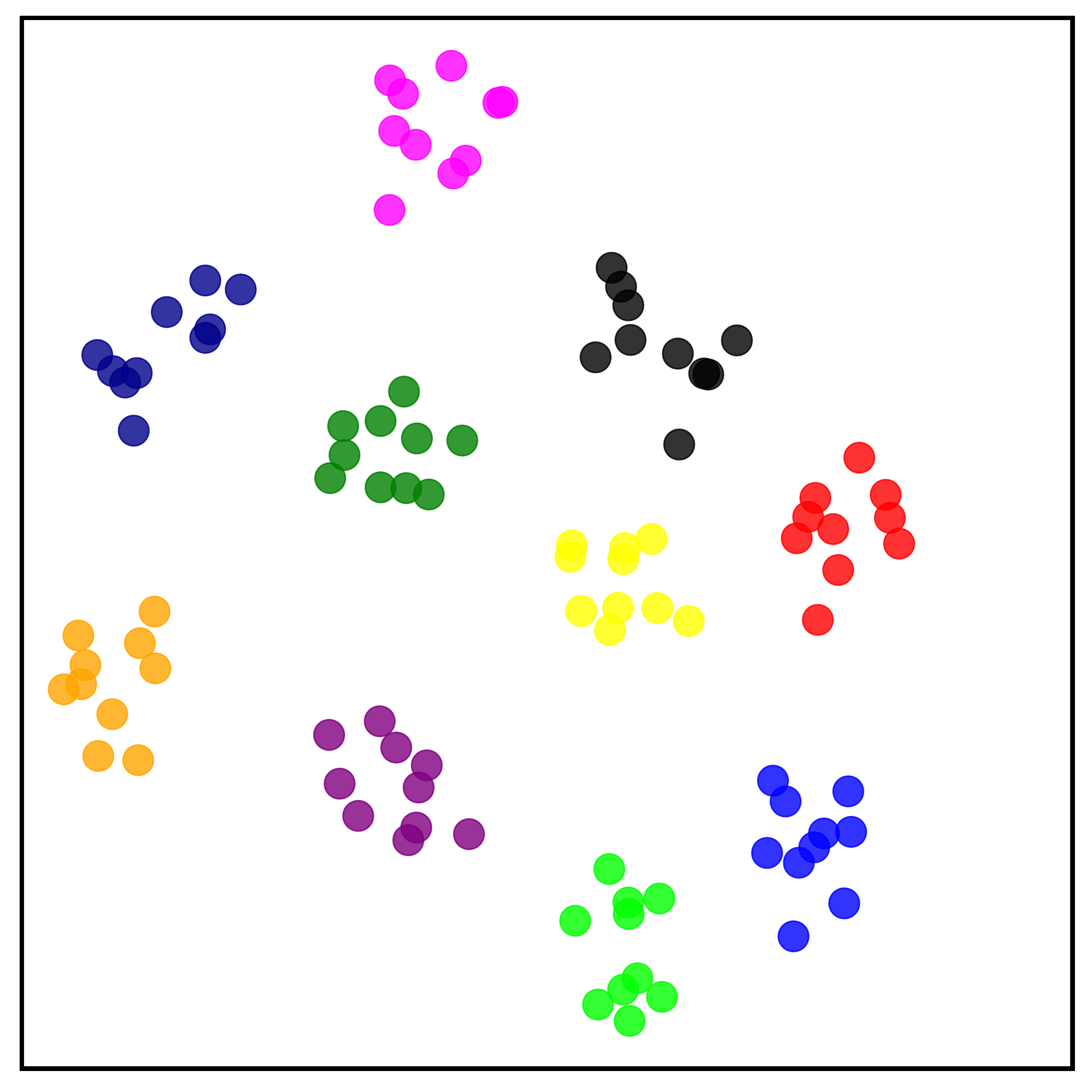}
\end{minipage}%
\begin{minipage}[b]{0.162\columnwidth}
    \centering
    \includegraphics[width=1\columnwidth,trim=0 10 0 0, clip]{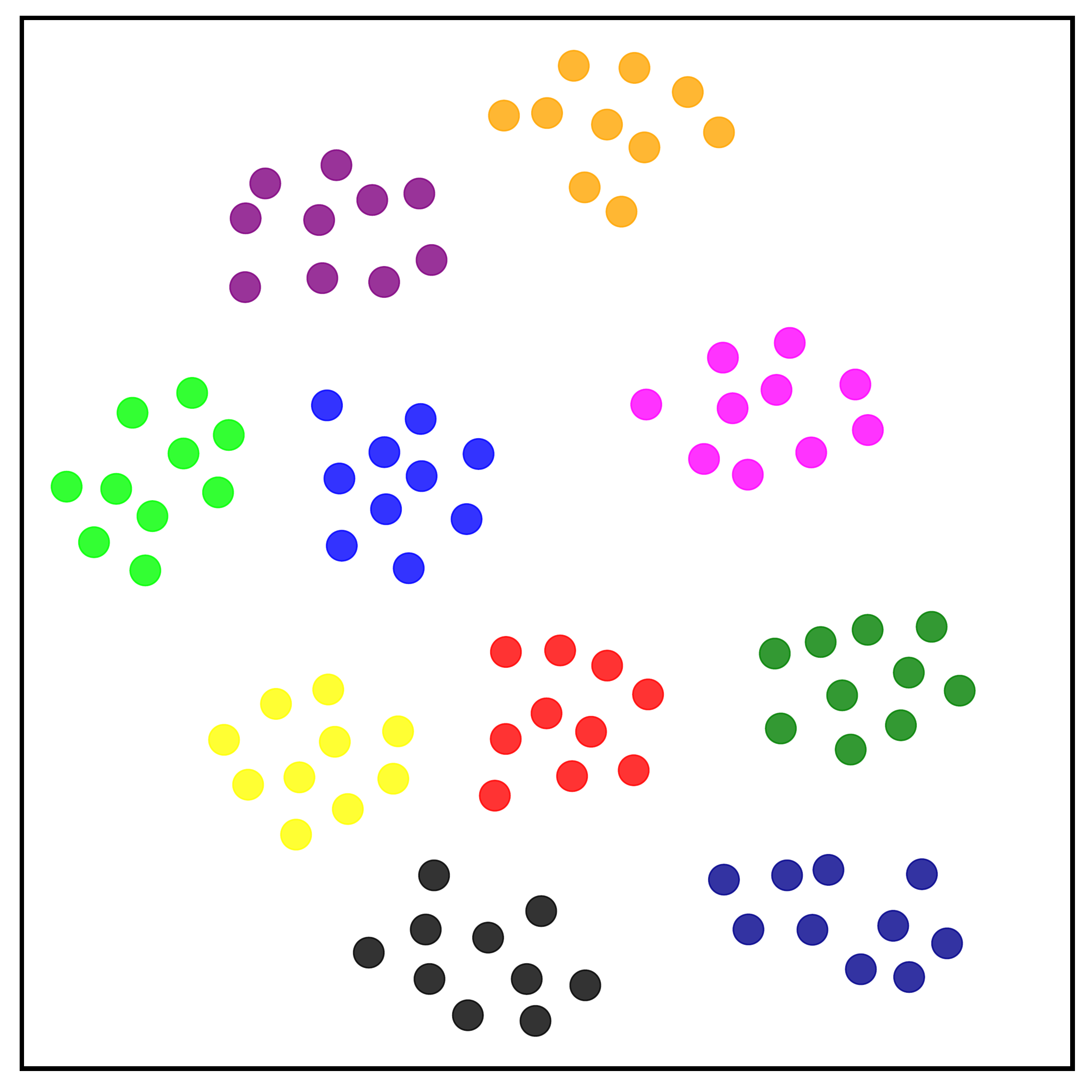}
\end{minipage}%
\begin{minipage}[b]{0.162\columnwidth}
    \centering
    \includegraphics[width=1\columnwidth,trim=0 10 0 0, clip]{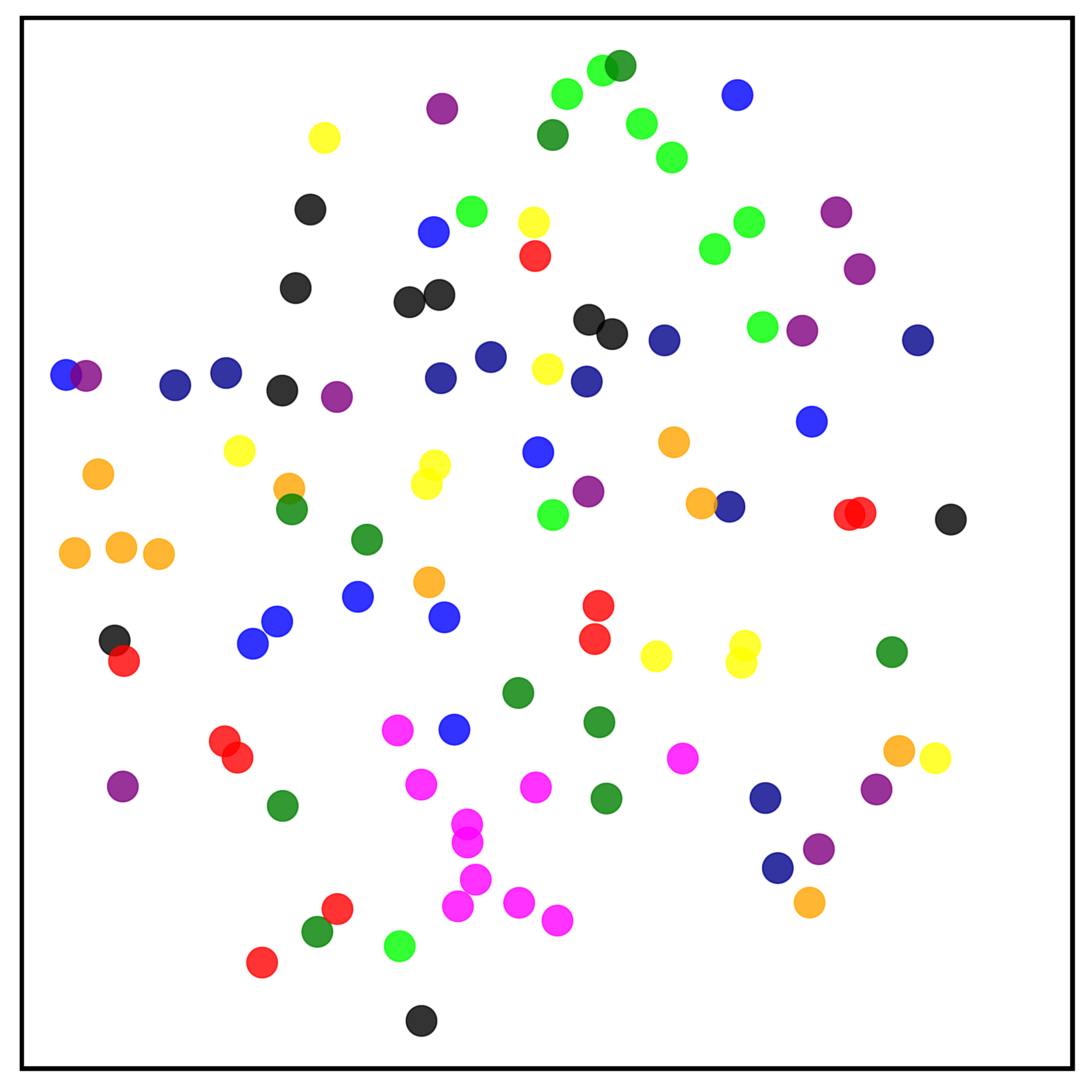}
\end{minipage}%
\begin{minipage}[b]{0.162\columnwidth}
    \centering
    \includegraphics[width=1\columnwidth,trim=0 10 0 0, clip]{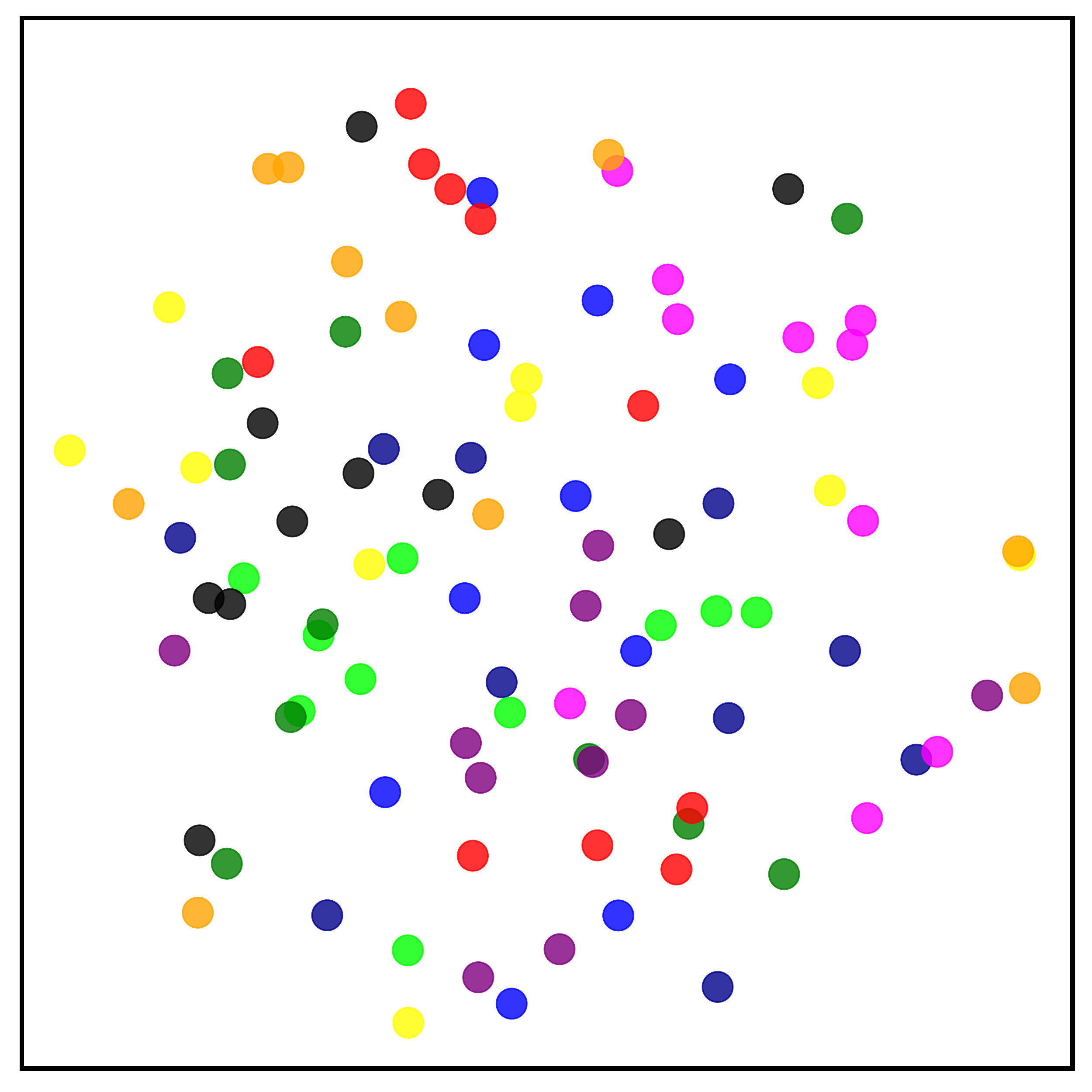}
\end{minipage}
\hspace{-0.2cm}
\begin{minipage}[b]{0.162\columnwidth}
    \centering
    \includegraphics[width=1\columnwidth,trim=0 10 0 0, clip]{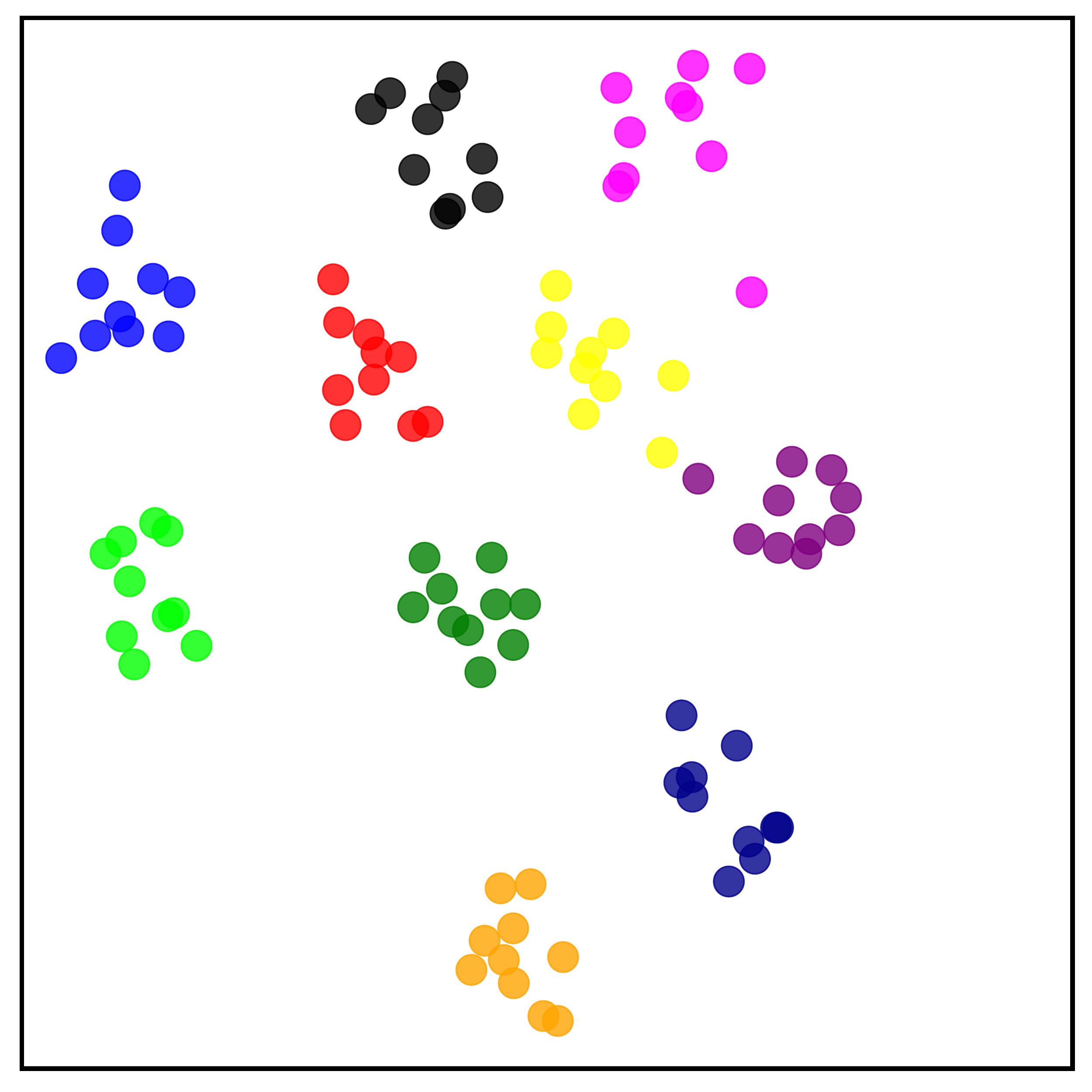}
\end{minipage}

\caption{$t$--SNE plots (in 2D) of gender and identity features. The gender plots in the top row are generated from $700$ randomly sampled LFW images of each gender ($f$--female, $m$--male). The subject--conditioned distributions in the bottom row are generated based on $10$ randomly selected images of the $10$ largest classes from LFW (marked (a) -- (j)). By comparing the distribution of the original and processed images the impact of the privacy enhancement can be observed.\label{fig: tSNE on LFW}\vspace{-3mm}}
\end{figure}

Next, we investigate the effect of privacy enhancement on the distribution of the features generated by the last fully connected layers of the \textit{(i)} ResNet--50 face recognition model and the \textit{(ii)} VGG16 gender classifier. Here, the same setup is used as with the baseline experiments discussed above. The goal of this series of experiments is to better understand what is happening at the representation level as a result of the privacy enhancement and to gain additional insight into the characteristics of the privacy models. 
To study the ResNet--50 (identity) features, images corresponding to the $10$ largest classes of LFW are selected and $t$-distributed Stochastic Neighbor Embedding ($t$--SNE) is used to visualize the feature distributions (in 2D) before and after privacy enhancement. For the gender features $700$ images from LFW are randomly sampled from the dataset for each gender. 

The $t$--SNE plots in Fig.~\ref{fig: tSNE on LFW} show that both gender (top row) as well as subjects/identities  (bottom row) are well separated with the original images. After privacy enhancement with $k$--AAP and FGSM  most of the separation between subjects is preserved. The gender distributions, on the other hand, are less clustered and now exhibit a multimodal distribution. Neveretheless, the overlap between the male and female data points is still limited, indicating that the two classes can still be distinguished using a suitable classification model. These observations support the results from the previous section where only small values in IL scores were observed with these two techniques, while the ROC curves from the gender--recognition experiments showed good separation, but with inverted labels. 

When looking at the FlowSAN models, a different behavior can be observed. Here, the identity features are not well separated, but for many data points exhibit (reasonably) correct pair--wise similarities. The gender distributions, on the other hand, overlap significantly, with a higher overlap for the FlowSAN--5 model, which is expected given the objective of the privacy enhancement. \textcolor{black}{PrivacyNet appears to combine the best of both worlds and ensures reasonable preservation of the identity clusters, while leading to considerable overlap in the feature distributions w.r.t. gender.} The observed distributions 
point to the shortcoming of the adversarial techniques $k$--AAP and FGSM, where the (gender) label swap could easily be identified using manual inspection of a few sample images. This is not the case with the FlowSAN and \textcolor{black}{PrivacyNet} models. 
\begin{figure}[t]
	\centering
	\includegraphics[width=0.95\columnwidth]{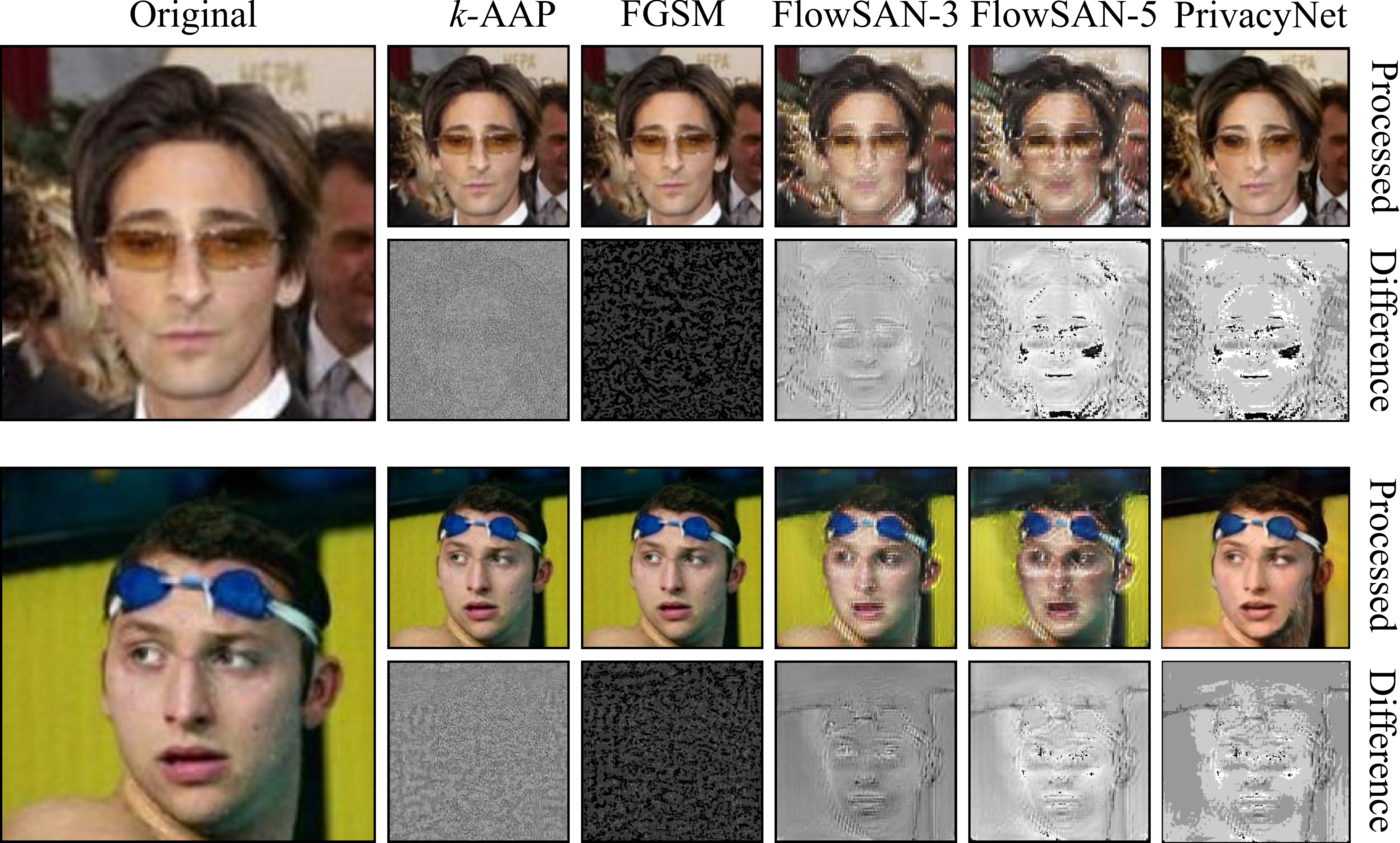}
	\caption{Illustration of the visual effect of the privacy models on two sample images from LFW. The top row next to each original image shows the privacy-enhanced images (with gender privacy), whereas the bottom row depicts the difference between the original and the modified images. The difference was scaled for visualization purposes (by $10\times$ for $k$--AAP, $40\times$ for FGSM and $1\times$ for the FlowSAN models) and then normalized to the valid image range. Note that $k$--AAP and FGSM result in imperceptible appearance changes, while the FlowSAN models introduce visible changes.}
	{
	}
	\label{fig:vis_impacts}
\end{figure}
\begin{table*}[!tb]
\centering
\caption{PrivacyProber variants implemented for the experimental evaluation. The models differ in terms of whether generative or domain--specific components and whether one or two processing steps are utilized for attribute recovery.\vspace{-1mm} 
}
\renewcommand{\arraystretch}{1.11}
\label{tab: Privacy_probers}
\resizebox{\textwidth}{!}{%
\begin{tabular}{lccccccccr} \hline\hline
\multirow{ 2}{*}{PrivacyProber} &\multirow{ 2}{*}{$\chi$} &&  \multicolumn{3}{c}{Generative components} && \multicolumn{1}{c}{Domain components} && \multirow{ 2}{*}{Characteristic} \\ \cline{4-6} \cline{8-8}
         &&& Denoising & Inpainting &    \textcolor{black}{Auto--encoder}  && Background removal  &&  \\\hline
PP--D & $\chi_d$  && \cmark & \gmark & \gmark && \gmark  && Generative, one--stage (GOS)\\
PP--I& $\chi_{in}$   && \gmark & \cmark & \gmark && \gmark    &&  Generative, one--stage (GOS)\\
\textcolor{black}{PP--A}& $\chi_{a}$    && \gmark & \gmark & \cmark && \gmark   &&  Generative, one--stage (GOS)\\
PP--B& $\chi_{br}$    && \gmark & \gmark & \gmark && \cmark    &&  Domain--specific, one--stage (DOS)\\\hline
PP--DI& $\chi_{d}\circ\chi_{in}$  && \cmark & \cmark & \gmark && \gmark  &  & Generative, two--stage (GTS)\\
\textcolor{black}{PP--DA}& $\chi_{d}\circ\chi_{a}$  && \cmark & \gmark & \cmark && \gmark  &  & Generative, two--stage (GTS)\\ \hline

PP--DB& $\chi_{d}\circ\chi_{br}$  && \cmark & \gmark & \gmark &&  \cmark &  & Generative+domain--specific, two--stage (GDT)\\
PP--IB& $\chi_{in}\circ\chi_{br}$  && \gmark & \cmark & \gmark && \cmark  &  & Generative+domain--specific, two--stage (GDT)\\
\textcolor{black}{PP--AB}& $\chi_{a}\circ\chi_{b}$  && \gmark & \gmark & \cmark &&  \cmark &  & Generative+domain--specific, two--stage (GDT)\\
\hline \hline
\multicolumn{7}{l}{\scriptsize $\circ$ denotes a function composition operator.}
\end{tabular}}\vspace{-2mm}
\end{table*}

%



\subsubsection{Qualitative assessment}\label{Sec: Qualitative}

The impact of the privacy models on the visual appearance of a couple of sample images from the LFW dataset is shown in Fig.~\ref{fig:vis_impacts}. As can be seen, $k$--AAP and FGSM generate privacy--enhanced images that are almost identical to the original images, \textcolor{black}{the FlowSAN models introduce bigger and visually noticeable changes, whereas PrivacyNet processed images are somewhere in between}. The bottom row in each of the two examples presents a visualization of the changes introduced by the privacy models. $k$--AAP and FGSM add a relatively uniformly distributed noise pattern to the input images, while the  FlowSAN and \textcolor{black}{PrivacyNet} models introduce a structured pattern focused predominantly on the facial area and less so on the background. This observation can be attributed to the design of the privacy models, where FlowSAN and \textcolor{black}{PrivacyNet} are designed specifically for facial images, while the other two models are applicable to arbitrary images and classification problems and, hence, are not \textcolor{black}{explicitly targeting} specific visual categories, i.e., faces. From a qualitative perspective, the adversarial models have an edge over the FlowSAN and \textcolor{black}{PrivacyNet} models, but as suggested in the previous sections this edge comes at the expense of robustness to unseen classification models and the simpler privacy mechanism that aim to induce an incorrect prediction with a chosen classifier. 

\input{sestavljena_slika}

\subsection{Robustness to Recovery Attempts}

The main contribution of this study is a rigorous evaluation of existing (soft--biometric) privacy models with respect to their performance beyond zero-effort recognition experiments. 
The next series of experiments, therefore, explores the robustness of the evaluated models against attribute recovery attempts facilitated by our PrivacyProber.

\subsubsection{One--Stage and Two--Stage Attribute Recovery}\label{Sec: oneStageTwoSTage}

Multiple versions of PrivacyProber are implemented for the robustness experiments using either \textit{one--stage} or \textit{two--stage} attribute recovery. The simpler one--stage implementations consist of a single (generative or domain--specific)  transformation, whereas the more complex two--stage implementations use sequential combinations of generative and domain--specific operations. For the one--stage implementations inpainting, \textcolor{black}{image reconstruction}, denoising and background removal are considered as stand--alone recovery options, whereas multiple different combination of image transformations are incorporated into the two--stage implementations. A high--level overview of the \textcolor{black}{considered} PrivacyProber variants is presented in Table~\ref{tab: Privacy_probers}. Note that these combinations are not exhaustive with respect to all possible attribute recovery options discussed in Section~\ref{Sec: PrivacyProber}. However, they do provide a representative cross--section of the existing options for the experimental evaluation.

For the implementation of the recovery techniques state--of--the--art backbone models are selected. Specifically, the GMCNN model\footnote{GMCNN: \url{https://github.com/shepnerd/inpainting_gmcnn}}~\cite{wang2018image} trained on  CelebA-HQ~\cite{karras2018progressive} is \textcolor{black}{utilized for the proposed inpainting scheme, the auto-encoder from~\cite{folz2020adversarial} trained on a selection of real-world images of objects is selected for the image reconstruction procedure}, the WDnCNN\footnote{WDnCNN: \url{https://github.com/lixiaopeng123456/WDnCNN}}~\cite{zhao2019enhancement} model trained on the Waterloo Exploration Database~\cite{ma2017waterloo}, the Berkeley segmentation dataset~\cite{roth2005fields} and part of ImageNet~\cite{deng2009imagenet} is chosen for the denoising operation and the DeepLabv3\footnote{DeepLabv3: \url{https://github.com/tensorflow/models/tree/master/research/deeplab}} face parser~\cite{liang2017rethinking} trained on CelebAMask-HQ~\cite{CelebAMask-HQ} is selected for the background removal process. The backbone techniques ensure state--of--the--art performance for each task and come with publicly available implementations and pretrained weights. It is also important to note that the data used to train the backbone techniques does not overlap (in terms of images or subjects) with any of the test datasets used.  

The impact of different PrivacyProbers on the visual appearance of a sample face image is presented in Fig.~\ref{fig:examples}. Here, the numbers below the images correspond to probabilities that the subject on the image is male and the color coded frames indicate whether a gender classifier with a decision threshold at $0.5$ correctly determines the subject's gender (green) or not (red). As can be seen, \textcolor{black}{all tested privacy models significantly reduce the correct-gender-prediction probability compared to the original image, whereas the majority of} PrivacyProbers successfully recover a considerable amount of gender information regardless of the privacy model used. The one--stage implementations based on inpaiting (PP-I), \textcolor{black}{image reconstruction (PP-A) and} denoising (PP-D) also retain all of the visible semantic content. In terms of image quality, the inpainting scheme results in a slight loss of image contrast \textcolor{black}{and the use of the auto-encoder-based leads to somewhat blurrier outputs}. The denoising--based implementation, on the other hand, additionally improves on the perceived quality of the facial images. Background subtraction is the only one--stage approach that removes part of the information and drastically changes the image content. Such images may be of limited use for certain applications, but can still facilitate automatic processing and analysis. The two--stage versions of PrivacyProber in general inherit the properties of the single--stage components and depending on the operations used may again retain all of the semantic content, e.g., PP-DI, \textcolor{black}{PP-DA}, or remove part of it due to background subtraction, e.g., PP-DB. 

\subsubsection{Robustness Analysis}
\begin{table*}[!tb]
\centering
\caption{Attribute recovery robustness (ARR - higher scores indicate better robustness) scores generated for the evaluated privacy models with different PrivacyProber variants. Results are presented in terms of ($\mu\pm\sigma$) ARR scores computed over 4 experimental splits and grouped by PrivacyProber type: GOS -- Generative and One Stage; GDT -- Generative and Two Stage; GDT -- Generative and Domain--specific and Two Stage. The lowest robustness against attribute recovery for each privacy model on each dataset (in rows) is marked red, the highest is marked blue.}
\renewcommand{\arraystretch}{1.15}
\label{tab: robustness table}
\resizebox{\textwidth}{!}{%
\begin{tabular}{lrrccccccccccc} \hline\hline
\multirow{ 2}{*}{Privacy Model} & \multirow{ 2}{*}{Dataset}  &&  \multicolumn{4}{c}{ARR -- GOS models} && \multicolumn{2}{c}{ARR -- GTS models} && \multicolumn{3}{c}{ARR -- GDT models}\\ \cline{4-7} \cline{9-10} \cline{12-14}
         &&& PP--D & PP--I & PP--A & PP-B && PP--DI & PP-DA && PP--DB & PP--IB & PP--AB  \\\hline
\multirow{ 3}{*}{$k$--AAP}  & LFW  &&   
$0.251\pm 0.076$ &  $0.149\pm0.063$ & $0.034\pm 0.017$ & \color{blue}{$0.410\pm0.030$} && $0.091\pm0.004$ & \color{red}{$0.026\pm 0.012$} && $0.163\pm0.062$ &  $0.108\pm0.045$ & $0.031\pm 0.012$ \\
                            & MUCT && 
\color{blue}{$0.089\pm 0.043$} & $0.078\pm 0.028$ & $0.028\pm 0.013$ & $0.082\pm 0.014$ &&	$0.031\pm 0.010$ &	\color{red}{$0.016\pm 0.007$} && $0.072\pm 0.040$	& $0.061\pm 0.023$ & $0.027\pm 0.015$
                            
                            \\
                            & Adience  && 
\color{red}{$0.024\pm 0.022$} &	$0.392\pm 0.024$	& $0.358\pm 0.046$ &	$0.079\pm 0.018$ &&	$0.332\pm 0.026$& 	\color{blue}{$0.399\pm 0.015$} &&	$0.096\pm 0.027$ & $0.304\pm 0.038$ & $0.370\pm 0.029$          \\
                            \arrayrulecolor{gray}\hline
\multirow{ 3}{*}{FGSM} & LFW  && 
$0.084\pm 0.022$ & $0.040\pm 0.016$ & $0.013\pm 0.007$ & \color{blue}{$0.353\pm 0.028$}	&& $0.020\pm 0.011$ & \color{red}{$0.007\pm 0.004$} && $0.053\pm 0.016$ & $0.032\pm 0.014$ & $0.016\pm 0.006$ \\ 
                            & MUCT && 
$0.016\pm 0.016$ &	$0.014\pm 0.007$	& $0.011\pm 0.009$ &	\color{blue}{$0.162\pm 0.035$} &&	\color{red}{$0.003\pm 0.002$} & $0.006\pm 0.004$ && $0.014\pm 0.013$ & $0.011\pm 0.005$ & $0.007\pm 0.007$   \\
                            & Adience  &&  
\color{blue}{$0.379\pm 0.029$} & $0.147\pm 0.003$ & $0.192\pm 0.014$ & \color{red}{$0.012\pm 0.007$} && $0.123\pm 0.005$ & $0.138\pm 0.011$ && $0.306\pm 0.023$ & $0.160\pm 0.010$ & $0.193\pm 0.019$ \\
\arrayrulecolor{gray}\hline

\multirow{3}{*}{FlowSAN--3} & LFW  && 
$0.433\pm 0.030$ &	\color{red}{$0.425\pm 0.016$} &	$0.542\pm 0.028$ & $0.430\pm 0.041$ && $0.443\pm 0.028$ &	$0.551\pm 0.032$ && $0.445\pm 0.036$ & $0.552\pm 0.022$ & \color{blue}{$0.631\pm 0.010$} \\
                            & MUCT && 
$0.390\pm 0.084$ &	 $0.324\pm 0.155$ &	$0.424\pm 0.130$	& $0.270\pm 0.055$ && $0.338\pm 0.114$ & $0.405\pm 0.129$ && $0.319\pm 0.079$ & \color{red}{$0.248\pm 0.100$} & \color{blue}{$0.731\pm 0.074$} \\
                            & Adience  &&  
$0.554\pm 0.026$ & $0.422\pm 0.036$ & $0.490\pm 0.037$ & \color{red}{$0.419\pm 0.040$} && $0.420\pm 0.035$ & $0.480\pm 0.039$ && $0.550\pm 0.032$ & $0.629\pm 0.064$ & \color{blue}{$0.641\pm 0.040$}
\\
\arrayrulecolor{gray}\hline
\multirow{ 3}{*}{FlowSAN--5} & LFW  && 
$0.710\pm 0.057$ & \color{red}{$0.578\pm 0.048$} & $0.698\pm 0.035$ & $0.674\pm 0.066$ &&	$0.590\pm 0.043$	& $0.708\pm 0.031$	&& $0.676\pm 0.059$ & $0.661\pm 0.042$ & \color{blue}{$0.741\pm 0.032$}

\\
                            & MUCT &&
$0.797\pm 0.055$ & $0.730\pm 0.109$ & \color{blue}{$0.803\pm 0.087$} & $0.723\pm 0.063$	&& $0.703\pm 0.103$ & $0.773\pm 0.088$ &&	$0.658\pm 0.061$ & $0.652\pm 0.133$ & \color{red}{$0.605\pm 0.144$}

                            \\
                            & Adience && 
$0.750\pm 0.057$ & $0.570\pm 0.054$ & $0.666\pm 0.030$ & $0.726\pm 0.021$ && \color{red}{$0.565\pm 0.050$} & $0.660\pm 0.031$ && $0.712\pm 0.040$ & \color{blue}{$0.763\pm 0.048$} & $0.681\pm 0.053$

                            \\
\arrayrulecolor{gray}\hline
\multirow{ 3}{*}{PrivacyNet} & LFW  && 
{$0.845\pm 0.075$} & {$0.494\pm 0.097$} & $0.520\pm 0.117$ & $0.876\pm 0.045$ &&	$0.721\pm 0.040$	& $0.786\pm 0.075$	&& \color{red}{$0.399\pm 0.108$} & $0.505\pm 0.099$ & \color{blue}{$0.857\pm 0.064$}

\\
                            & MUCT &&
{$0.016\pm 0.009$} & $0.026\pm 0.014$ & $0.022\pm 0.019$ & \color{red}{$0.011\pm 0.009$}	&& \color{blue}{$0.670\pm 0.105$} & $0.553\pm 0.138$&&	$0.012\pm 0.010$ & $0.033\pm 0.012$ & $0.656\pm 0.064$

                            \\
                            & Adience && 
\color{red}{$0.562 \pm 0.051$} & $0.573\pm 0.073$ & $0.610\pm 0.075$ & $0.579\pm 0.039$ && \color{blue}{$0.807\pm 0.047$} & $0.754\pm 0.027$ && $0.572\pm 0.036$ & $0.672\pm 0.063$ & $0.785\pm 0.042$
\\
\arrayrulecolor{black}
\hline \hline
\end{tabular}
}
\end{table*}
\begin{figure*}[!th]
\vspace{-2mm}
\begin{minipage}{0.33\textwidth}
\centering
\vspace{1mm}
\text{\footnotesize \ \ \ \hspace{3mm}LFW} 
\end{minipage}
\hfill
\begin{minipage}{0.33\textwidth}
\centering
\vspace{1mm}
\text{\footnotesize \ \ \ \hspace{4mm}MUCT}
\end{minipage}
\hfill
\begin{minipage}{0.33\textwidth}
\centering
\vspace{1mm}
\text{\footnotesize \ \ \ \hspace{5mm}Adience}
\end{minipage}\\ \vspace{-3mm}

\begin{subfloat}
    \centering
    \includegraphics[width=0.33\textwidth,
     trim=0mm 0mm 0mm 0mm,
     clip]{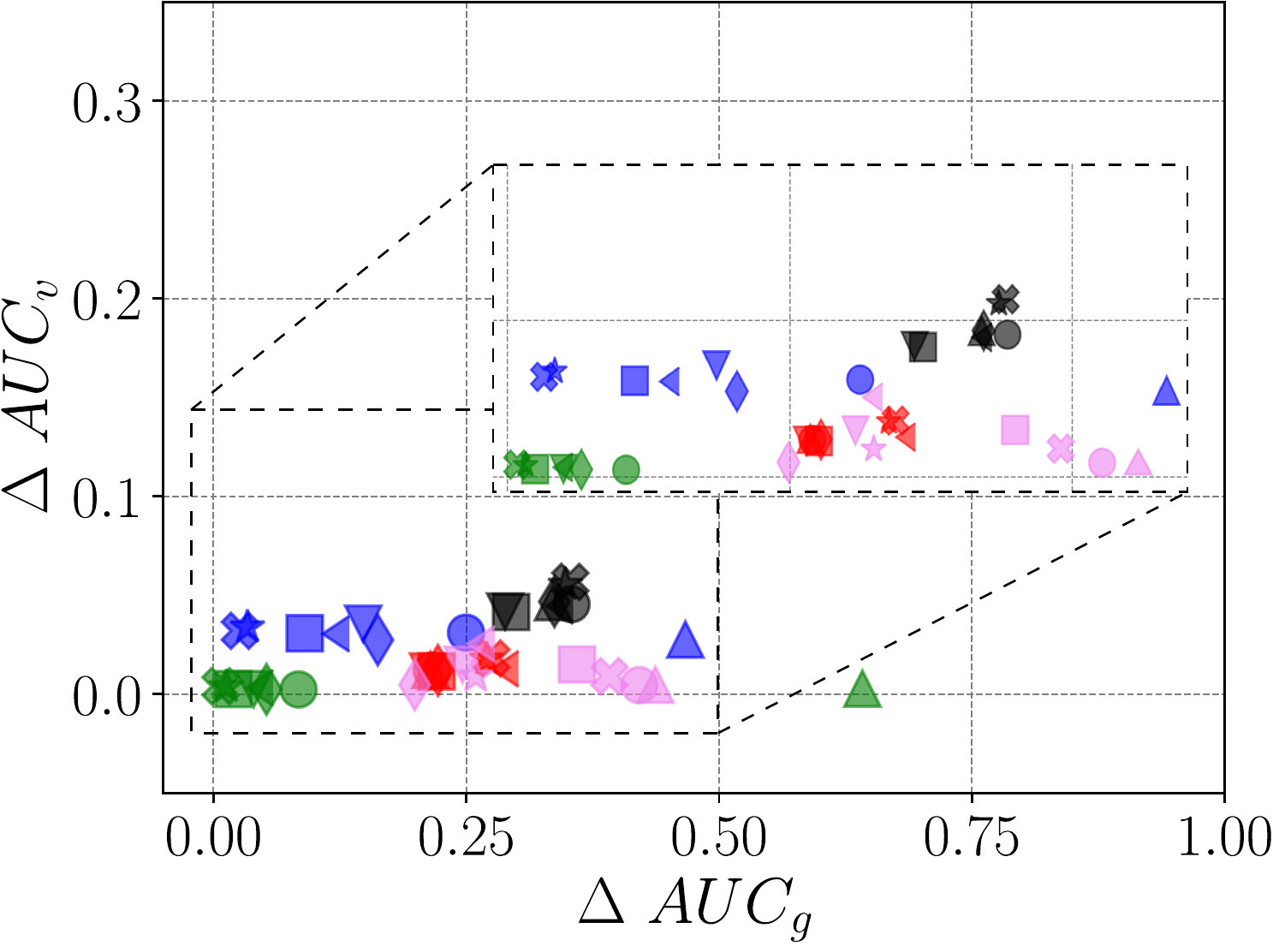}
     %
\end{subfloat}
\begin{subfloat}
    \centering
    \includegraphics[width=0.33\textwidth,
     trim=0mm 0mm 0mm 0mm,
     clip]{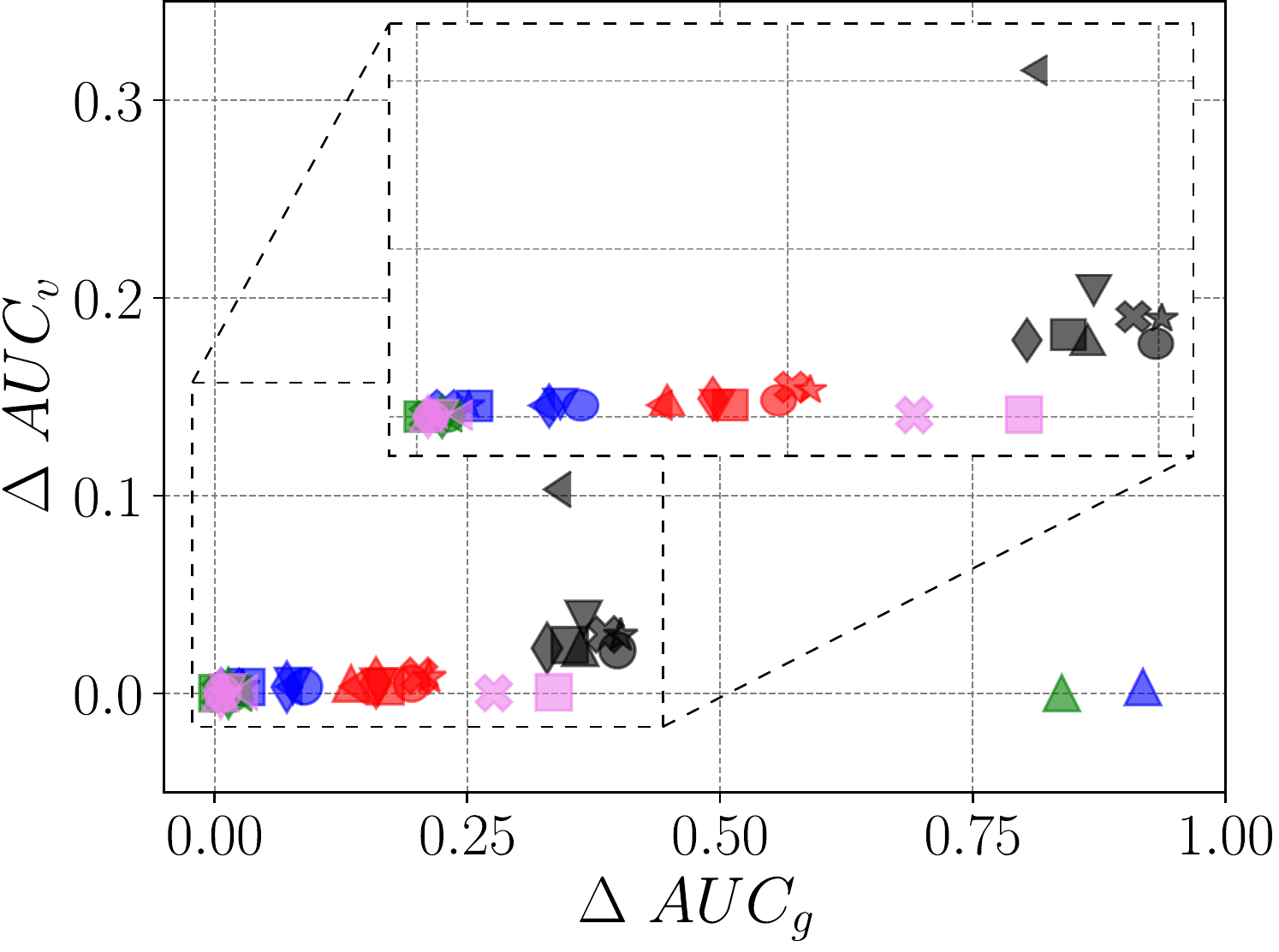}
\end{subfloat}
\begin{subfloat}
    \centering
    \includegraphics[width=0.33\textwidth,
     trim=0mm 0mm 0mm 0mm,
     clip]{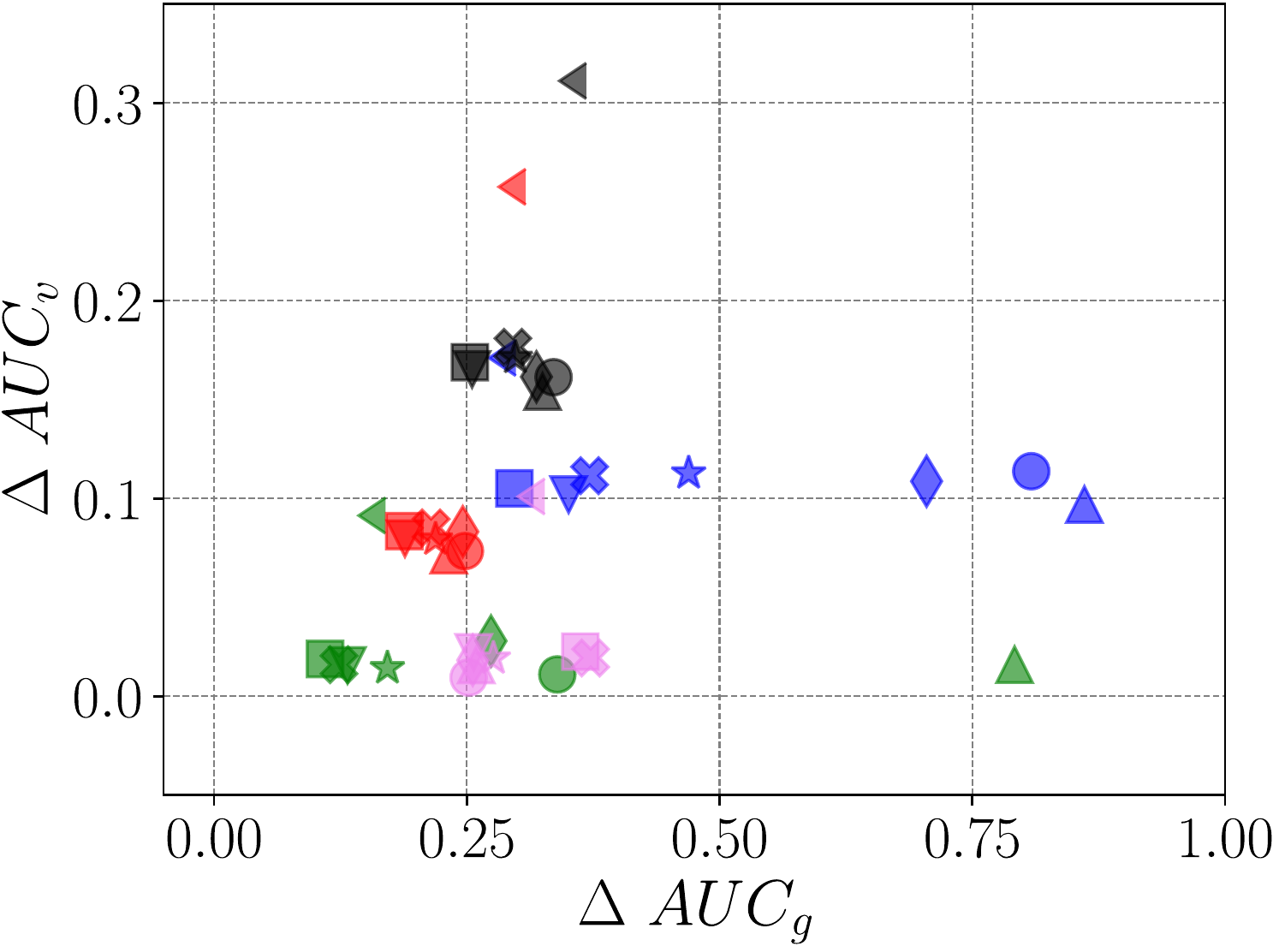}
     
\end{subfloat}
\arrayrulecolor{gray}
\color{gray}\noindent\makebox[\linewidth]{\rule{\textwidth}{0.05pt}}\vspace{-3.5mm}\\
\color{gray}\noindent\makebox[\linewidth]{\rule{\textwidth}{0.05pt}}
\begin{minipage}{1\textwidth}
\centering
\includegraphics[width=0.99\textwidth,
     trim=0mm 0mm 0mm 0mm,
     clip]{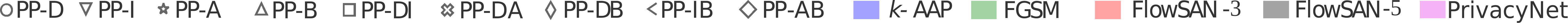} 
\end{minipage}\vspace{-2mm}
\color{gray}\noindent\makebox[\linewidth]{\rule{\textwidth}{0.05pt}}\vspace{-3.4mm}\\
\color{gray}\noindent\makebox[\linewidth]{\rule{\textwidth}{0.05pt}}
\arrayrulecolor{black}

\caption{Impact of application of different PrivacyProbers on the verification and gender--recognition performance. The $x$--axis shows the  differences in verification performance observed with the original and privacy--enhanced images that were subjected to the proposed PrivacyProber, i.e., $\Delta AUC_v = AUC_{vo}-AUC_{vr}$. The $y$--axis shows a similar difference for gender recognition, i.e., $\Delta AUC_g = AUC_{go}-AUC_{gr}$. Results are presented across all tested privacy models, PrivacyProbers and experimental datasets. The figure is best viewed in color.\label{fig: coors_plots}\vspace{-4.5mm}}
\end{figure*}

To assess the robustness of the privacy models to attribute recovery attempts, privacy--enhanced images, $I_{pr}$, from all three experimental datasets are first subjected to the implemented PrivacyProbers and then analyzed through verification and gender--recognition experiments. For the evaluation, the same gender classifier that steered the privacy enhancement on each dataset is again used to score gender--recognition performance. Note that the privacy models exhibited the strongest performance against this classifier (see Figs.~\ref{fig: ROC curves} and~\ref{fig: scalar scores}), which makes attribute recovery particularly challenging. For the verification experiments, the ResNet--50 face recognition model is utilized.

\textbf{1) ARR analysis.} Table~\ref{tab: robustness table} presents the \textit{attribute recovery robustness (ARR)} scores (from Eq.~\eqref{eq: recovery_rate}) generated from the attribute--recovered images. 
For each privacy model (and for each dataset), ARR scores for the PrivacyProber that resulted in the lowest robustness are colored red and the scores that correspond to the highest robustness are colored blue. Several interesting observations can be made from the presented results: (i) First, considerably more information can be recovered from the adversarial techniques, $k$--AAP and FGSM \textcolor{black}{on average, than from the FlowSAN and PrivacyNet models}. For these techniques at least one of the PrivacyProbers results in a ARR score in the $0.1$ range, which suggests that a significant amount of the gender information can be restored. Among the FlowSAN models, FlowSAN--3 is less robust to attribute recovery attempts than FlowSAN--5, but still significantly more so than $k$--AAP or FGSM. As expected, the most robust among all models is FlowSAN--5, where the weakest ARR scores across all PrivacyProbers are between $0.4$ (on LFW) and $0.55$ (on Adience). \textcolor{black}{PrivacyNet is somewhere in between FlowSAN--3 and FlowSAN--5 with the weakest ARR score on LFW being $0.399$ and on Adience $0.562$.} (ii) Second, the robustness of the privacy models varies from dataset to dataset. \textcolor{black}{While the results on LFW and MUCT are relatively consistent for most privacy models}, the weakest ARR scores on Adience are higher than the weakest scores on LFW and MUCT. This observation implies that attribute recovery robustness is in parts also affected by the initial data characteristics and not only by the recovery capabilities of the tested PrivacyProbers. Because the gender recognition performance  was already weaker on Adience than on the remaining two datasets, even small degradtions from the initial images, $I_{or}$, cause further degradations. This fact is then also reflected in the success of the attribute recovery attempts. \textcolor{black}{An exception to these observations is PrivacyNet, which performs well on LFW and Adience, but exhibits very low robustness on the MUCT datasets with several variants of PrivacyProber. The reason for this setting lies in the characteristics of the privacy model, which requires images aligned in a specific way. Since the faces in the MUCT dataset are aligned differently than those in LFW and Adience, the privacy enhancement with this data is less stable and, consequently, more susceptible to attribute-recovery attempts. }  (iii) Third, all evaluated PrivacyProbers are able to recover some level of gender information from the privacy--enhanced images, suggesting that the reported performance with zero-effort evaluation scenarios typically reported in the literature often overestimates the actual capabilities of the existing privacy models. 
\begin{figure*}[t]
\begin{minipage}[b]{0.085\textwidth}
    \centering
    {\footnotesize Original}
\end{minipage}%
\hfill
\begin{minipage}[b]{0.085\textwidth}
    \centering
    {\footnotesize PP-D\\ $k$--AAP}
\end{minipage}%
\begin{minipage}[b]{0.085\textwidth}
    \centering
    {\footnotesize PP-DI\\ $k$--AAP}
\end{minipage}%
\hfill
\begin{minipage}[b]{0.085\textwidth}
    \centering
    {\footnotesize PP-D\\ FGSM}
\end{minipage}%
\begin{minipage}[b]{0.085\textwidth}
    \centering
    {\footnotesize PP-DI\\ FGSM}
\end{minipage}%
\hfill
\begin{minipage}[b]{0.085\textwidth}
    \centering
    {\footnotesize PP-D\\ FlowSAN--3}
\end{minipage}%
\begin{minipage}[b]{0.085\textwidth}
   \centering
    {\footnotesize PP-DI\\ FlowSAN--3}
\end{minipage}%
\hfill
\begin{minipage}[b]{0.085\textwidth}
    \centering
    {\footnotesize PP-D\\ FlowSAN--5}
\end{minipage}%
\begin{minipage}[b]{0.085\textwidth}
    \centering
    {\footnotesize PP-DI\\ FlowSAN--5}
\end{minipage}%
\hfill
\begin{minipage}[b]{0.085\textwidth}
    \centering
    {\footnotesize PP-D\\ PrivacyNet}
\end{minipage}%
\begin{minipage}[b]{0.085\textwidth}
    \centering
    {\footnotesize PP-DI\\ PrivacyNet}
\end{minipage}%

\centering
\begin{minipage}[b]{0.085\textwidth}
    \centering
    \includegraphics[width=1\columnwidth, trim=0 10 0 0, clip]{tSNE_gender/original2.pdf}
\end{minipage}%
\hfill
\begin{minipage}[b]{0.085\textwidth}
    \centering
    \includegraphics[width=1\columnwidth, trim=0 10 0 0, clip]{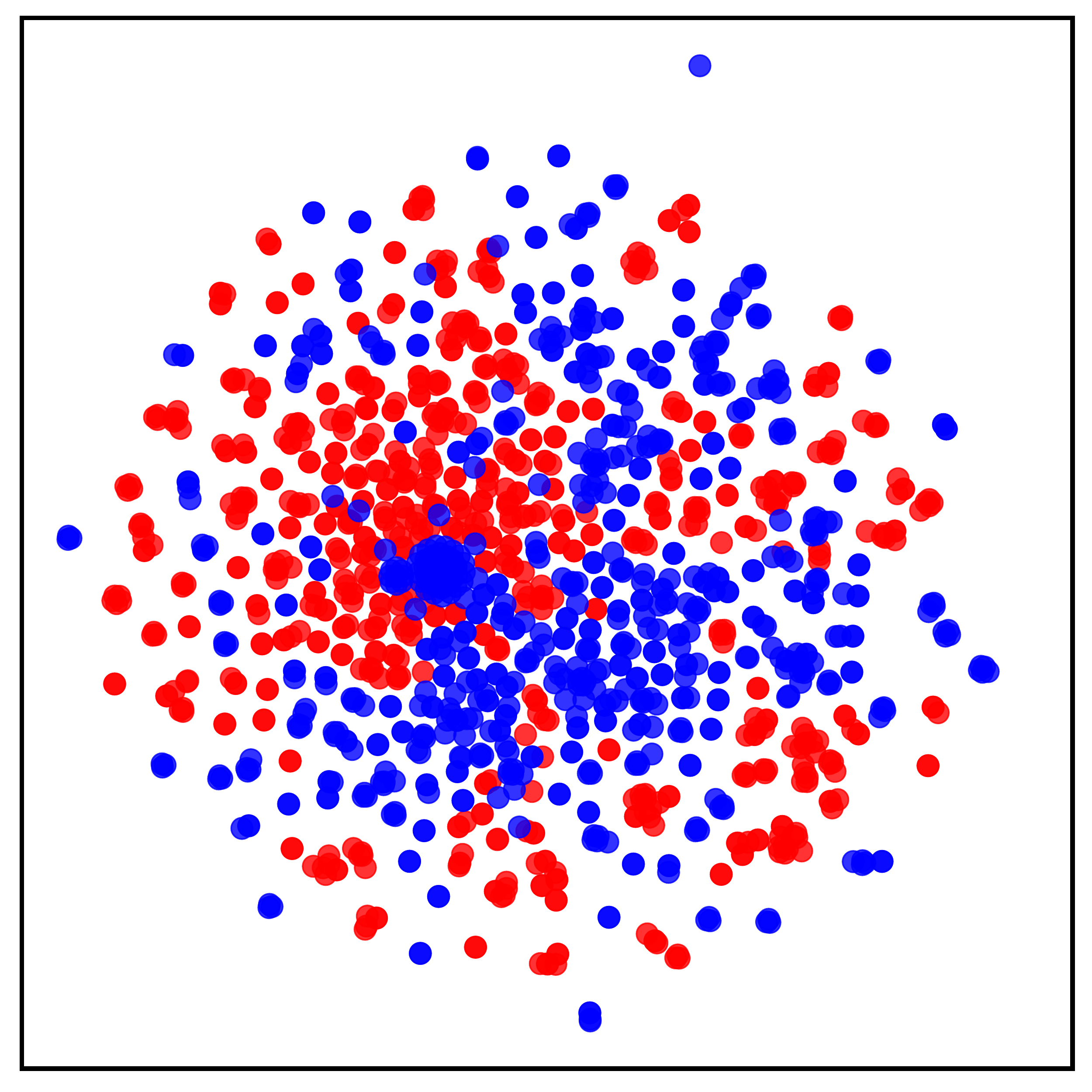}
\end{minipage}%
\begin{minipage}[b]{0.085\textwidth}
    \centering
   \includegraphics[width=1\columnwidth, trim=0 10 0 0, clip]{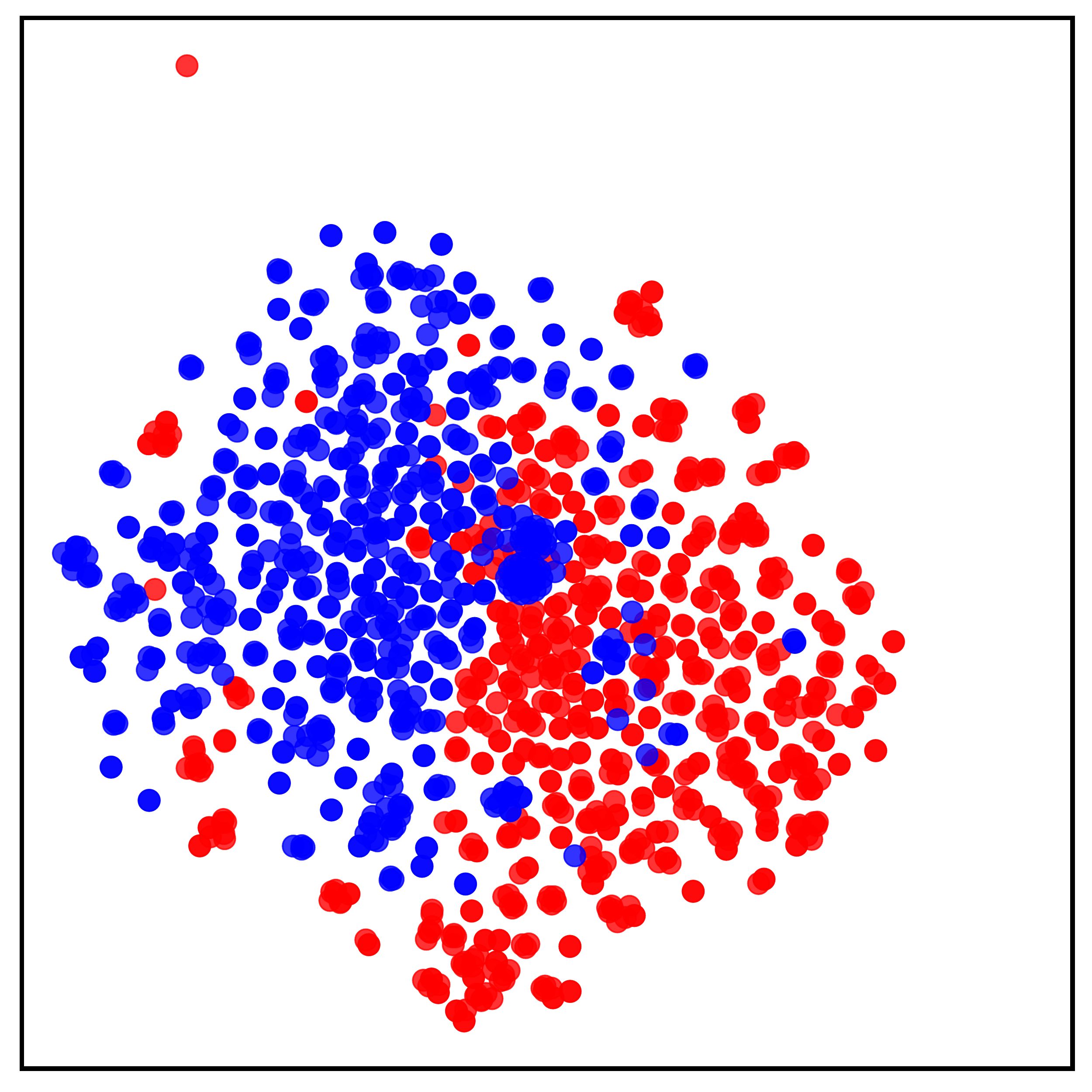}
\end{minipage}%
\hfill
\begin{minipage}[b]{0.085\textwidth}
    \centering
    \includegraphics[width=1\columnwidth, trim=0 10 0 0, clip]{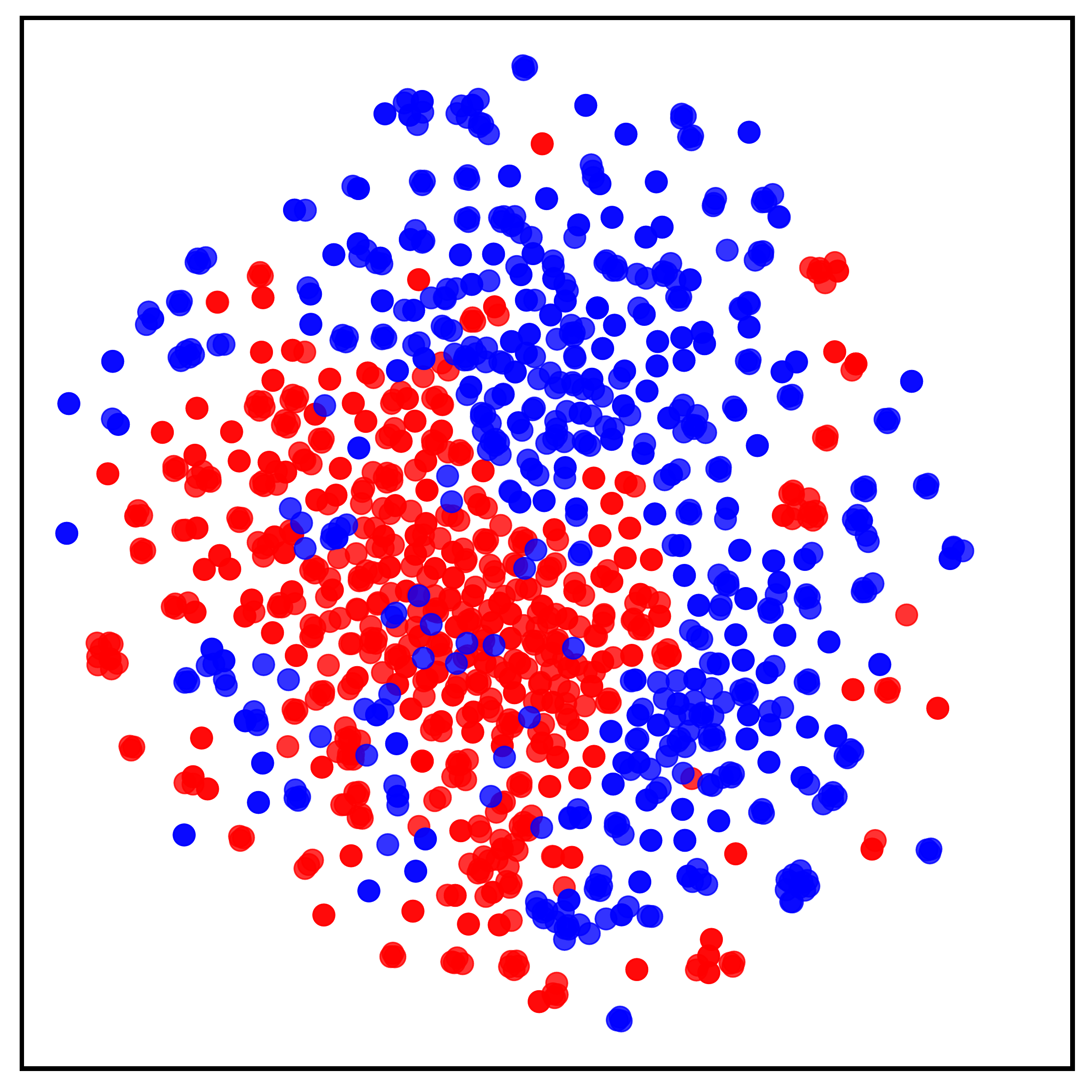}
\end{minipage}%
\begin{minipage}[b]{0.085\textwidth}
    \centering
    \includegraphics[width=1\columnwidth,trim=0 10 0 0, clip]{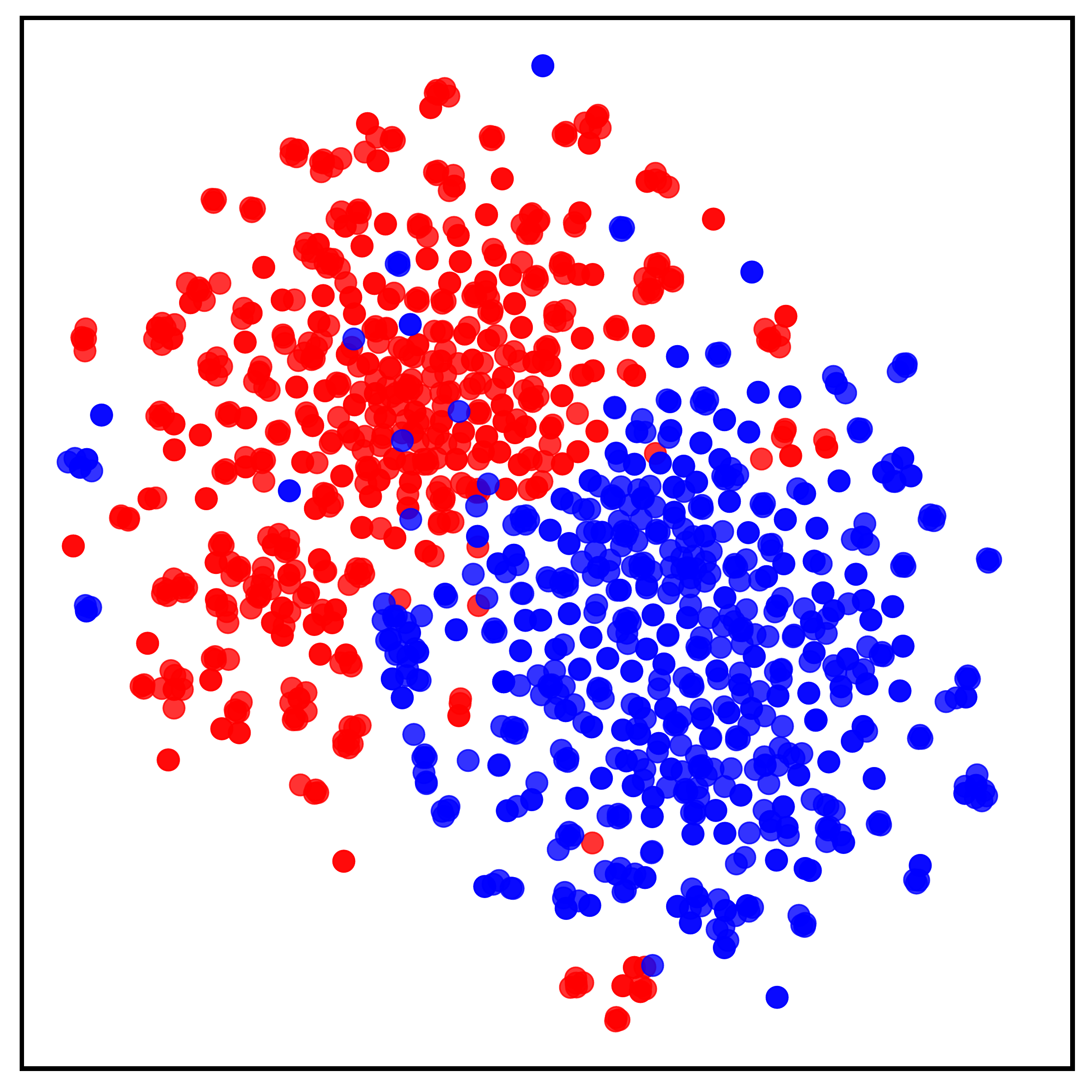}
\end{minipage}%
\hfill
\begin{minipage}[b]{0.085\textwidth}
    \centering
    \includegraphics[width=1\columnwidth, trim=0 10 0 0, clip]{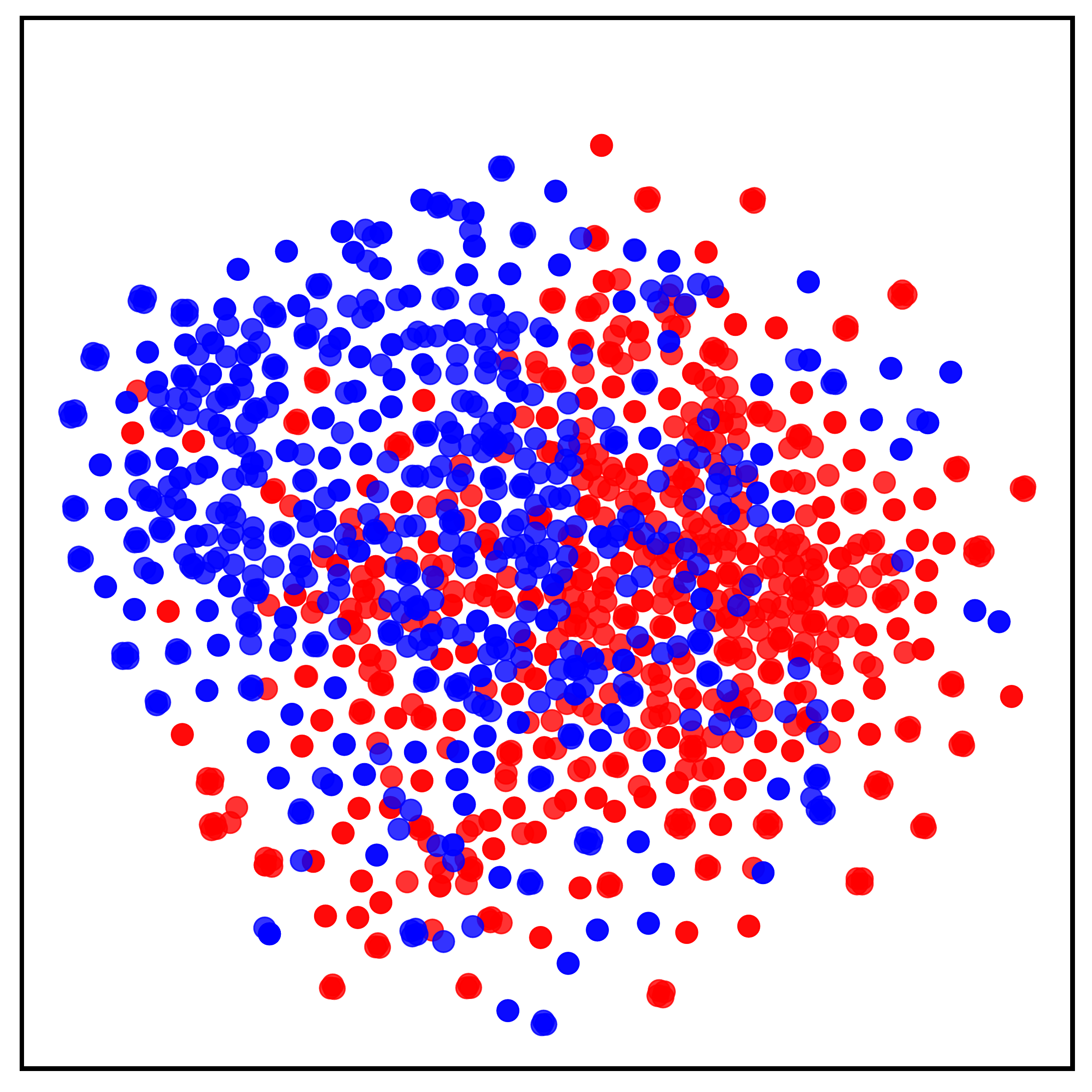}
\end{minipage}%
\begin{minipage}[b]{0.085\textwidth}
    \centering
   \includegraphics[width=1\columnwidth, trim=0 10 0 0, clip]{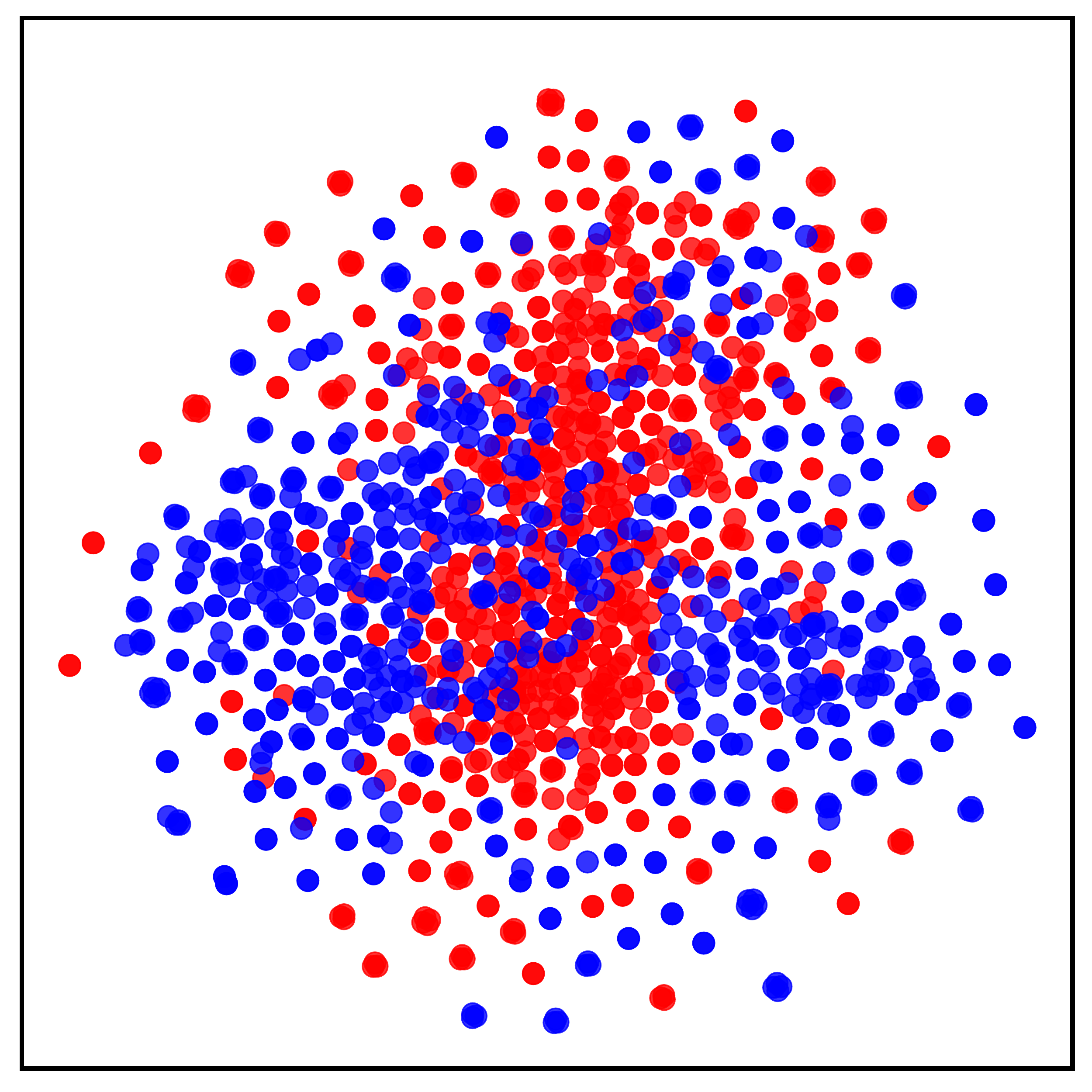}
\end{minipage}%
\hfill
\begin{minipage}[b]{0.085\textwidth}
    \centering
    \includegraphics[width=1\columnwidth, trim=0 10 0 0, clip]{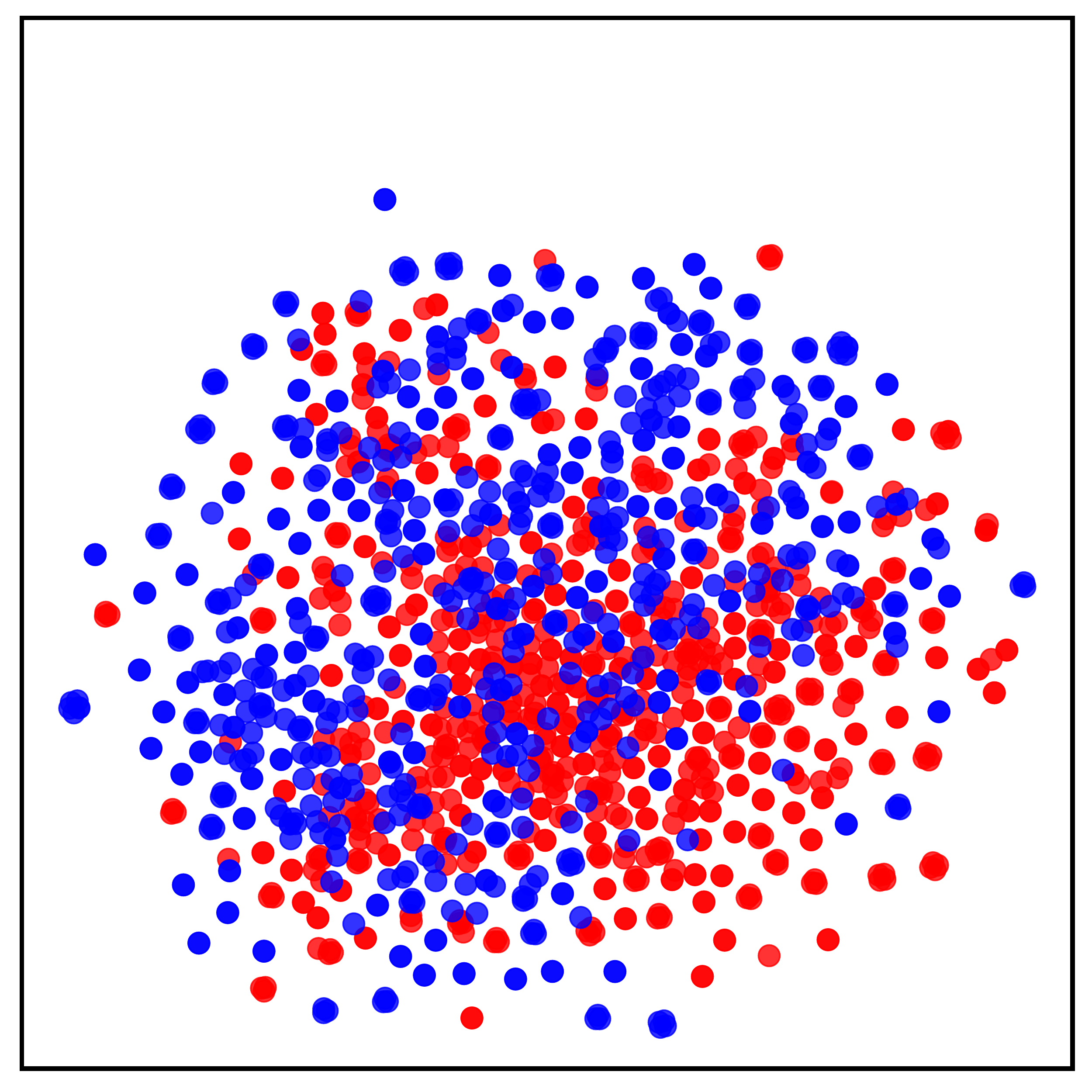}
\end{minipage}%
\begin{minipage}[b]{0.085\textwidth}
    \centering
    \includegraphics[width=1\columnwidth,trim=0 10 0 0, clip]{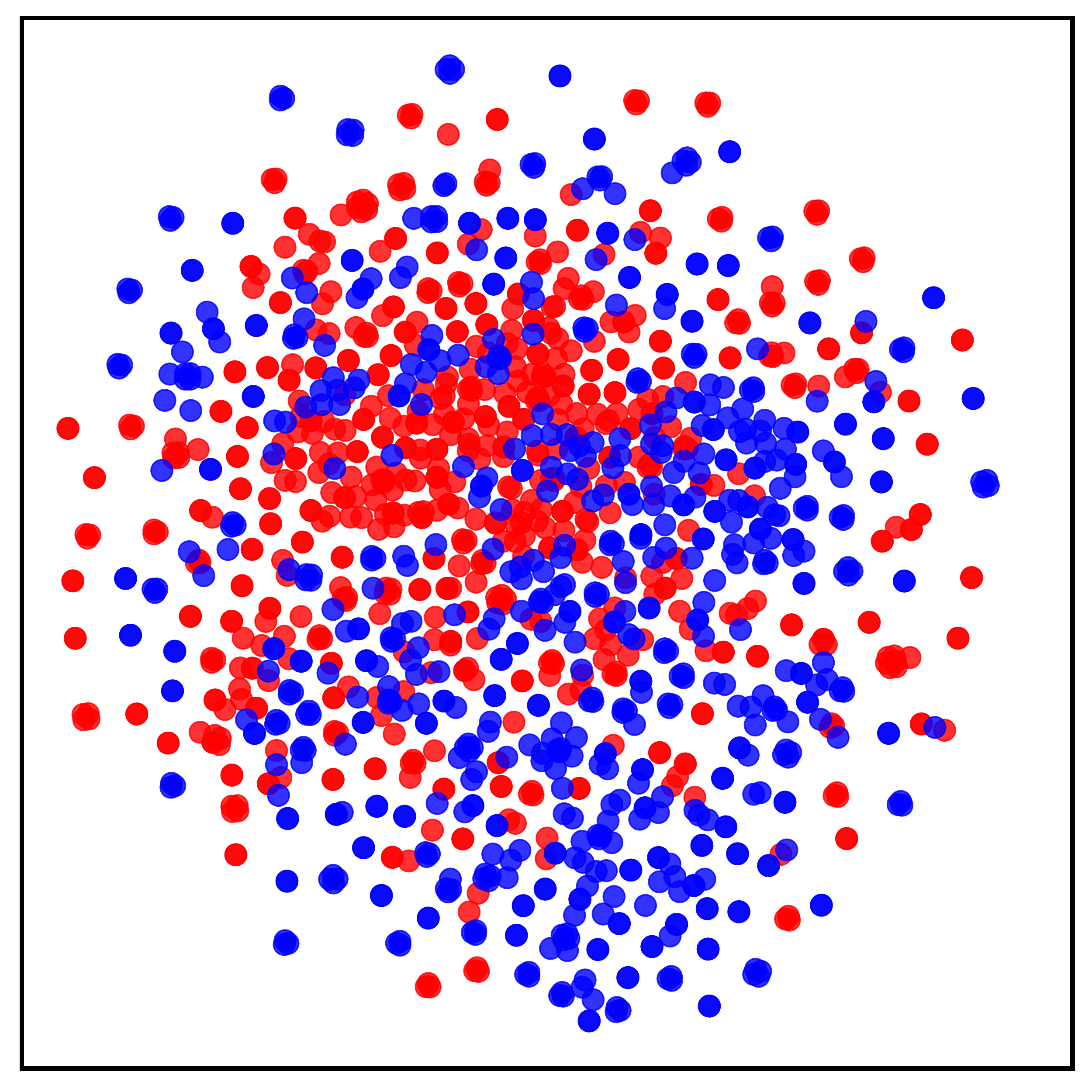}
\end{minipage}%
\hfill
\begin{minipage}[b]{0.085\textwidth}
    \centering
    \includegraphics[width=1\columnwidth,trim=0 10 0 0, clip]{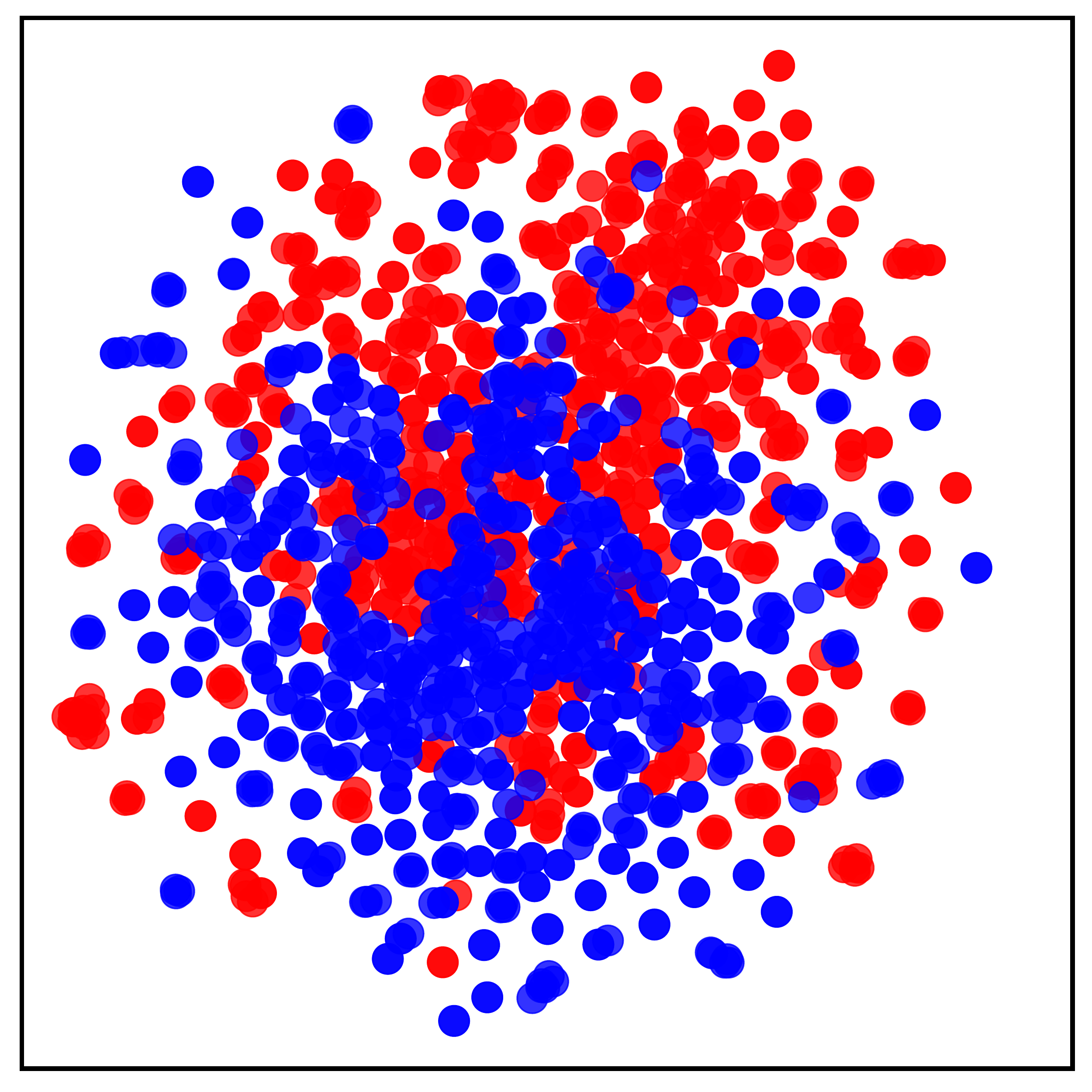}
\end{minipage}%
\begin{minipage}[b]{0.085\textwidth}
    \centering
    \includegraphics[width=1\columnwidth,trim=0 10 0 0, clip]{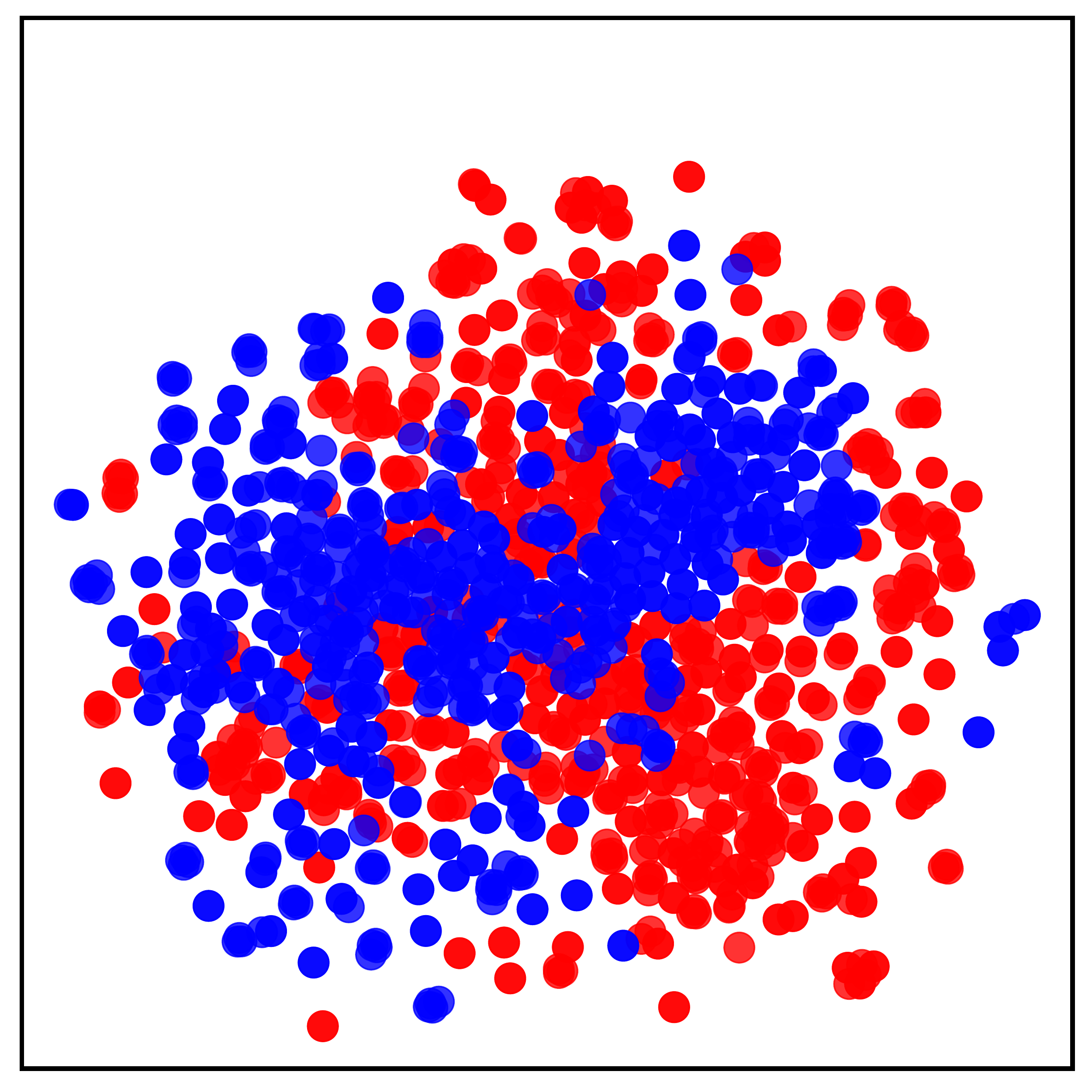}
\end{minipage}%
\\
\begin{minipage}[b]{0.085\textwidth}
    \centering
    \includegraphics[width=1\columnwidth, trim= 0 10 0 0, clip]{tSNE_id/original_id2.pdf}
\end{minipage}%
\hfill
\begin{minipage}[b]{0.085\textwidth}
    \centering
    \includegraphics[width=1\columnwidth, trim= 0 10 0 0, clip]{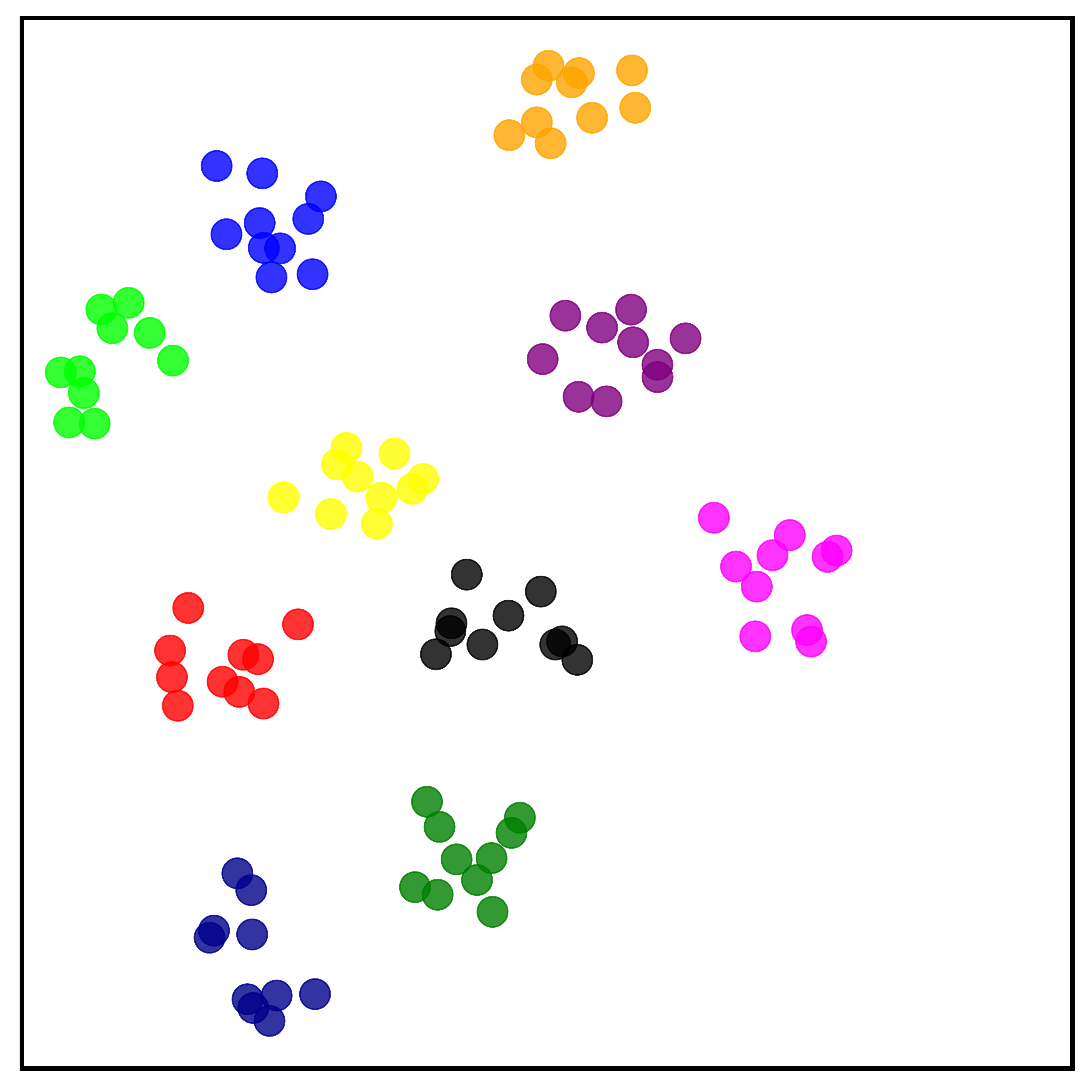}
\end{minipage}%
\begin{minipage}[b]{0.085\textwidth}
    \centering
   \includegraphics[width=1\columnwidth, trim= 0 10 0 0, clip]{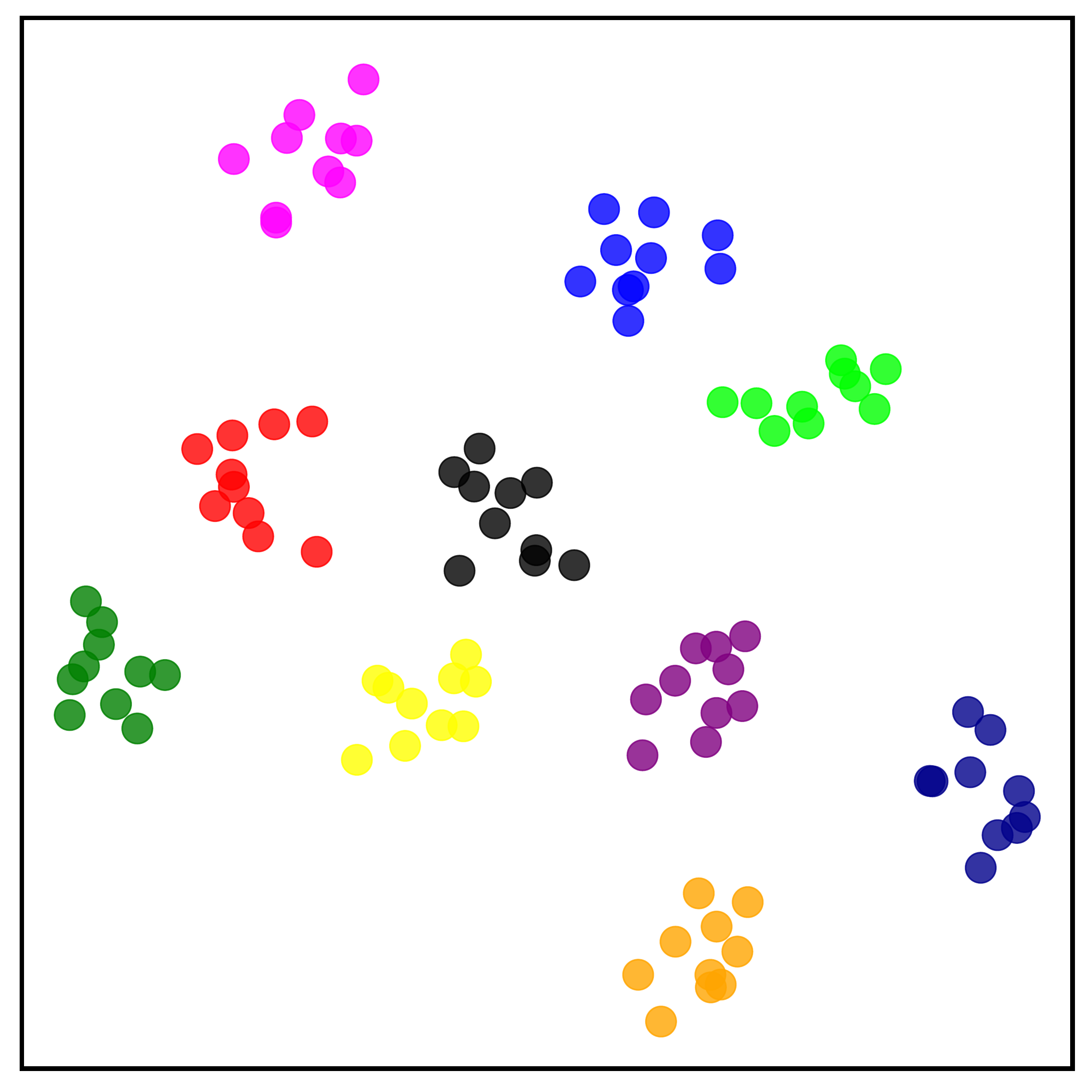}
\end{minipage}%
\hfill
\begin{minipage}[b]{0.085\textwidth}
    \centering
    \includegraphics[width=1\columnwidth,trim= 0 10 0 0, clip]{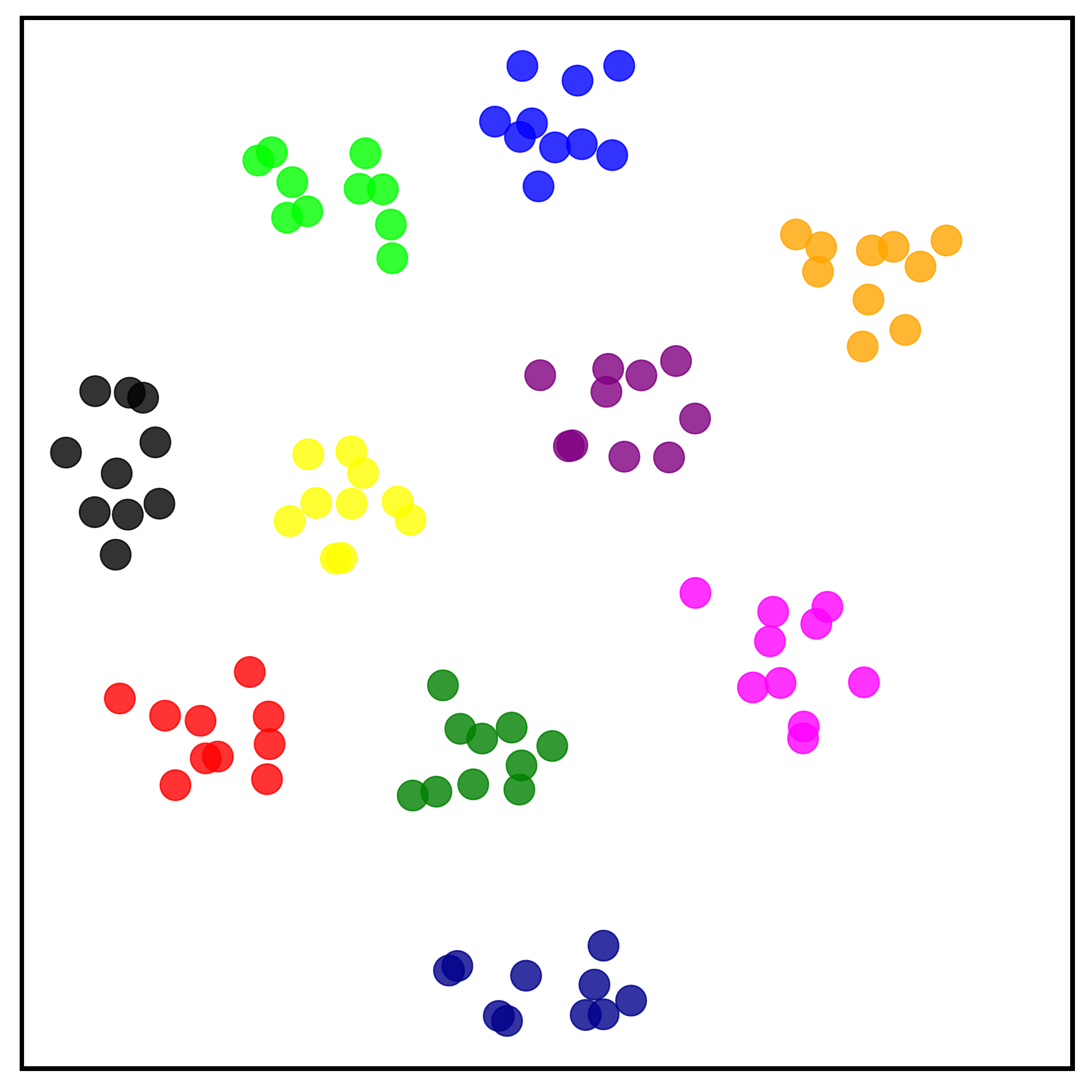}
\end{minipage}%
\begin{minipage}[b]{0.085\textwidth}
    \centering
    \includegraphics[width=1\columnwidth,trim= 0 10 0 0, clip]{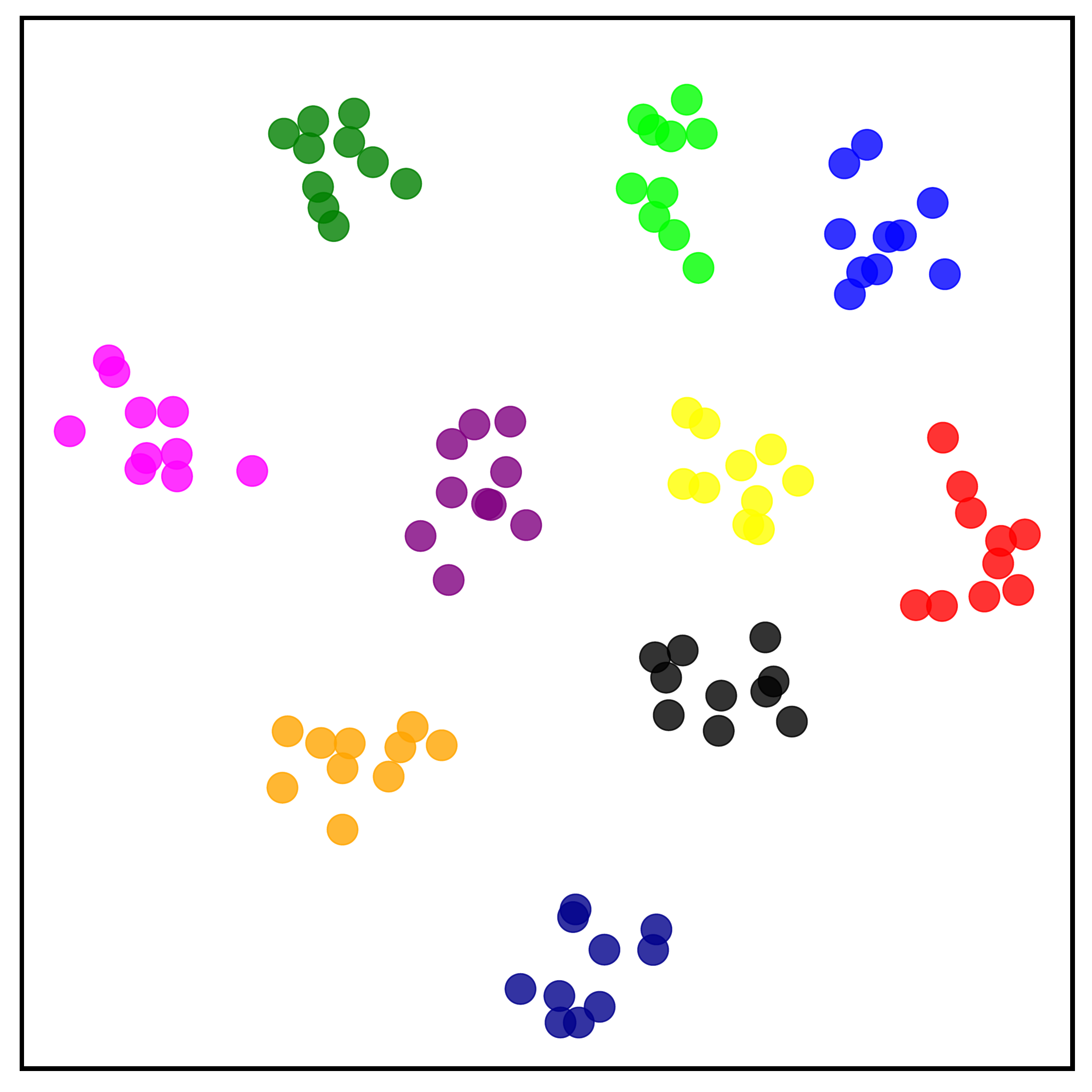}
\end{minipage}%
\hfill
\begin{minipage}[b]{0.085\textwidth}
    \centering
    \includegraphics[width=1\columnwidth, trim= 0 10 0 0, clip]{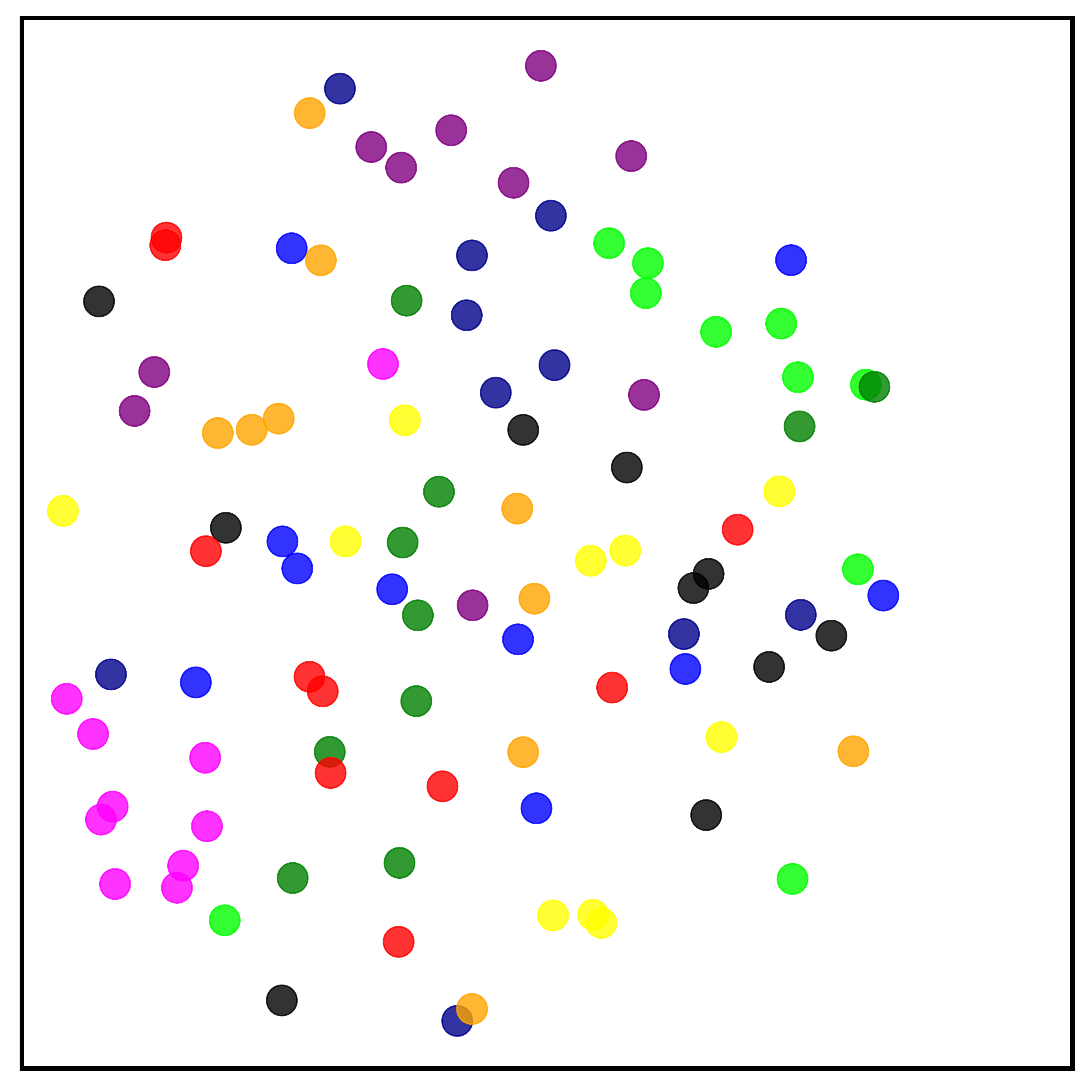}
\end{minipage}%
\begin{minipage}[b]{0.085\textwidth}
    \centering
   \includegraphics[width=1\columnwidth, trim= 0 10 0 0, clip]{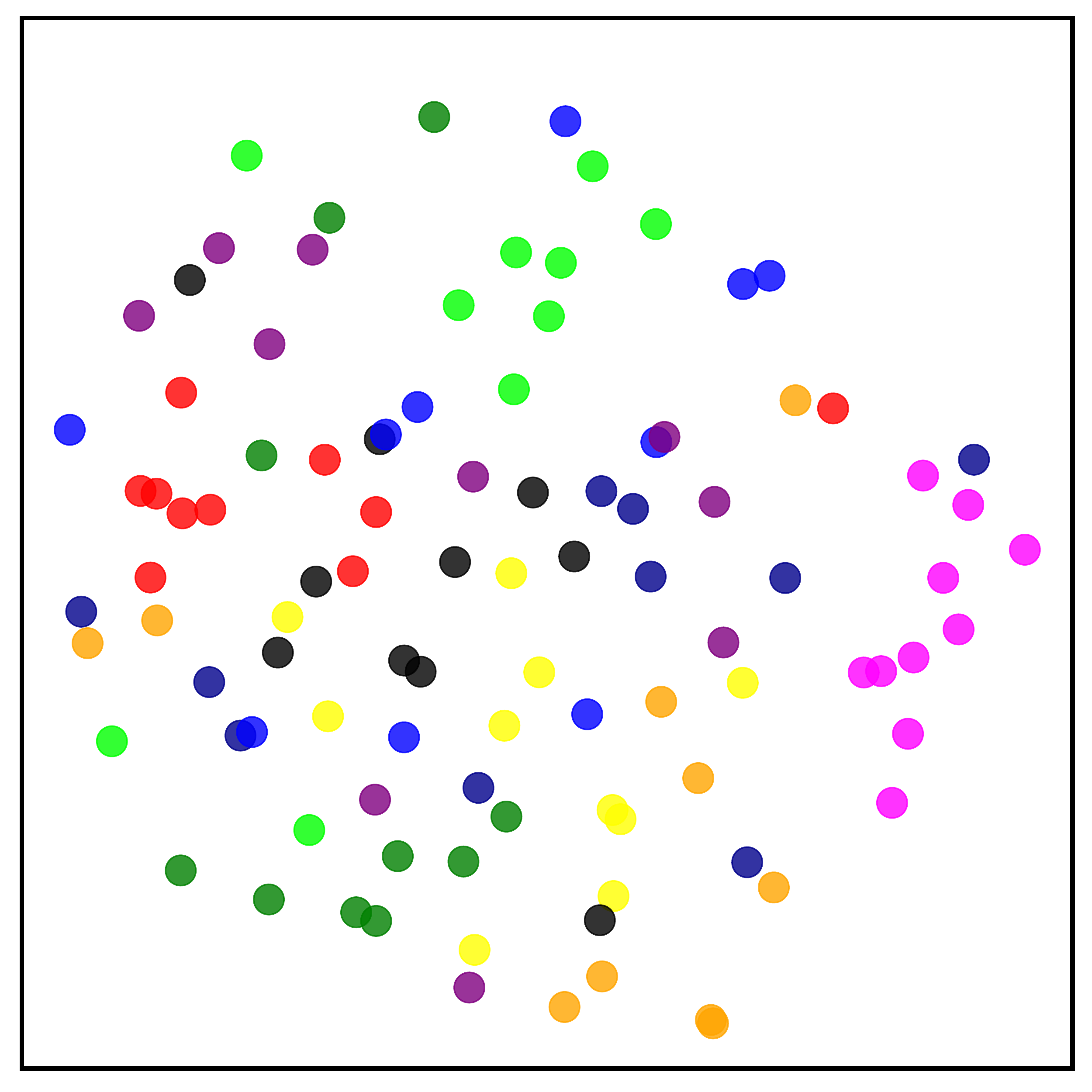}
\end{minipage}%
\hfill
\begin{minipage}[b]{0.085\textwidth}
    \centering
    \includegraphics[width=1\columnwidth, trim= 0 10 0 0, clip]{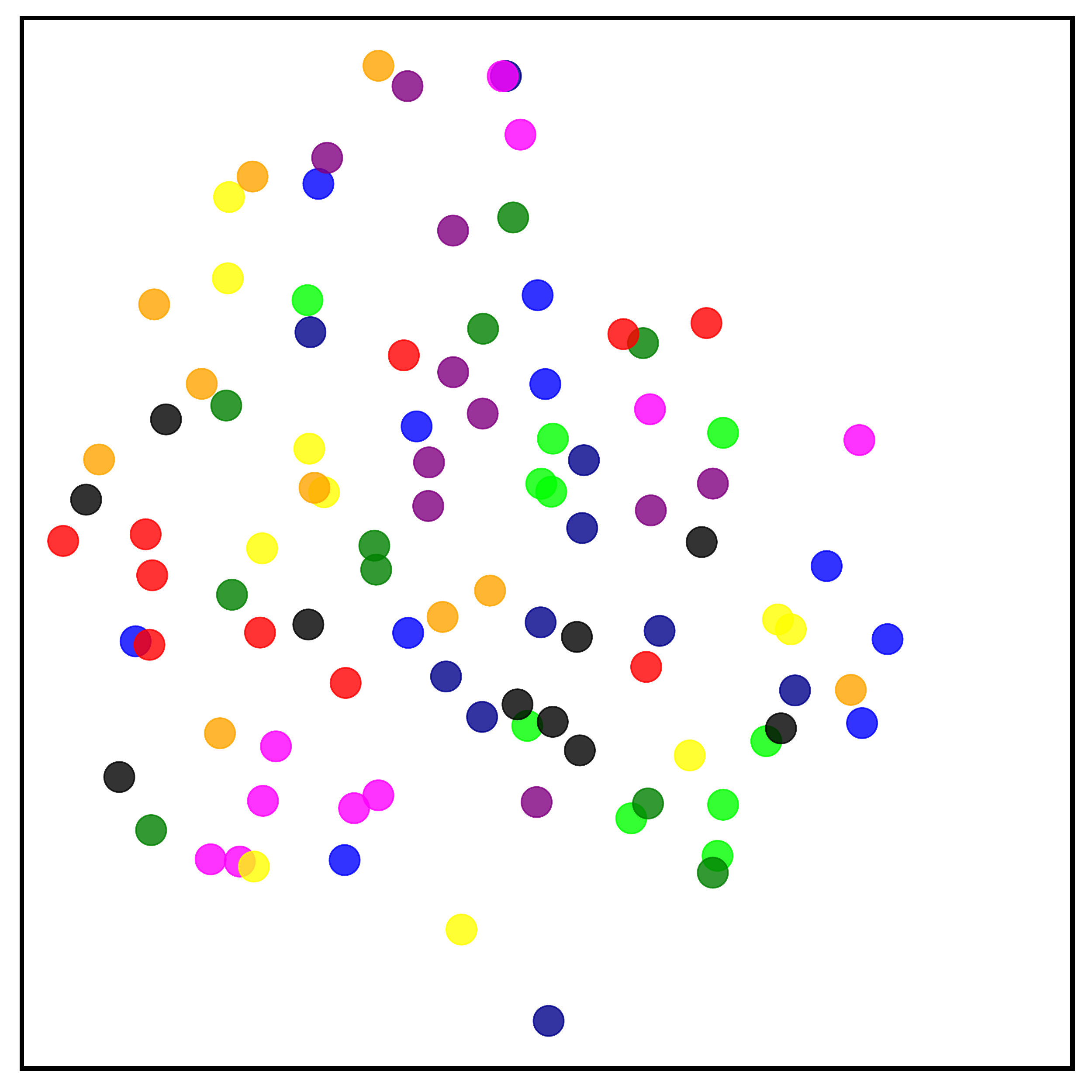}
\end{minipage}%
\begin{minipage}[b]{0.085\textwidth}
    \centering
    \includegraphics[width=1\columnwidth,trim= 0 10 0 0, clip]{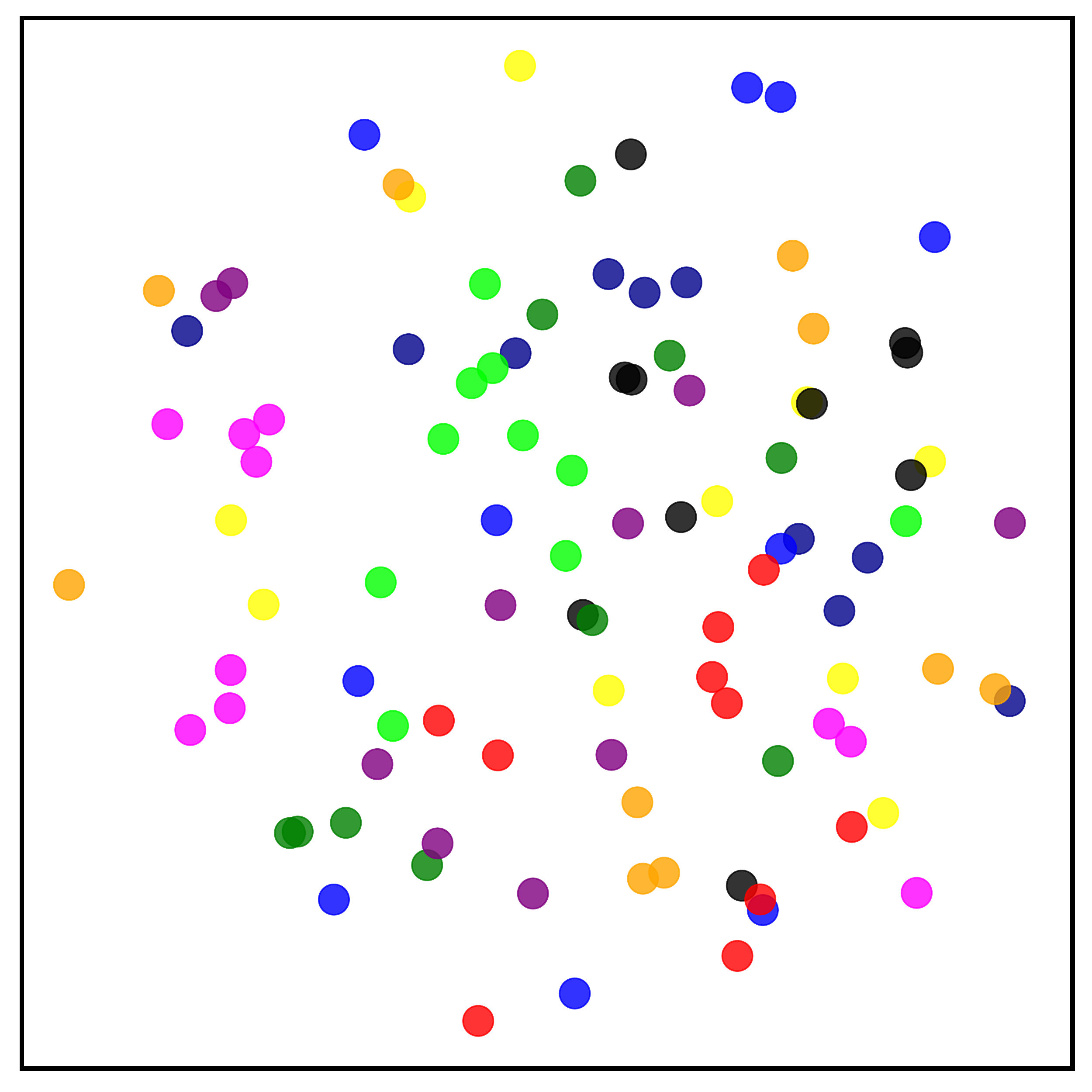}
\end{minipage}%
\hfill
\begin{minipage}[b]{0.085\textwidth}
    \centering
    \includegraphics[width=1\columnwidth,trim= 0 10 0 0, clip]{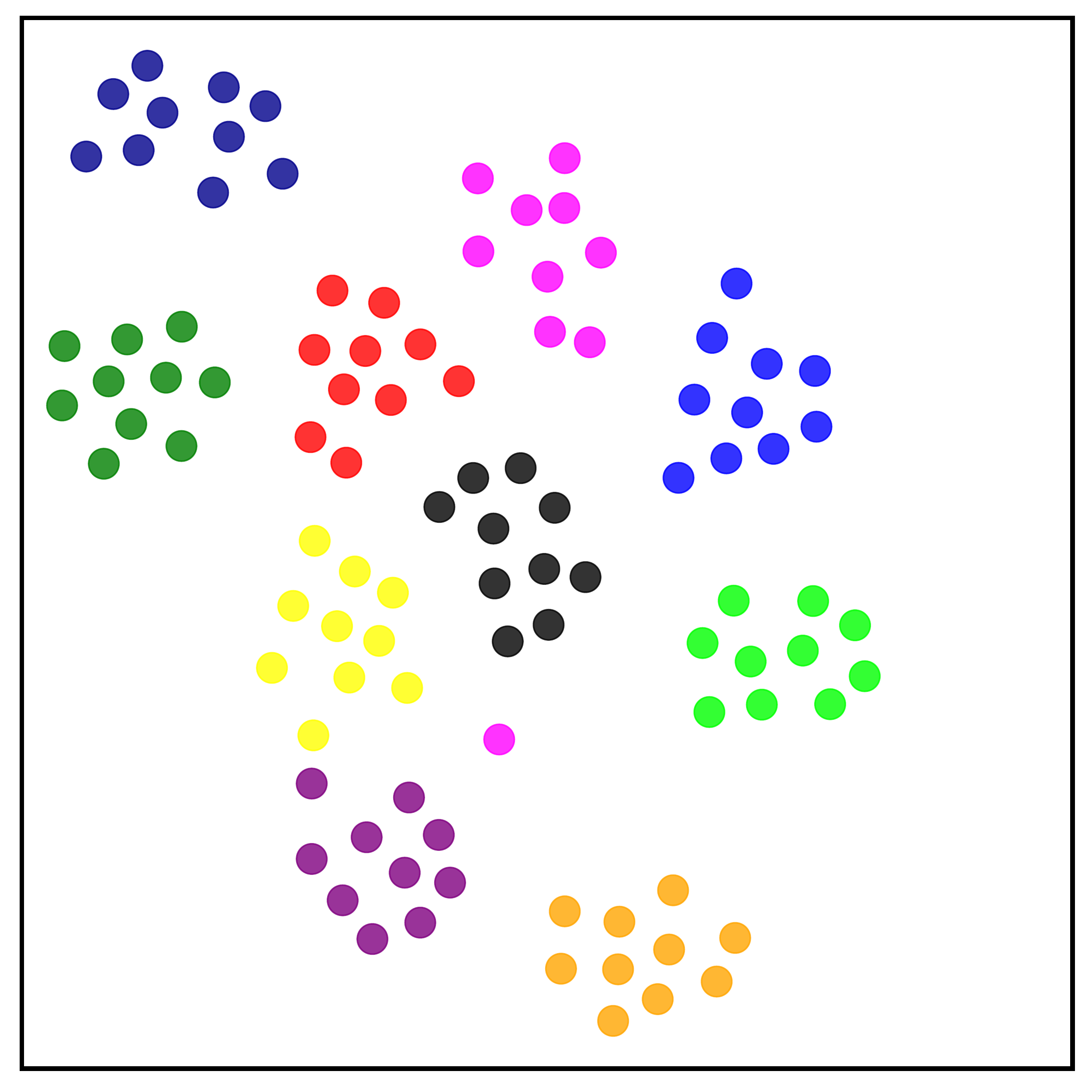}
\end{minipage}%
\begin{minipage}[b]{0.085\textwidth}
    \centering
    \includegraphics[width=1\columnwidth,trim= 0 10 0 0, clip]{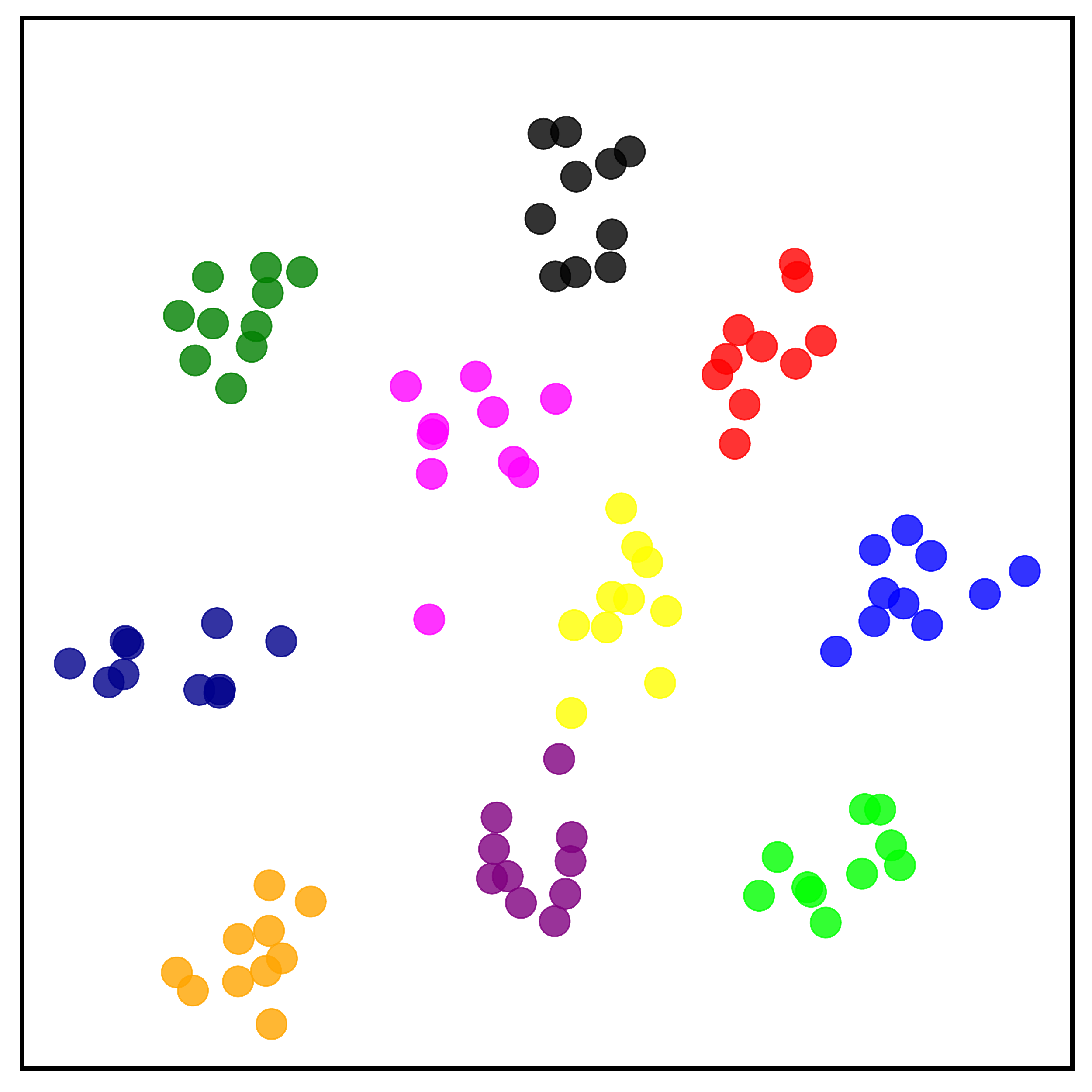}
\end{minipage}
\caption{$t$--SNE plots (in 2D) for gender and identity features. Results are presented for the original face images (far left) and images processed from with the best performing one--stage and two stage PrivacyProbers, i.e., PP--D and PP--DI. The gender in the top row are generated from $700$ randomly sampled LFW images of each gender and the subject plots in the bottom row are generated based on $10$ randomly selected images of the $10$ largest LFW classes. Best viewed in color.\label{fig: tSNE on LFW PP}}
\end{figure*}

\textbf{2) Attribute recovery vs. verification performance.} The implemented PrivacyProbers aim to reconstruct attribute information and, as a result, alter the characteristics of the facial images. To analyze the impact of attribute recovery on the trade--off between identity preservation and attribute recovery robustness, we plot the difference in AUC scores between the original and the attribute--recovered images for both verification and gender recognition experiments in Fig.~\ref{fig: coors_plots}. Points indicating that all the identity and gender information contained in the original images was recovered are located at the origin of the plots. These points correspond to the least robust privacy models. Different from the scalar ARR scores analyzed above, the presented plots offer more insight into the behavior of the tested models, but also the characteristics of the attribute recovery attempts. 

\begin{table*}[t]
\centering
\renewcommand{\arraystretch}{1.15}
\caption{\textcolor{black}{PrivacyProber as an adversarial defense mechanism. While PrivacyProber is designed as an attribute recovery approach for soft-biometric privacy models, it also ensures competitive performance as an adversarial defense.}}
\label{Tab:adversarial}
\textcolor{black}{
\begin{tabular}{lcccc} \hline \hline
\multirow{ 2}{*}{Attack model} & \multirow{ 2}{*}{AUC after attack} & \multicolumn{3}{c}{AUC after defense}\\ \cline{3-5}
& & AE recovery \cite{folz2020adversarial} & PP--D recovery (ours) & PP--DI recovery (ours) \\
 \hline
FGSM~\cite{goodfellow2014explaining} & $0.005\pm 0.000$ & $0.983 \pm 0.008$ & $0.957 \pm 0.017$ & $0.976 \pm 0.012$ \\
Carlini-Wagner~\cite{carlini2017towards} & $0.218\pm 0.059$ & $0.962 \pm 0.017$ & $0.848 \pm 0.062$ & $0.906 \pm 0.040$\\
AdvDrop~\cite{duan2021advdrop} & $0.299\pm 0.040$ & $0.600 \pm 0.017$ & $0.659 \pm 0.023$ & \textbf{$0.665 \pm 0.022$}  \\
\hline \hline
\end{tabular}\vspace{-2mm}}
\end{table*}

As can be seen from Fig.~\ref{fig: coors_plots}, the information content can be restored to close to the same level as before the application of $k$--AAP and FGSM on LFW and MUCT given the most effective PrivacyProbers. On these two datasets, FlowSAN--3--recovered images perform similarly to the original ones in terms of verification performance, but still offer a certain level of soft--biometric privacy. This result points to the robustness of the privacy model, but also shows that gender information can be recovered without affecting identity cues. FlowSAN--5 exhibits the greatest level of robustness to recovery attempts, but at the expense of the largest loss of identity information among all models. \textcolor{black}{PrivacyNet preserves identity information very well on LFW and MUCT, but (across these two datasets) offers robustness to recovery attempts only on LFW. On MUCT, on the other hand,  a significant amount of soft-biometric information can still be recovered. On Adience, similar observations as above can again be made for FGSM, that is, FGSM is again the least robust of all tested privacy models. \textcolor{black}{$k$--AAP, FlowSAN--3, FlowSAN--5 and PrivacyNet exhibit} higher levels of robustness, i.e., $\Delta AUC_g$ is comparably larger, but for most methods (except for PrivacyNet) these appear to be a consequence of a general loss of useful visual information, as indicated by the drop in the verification performance compared to the original images}. Among the PrivacyProbers, the most effective (on average) for the robustness analysis, with respect to both gender and identity information (on average) is PP--DI. The combination of image denoising and context--based inpainting appears to be highly effective in reconstructing attribute information and for evaluating the robustness of soft--biometric privacy models, tough other types of attribute recovery may offer additional insights.    

\subsubsection{Attribute Recovery and Feature Distributions}

Attribute recovery attempts change the visual characteristics of facial images, as illustrated in Fig.~\ref{fig:vis_impacts}. Because these changes also affect the feature representations generated by the recognition models, we next analyze the feature distributions generated from the attribute--recovered images. 
Fig.~\ref{fig: tSNE on LFW PP} compares the $t$--SNE based distributions produced by the face and gender-recognition models. Here, the same setup (involving LFW) as in Section~\ref{Sec:featureDistributionLFW} is utilized to generate feature vectors for the plots. To keep the analysis concise, only the best performing one-- and two--stage PrivacyProbers are considered, i.e., PP--D and PP--DI.

\begin{table*}[t]
\centering
\caption{\textcolor{black}{AUC scores ($\mu\pm\sigma$) for privacy-enhancement detection experiments. The learning-free (black box) APEND approach is compared against the classification--based T--SVM techniques for adversarial attack detection from~\cite{agarwal2020image} as well as the super-resolution based PREM detector  from~\cite{rot2022detecting}.} }
\label{Tab:detection_table}
\renewcommand{\arraystretch}{1.15}
\begin{tabular}{lccccc} \hline\hline
{Privacy Model} & {Dataset}  & APEND (ours) & PREM~\cite{rot2022detecting} & T--SVM (A)$^\dagger$~\cite{agarwal2020image} & T--SVM (B)$^\ddagger$~\cite{agarwal2020image} 
 \\\hline
\multirow{3}{*}{$k$--AAP}  & LFW  & $0.982 \pm 0.002$ &   
                                     $0.957 \pm 0.005$ &  
                                     $0.743 \pm 0.013$ &
                                     $0.984\pm 0.007$  \\
                            & MUCT & $0.984 \pm 0.002$ & 
                                     $0.965 \pm 0.005$ & 
                                     $0.727 \pm 0.030$ &
                                     $0.534\pm 0.021$ \\
                            & Adience  & $0.820 \pm 0.004$ & 
                                    $0.629 \pm 0.007$ &  
                                    $0.604 \pm 0.008$ &
                                    $0.541\pm 0.007$  \\
                            \arrayrulecolor{gray}\hline
\multirow{ 3}{*}{FGSM} & LFW  & $0.993 \pm 0.002$ &
                                $0.989 \pm 0.003$ &  
                                $0.894 \pm 0.013$&
                                $0.921\pm 0.002$ \\
                            & MUCT &  $0.995 \pm 0.002$ &
                                $0.989 \pm 0.003$ &
                                $0.858 \pm 0.015$ &
                                $0.552\pm 0.002$  \\
                            & Adience  & $0.907 \pm 0.001$ &
                                $0.858 \pm 0.003$ &  
                                $0.595 \pm 0.016$ &
                                $0.471\pm 0.004$ \\
\arrayrulecolor{gray}\hline

\multirow{3}{*}{FlowSAN--3} & LFW  & $0.984\pm 0.002$ &
                                 $0.987 \pm 0.002$ &  
                                 $1.000 \pm 0.000$ &
                                 $0.997\pm 0.002$ \\
                            & MUCT & $0.991 \pm 0.002$ &
                                 $0.993 \pm 0.001$ &  
                                 $1.000 \pm 0.000$&
                                 $0.524\pm 0.020$ \\
                            & Adience  & $0.754 \pm 0.003$ &
                                $0.775 \pm 0.002$ &  
                                $1.000 \pm 0.000$&
                                $0.609\pm 0.005$ \\
\arrayrulecolor{gray}\hline
\multirow{3}{*}{FlowSAN--5} & LFW  & $0.981 \pm 0.002$ &
                                $0.988 \pm 0.001$ &  
                                $1.000 \pm 0.000$&
                                $0.998\pm 0.001$ \\
                            & MUCT & $0.991 \pm 0.003$ &
                            $0.994 \pm 0.001$ &  
                            $1.000 \pm 0.000$&
                            $0.531\pm 0.018$ \\
                            & Adience & $0.770 \pm 0.006$ &
                            $0.793 \pm 0.011$ &  
                            $1.000 \pm 0.000$&
                            $0.563\pm 0.005$ \\
\arrayrulecolor{gray}\hline
\multirow{3}{*}{PrivacyNet} & LFW  & $0.989 \pm 0.002$ &
                             $ 0.978 \pm 0.002 $ &  
                             $0.533\pm 0.017$ &
                             $0.765\pm 0.004$ \\ 
                            & MUCT & $0.995 \pm 0.002 $ &
                                     $0.995 \pm 0.002 $ &  
                                     $0.501\pm 0.029$ &
                                     $0.205\pm 0.020$  \\
                            & Adience & $0.931 \pm 0.002 $ &
                                     $ 0.948 \pm 0.002 $ &  
                                     $0.439\pm 0.014$ &
                                     $0.254\pm 0.006$ \\
\hline

Average performance & & $\mathbf{0.940 \pm 0.084}$ &
 $0.926 \pm 0.108 $ & $0.792 \pm 0.009$ & $0.630 \pm 0.008$ \\ 
\arrayrulecolor{black}
\hline \hline
\multicolumn{2}{l}{\scriptsize $^\dagger$ Detection model trained on FlowSAN--5 and LFW.} \\
\multicolumn{2}{l}{\scriptsize $^\ddagger$ Detection model trained on $k$--AAP and LFW.}\\
\end{tabular}
\end{table*}

The presented plots support the observations made in the previous section. Attribute recovery (with PP--D and PP--DI) contributes towards well separated gender classes in the feature space for $k$--AAP and FGSM with minimal impact of the separability of the identity classes. For FlowSAN-3 and FlowSAN--5 we see less gender separation due to higher robustness of the privacy models and again minimal impact on identity information. \textcolor{black}{A similar observation can also be made for PrivacyNet, where the gender overlap is reduced compared to the original privacy-enahnced images,while the identitiy separation is not degraded.}

\textcolor{black}{\subsubsection{PrivacyProber and Adversarial Attacks}\vspace{1mm}}

\textcolor{black}{PrivacyProber is designed to recover attribute information, when this information is  protected by soft-biometric privacy models. While such models often also incorporate adversarial perturbations, they are typically implemented with mechanisms that go beyond adversarial methods for two reasons. Firstly, they strive towards a dual goal, i.e., (i) to conceal information by fooling a classifier and (ii) to preserve the utility of the data (e.g. verification accuracy, image quality etc.). Typical adversarial examples only try to achieve the first goal. Secondly, besides relying on adversarial noise, methods for soft--biometric privacy--enhancement also often include an image synthesis step, whereas classical adversarial methods do not. }

\textcolor{black}{Nevertheless, because adversarial methods and  soft bio\-metric privacy models share some commonalities, we explore in this section how PrivacyProber fares as a tool for adversarial defense. For the evaluation, we implement three adversarial attack methods, AdvDrop~\cite{duan2021advdrop}, Carlini-Wagner~\cite{carlini2017towards} and FGSM~\cite{goodfellow2014explaining}, and test them again on test images from LFW and with the two best-performing ProvacyProber variants, PP-D and PP-DI. Additionally, we compare PrivacyProber against a recent basline adversarial defense mechanism, i.e., the auto-encoder (AE) defense from~\cite{folz2020adversarial}. The adversarial attacks are designed to induce gender misclassifications and performance is measured in terms of AUC of the ROC cruves, generated in gender recognition experiments.} 

\textcolor{black}{We observe from the results in Table~\ref{Tab:adversarial} that PrivacyProber: (i) performs comparable to the state-of-the-art defense with the FGSM attack, (ii) is slightly behind, but still competitive with the Carlini-Wagner attack, and (iii) yields somewhat better performance with the most recent AdvDrop attack. The presented results suggest that despite the fact that PrivacyProber was primarily designed for robust evaluation of soft-biometric privacy models, it can also be used to explore the competitiveness of adversarial attacks.}\\

\subsection{Detecting (Soft--Biometric) Privacy Enhancement}

\textcolor{black}{In this section we now show that PrivacyProber can also be used to efficiently detect image tampering with soft-biometric privacy models. To this end, we implement the proposed APEND detector using three diverse PrivacyProber variants, i.e., PP-A, PP-DI and PP-B. We note that other combinations of PrivacyProber could be used for the experiments with different performance. However, our goal here is to provide a proof-of-concept for the detector and illustrate the benefits of attribute recovery and not to optimize performance indicators.} 

\subsubsection{Quantitative Evaluation \label{SubSEc: PD comparison}} 

\textcolor{black}{We note again that the problem of detecting privacy enhancement in facial images has, to the best of our knowledge, not yet been studied widely in the open literature. To the best of our knowledge, only the PREM approach from~\cite{rot2022detecting} aims to address the same task and is therefore also included in the evaluation. However, because soft--biometric privacy enhancement techniques share some characteristics with adversarial attacks, we also select the recent state--of--the art transformation--based detection technique (denoted T--SVM hereafter)} from~\cite{agarwal2020image} as a baseline for our experiments and compare it to APEND. T--SVM combines features from the discrete wavelet (DWT) and discrete sine  (DST) transforms with a support vector classifier (SVM) for tampering detection and requires training data to be able to learn how to discriminate between original and tampered images. We, therefore, consider the following setting for the experimental evaluation to ensure a fair comparison, i.e.:\vspace{1mm}
\begin{itemize}
   \item \textcolor{black}{\textbf{APEND and PREM}: These methods require no examples of tampered images or knowledge of the mechanism used for privacy enhancement and are, therefore, tested in a purely black-box scenario.}  \vspace{0.5mm}
    \item \textcolor{black}{\textbf{T-SVM:} Because T-SVM is supervised and does requires examples of tampered images or training, we design two experimental configurations that test for the generalization capabilities of this detector. 
    With the first configuration, T--SVM (A), is trained for the detection of FlowSAN--5 enhanced images only on data from LFW. For the second configuration, T--SVM (B), is trained for detecting $k$--AAP based enhancement, again only on LFW. The detection models from both configurations are then tested on all datasets and with all privacy models, thus, simulating a black-box evaluation scenario.}
\end{itemize}\vspace{1mm}

\textcolor{black}{Table~\ref{Tab:detection_table} shows that in the supervised approach,  T--SVM, is the most competitive on the datasets and privacy model it was trained on, while the performance with unseen models (and in most cases  also with unseen datasets) quickly deteriorates. The APEND and PREM techniques, on the other hand, are training-free and, therefore, generalize better to unseen models and across different datasets. While both PREM and APEND are quite competitive in most experiments, the aggregation of reconstruction evidence integrated in APEND leads to the best overall average performance of $0.940$ in terms of AUC, compared to $0.926$ for PREM. The added robustness through the aggregation process is especially evident in specific cases, such as for example with the $k$--AAP technique on the Adience dataset, where APEND outperforms PREM by more than $23\%$ in terms of AUC. The presented results clearly show the added value of training--free tampering detection as well as the aggregation-based approach to privacy-enhancement detection, which lead to highly encouraging results in the presented experiments.}

\begin{figure}[t]
\begin{minipage}{0.2\columnwidth}
    \centering
    \includegraphics[width=1\textwidth]{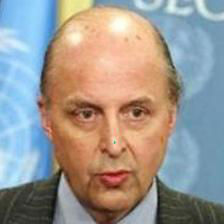}
\end{minipage}%
\begin{minipage}{0.2\columnwidth}
    \centering
    \includegraphics[width=1\textwidth]{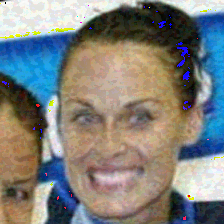}
\end{minipage}%
\begin{minipage}{0.2\columnwidth}
    \centering
    \includegraphics[width=1\textwidth]{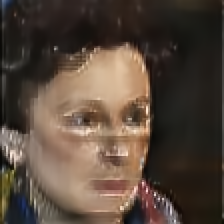}
\end{minipage}%
\begin{minipage}{0.2\columnwidth}
    \centering
    \includegraphics[width=1\textwidth]{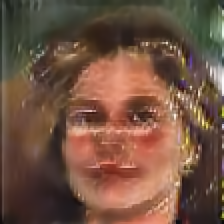}
\end{minipage}%
\begin{minipage}{0.2\columnwidth}
    \centering
    \includegraphics[width=1\textwidth]{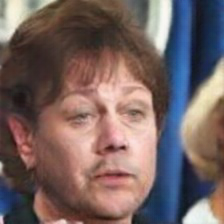}
\end{minipage}%
\\
\begin{minipage}{0.2\columnwidth}
    \centering
    \includegraphics[width=1\textwidth]{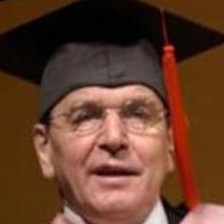}
    \text{\scriptsize (a) $k$--AAP}
\end{minipage}%
\begin{minipage}{0.2\columnwidth}
    \centering
    \includegraphics[width=1\textwidth]{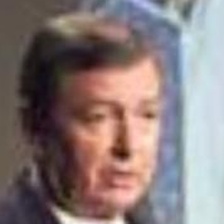}
    \text{\scriptsize (b) FGSM}
\end{minipage}%
\begin{minipage}{0.2\columnwidth}
    \centering
    \includegraphics[width=1\textwidth]{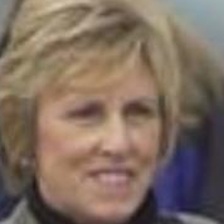}
    \text{\scriptsize (c) FlowSAN--3}
\end{minipage}%
\begin{minipage}{0.2\columnwidth}
    \centering
    \includegraphics[width=1\textwidth]{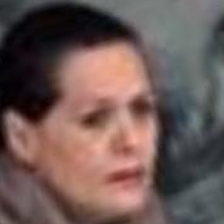}
    \text{\scriptsize (d) FlowSAN--5}
\end{minipage}%
\begin{minipage}{0.2\columnwidth}
    \centering
    \includegraphics[width=1\textwidth]{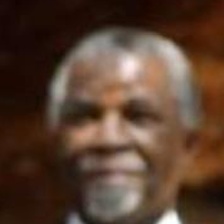}
    \text{\scriptsize (e) PrivacyNet}
\end{minipage}%
\caption{\textcolor{black}{Illustration of APEND failure cases: examples, where APEND did not detect the privacy enhancement (top row); examples, where APEND incorrectly flagged non--tampered images (bottom row)}.\label{fig:error_examples}}
\end{figure}

\subsubsection{Visual Analysis} 

\textcolor{black}{Fig.~\ref{fig:error_examples} shows a few example face images, \textcolor{black}{where the proposed APEND detector incorrectly flaged the presence of image tampering (i.e., privacy enhancement) at a decision threshold that ensures equal error (EER) rates in the (two-class) gender recognition experiments}. In the top row, we show images, where APEND failed to detect privacy-enhanced images. In most cases, these images contain strong image artifacts that make it challening to properly recover attribute information using the PrivacyProber variant used for the implementation of APEND. This leads to minute differences in the gender predictions between the privacy-enhanced and recovered images and eventually to misdetections. In the bottom row, we shows example images that have been flagged by APEND as being tampered, but in fact represent original images. As can be seen, such images are often of poor quality (due to blur, noise, etc.) and get improved by the recovery approaches in APEND. Thus, the predictions of the gender classifier change sufficiently to flag the images as tampered. While the performance of APEND is highly competitive (and close to ideal on most datasets), the presented examples suggest that there is room for further improvement by, for example, combining APEND with supervised detection techniques that should be able to also perform well with poor quality images.} 

\section{Conclusion\label{Sec:Conclusion}}

In this paper, we investigated the reliability of 
soft--biometric privacy--enhancing techniques and explored their robustness to attribute recovery attempts. We introduced PrivacyProber, a framework for the recovery of suppressed soft--biometric information from facial images, and used it 
in a comprehensive experimental evaluation. Experimental results on the LFW, MUCT and Adience datasets showed that there are considerable differences between the tested privacy models, both in terms of visual impact on the privacy--enhanced images as well as in terms of the level of privacy ensured. Additionally, we observed that (using our framework) is was possible to recover a considerable amount of suppressed (concealed) attribute information regardless of the privacy model used. However, the robustness of the tested synthesis--based techniques (i.e., FlowSANs \textcolor{black}{and PrivacyNet}) was observed to be considerable higher than that of the evaluated adversarial approaches (i.e., $k$--AAP and FGSM). These findings have considerable implications for future research in the area of biometric privacy enhancement, where more work is needed to improve robustness of existing models.

As another contribution, we showed that the proposed attribute recovery framework can also be used to detect privacy enhancement (e.g., tampering) in facial images. We proposed the APEND detector, and demonstrated that it can detect privacy enhancements with high accuracy even if different privacy models are utilized and data with diverse characteristics is used. This fact points to another threat vector with respect to biometric privacy models in that privacy--enhanced images can easily be identified and subjected to alternative means of processing that is less sensitive to artifacts and perturbations infused with the enhancement.

As part of our future work, \textcolor{black}{we plan to extend our robustness analysis to other biometric privacy models, such as deidentification techniques, which are based on different assumptions and require conceptually different models to restore the obscured information. The need for such robust evaluation schemes was also identified as a key issue in recent privacy surveys, e.g., in\cite{meden2021Survey}.}






%



\ifCLASSOPTIONcompsoc
  \section*{Acknowledgments}
\else
  \section*{Acknowledgment}
\fi

This research was supported in parts by the ARRS Project J2--1734 “Face deidentification with generative deep models”, ARRS Research Programs P2--0250 (B) “Metrology and Biometric Systems” and  P2--0214 (A) “Computer Vision”.

\ifCLASSOPTIONcaptionsoff
  \newpage
\fi



\bibliographystyle{IEEEtran}
\bibliography{refs}

\clearpage

%
%
%
\renewcommand{\thesection}{\Alph{subsection}.\arabic{subsection}}
\setcounter{subsection}{0}

\end{document}

%% file: sestavljena_slika.tex



\begin{figure*}[!t]

\minipage{0.16\textwidth}
\centering
    {\footnotesize Input image, $I_{or}$ \vspace{1mm}}\\
  \includegraphics[width=1\textwidth]{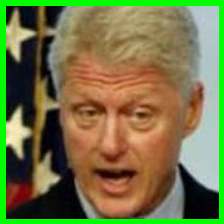}\\
  {\footnotesize $P(C_m|I_{or})=1.000$}
  
\endminipage
\hspace{2mm}
\minipage{0.01\textwidth}
\tikz{\draw[-,gray, densely dashed, thick](0,-2.8) -- (0,2.6);}
\endminipage
\hspace{1mm}
\minipage{0.78\textwidth}
\minipage{1\textwidth}
\minipage{0.09\textwidth}
\centering
\text{\footnotesize $I_{pr}$}
\endminipage
\hfill
\minipage{0.09\textwidth}
\centering
\text{\footnotesize PP--D}
\endminipage
\hfill
\minipage{0.09\textwidth}
\centering
\text{\footnotesize PP--I}
\endminipage
\hfill
\minipage{0.09\textwidth}
\centering
\text{\footnotesize \textcolor{blue}{PP--A}}
\endminipage
\hfill
\minipage{0.09\textwidth}
\centering
\text{\footnotesize PP--B}
\endminipage
\hfill
\minipage{0.09\textwidth}
\centering
\text{\footnotesize PP--DI}
\endminipage
\hfill
\minipage{0.09\textwidth}
\centering
\text{\footnotesize PP--DA}
\endminipage
\hfill
\minipage{0.09\textwidth}
\centering
\text{\footnotesize PP--DB}
\endminipage
\hfill
\minipage{0.09\textwidth}
\centering
\text{\footnotesize PP--IB}
\endminipage
\hfill
\minipage{0.09\textwidth}
\centering
\centering
\text{\footnotesize PP--AB}
\endminipage
\endminipage \\ 
\minipage{1\textwidth}
    \includegraphics[width=0.095\textwidth]{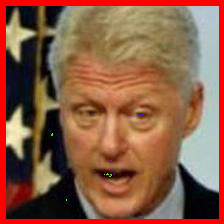}
        \includegraphics[width=0.095\textwidth]{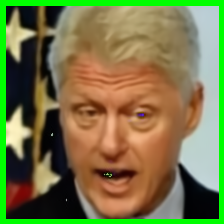}
    \includegraphics[width=0.095\textwidth]{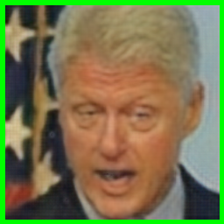}
     \includegraphics[width=0.095\textwidth]{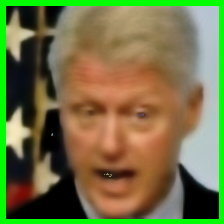}
         \includegraphics[width=0.095\textwidth]{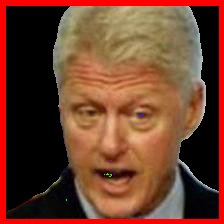}
                 \includegraphics[width=0.095\textwidth]{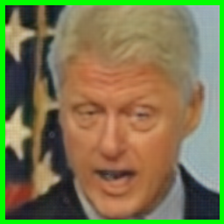}
       \includegraphics[width=0.095\textwidth]{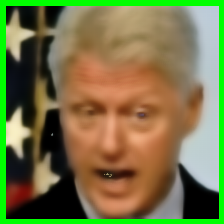}
        \includegraphics[width=0.095\textwidth]{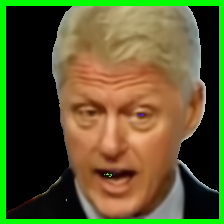}
        \includegraphics[width=0.095\textwidth]{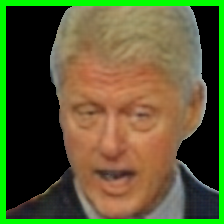}
        \includegraphics[width=0.095\textwidth]{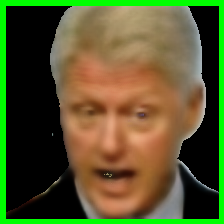}
\endminipage\\ \vspace{-2mm}     

\minipage{1\textwidth}
\minipage{0.09\textwidth}
\centering
\text{\scriptsize $0.298$}
\endminipage
\hfill
\minipage{0.09\textwidth}
\centering
\text{\scriptsize $0.999$}
\endminipage
\hfill
\minipage{0.09\textwidth}
\centering
\text{\scriptsize $1.000$}
\endminipage
\hfill
\minipage{0.09\textwidth}
\centering
\text{\scriptsize $1.000$}
\endminipage
\hfill
\minipage{0.09\textwidth}
\centering
\text{\scriptsize $0.224$}
\endminipage
\hfill
\minipage{0.09\textwidth}
\centering
\text{\scriptsize $1.000$}
\endminipage
\hfill
\minipage{0.09\textwidth}
\centering
\text{\scriptsize $1.000$}
\endminipage
\hfill
\minipage{0.09\textwidth}
\centering
\text{\scriptsize $0.991$}
\endminipage
\hfill
\minipage{0.09\textwidth}
\centering
\text{\scriptsize $1.000$}
\endminipage
\hfill
\minipage{0.09\textwidth}
\centering
\centering
\text{\scriptsize $1.000$}
\endminipage
\endminipage   \\ \vspace{-3mm}      
 
\minipage{1\textwidth}
    \includegraphics[width=0.095\textwidth]{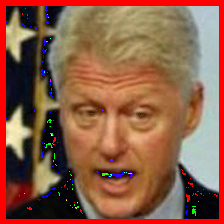}
        \includegraphics[width=0.095\textwidth]{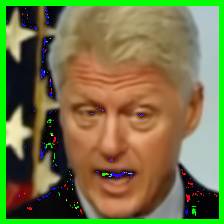}
     \includegraphics[width=0.095\textwidth]{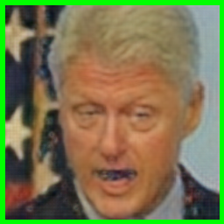}
                \includegraphics[width=0.095\textwidth]{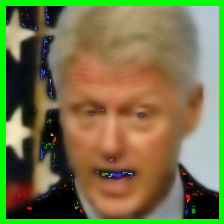}
         \includegraphics[width=0.095\textwidth]{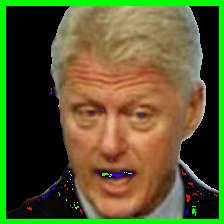}
         \includegraphics[width=0.095\textwidth]{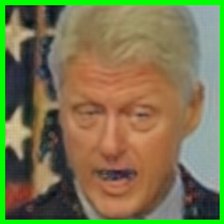}
                  \includegraphics[width=0.095\textwidth]{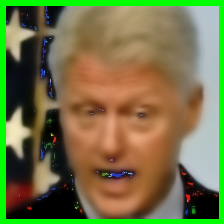}
         \includegraphics[width=0.095\textwidth]{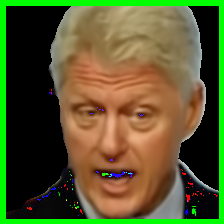}
        \includegraphics[width=0.095\textwidth]{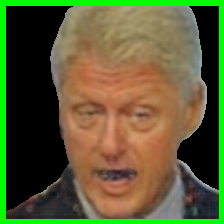}
        \includegraphics[width=0.095\textwidth]{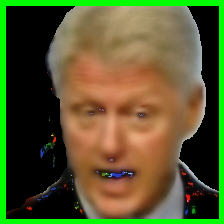}
\endminipage\\ \vspace{-2mm} 

\minipage{1\textwidth}
\minipage{0.09\textwidth}
\centering
\text{\scriptsize $0.199$}
\endminipage
\hfill
\minipage{0.09\textwidth}
\centering
\text{\scriptsize $1.000$}
\endminipage
\hfill
\minipage{0.09\textwidth}
\centering
\text{\scriptsize $1.000$}
\endminipage
\hfill
\minipage{0.09\textwidth}
\centering
\text{\scriptsize $1.000$}
\endminipage
\hfill
\minipage{0.09\textwidth}
\centering
\text{\scriptsize $0.854$}
\endminipage
\hfill
\minipage{0.09\textwidth}
\centering
\text{\scriptsize $1.000$}
\endminipage
\hfill
\minipage{0.09\textwidth}
\centering
\text{\scriptsize $1.000$}
\endminipage
\hfill
\minipage{0.09\textwidth}
\centering
\text{\scriptsize $1.000$}
\endminipage
\hfill
\minipage{0.09\textwidth}
\centering
\text{\scriptsize $1.000$}
\endminipage
\hfill
\minipage{0.09\textwidth}
\centering
\centering
\text{\scriptsize $1.000$}
\endminipage
\endminipage   \\ \vspace{-3mm}    

\minipage{1\textwidth}
     \includegraphics[width=0.095\textwidth]{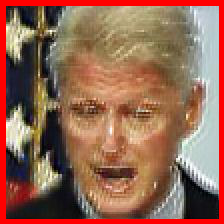}
       \includegraphics[width=0.095\textwidth]{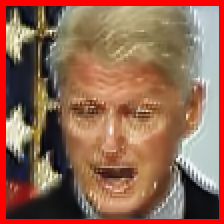}
      \includegraphics[width=0.095\textwidth]{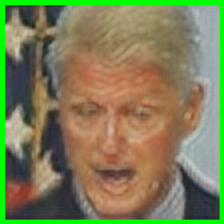}
        \includegraphics[width=0.095\textwidth]{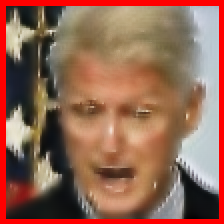}
        \includegraphics[width=0.095\textwidth]{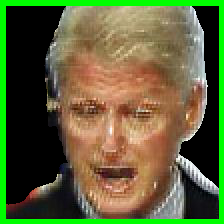}
       \includegraphics[width=0.095\textwidth]{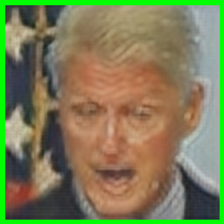}
       \includegraphics[width=0.095\textwidth]{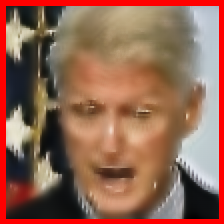}
        \includegraphics[width=0.095\textwidth]{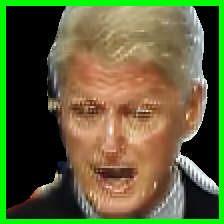}
         \includegraphics[width=0.095\textwidth]{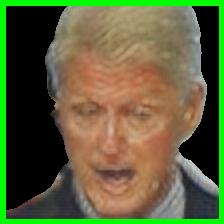}
       \includegraphics[width=0.095\textwidth]{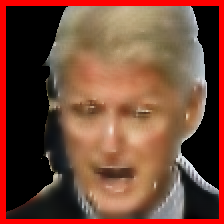}
\endminipage\\ \vspace{-2mm} 

\minipage{1\textwidth}
\minipage{0.09\textwidth}
\centering
\text{\scriptsize $0.184$}
\endminipage
\hfill
\minipage{0.09\textwidth}
\centering
\text{\scriptsize $0.256$}
\endminipage
\hfill
\minipage{0.09\textwidth}
\centering
\text{\scriptsize $0.634$}
\endminipage
\hfill
\minipage{0.09\textwidth}
\centering
\text{\scriptsize $0.111$}
\endminipage
\hfill
\minipage{0.09\textwidth}
\centering
\text{\scriptsize $0.562$}
\endminipage
\hfill
\minipage{0.09\textwidth}
\centering
\text{\scriptsize $0.705$}
\endminipage
\hfill
\minipage{0.09\textwidth}
\centering
\text{\scriptsize $0.141$}
\endminipage
\hfill
\minipage{0.09\textwidth}
\centering
\text{\scriptsize $0.817$}
\endminipage
\hfill
\minipage{0.09\textwidth}
\centering
\text{\scriptsize $0.817$}
\endminipage
\hfill
\minipage{0.09\textwidth}
\centering
\centering
\text{\scriptsize $0.448$}
\endminipage
\endminipage   \\ \vspace{-3mm} 

\minipage{1\textwidth}
        \includegraphics[width=0.095\textwidth]{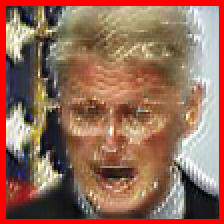}
        \includegraphics[width=0.095\textwidth]{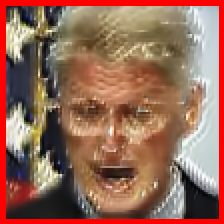}
        \includegraphics[width=0.095\textwidth]{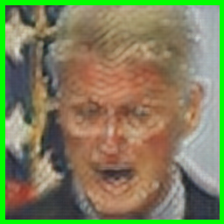}
       \includegraphics[width=0.095\textwidth]{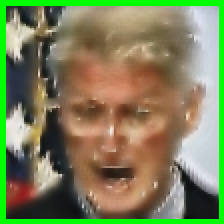}
        \includegraphics[width=0.095\textwidth]{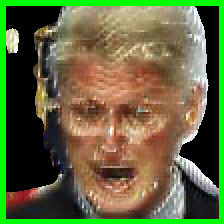}
        \includegraphics[width=0.095\textwidth]{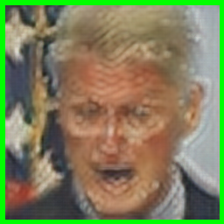}
        \includegraphics[width=0.095\textwidth]{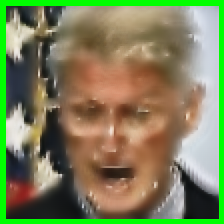}
        \includegraphics[width=0.095\textwidth]{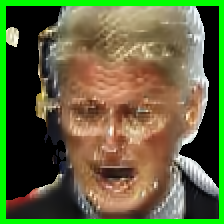}
        \includegraphics[width=0.095\textwidth]{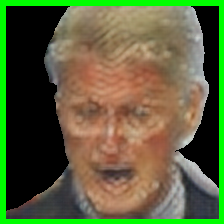}
       \includegraphics[width=0.095\textwidth]{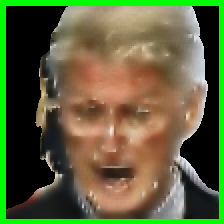}
\endminipage\\ \vspace{-2mm} 

\minipage{1\textwidth}
\minipage{0.09\textwidth}
\centering
\text{\scriptsize $0.467$}
\endminipage
\hfill
\minipage{0.09\textwidth}
\centering
\text{\scriptsize $0.452$}
\endminipage
\hfill
\minipage{0.09\textwidth}
\centering
\text{\scriptsize $0.695$}
\endminipage
\hfill
\minipage{0.09\textwidth}
\centering
\text{\scriptsize $0.576$}
\endminipage
\hfill
\minipage{0.09\textwidth}
\centering
\text{\scriptsize $0.932$}
\endminipage
\hfill
\minipage{0.09\textwidth}
\centering
\text{\scriptsize $0.682$}
\endminipage
\hfill
\minipage{0.09\textwidth}
\centering
\text{\scriptsize $0.567$}
\endminipage
\hfill
\minipage{0.09\textwidth}
\centering
\text{\scriptsize $0.936$}
\endminipage
\hfill
\minipage{0.09\textwidth}
\centering
\text{\scriptsize $0.921$}
\endminipage
\hfill
\minipage{0.09\textwidth}
\centering
\centering
\text{\scriptsize $0.875$}
\endminipage
\endminipage \\

 \vspace{-3mm} 

\minipage{1\textwidth}
        \includegraphics[width=0.095\textwidth]{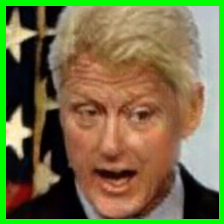}
        \includegraphics[width=0.095\textwidth]{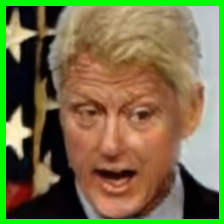}
        \includegraphics[width=0.095\textwidth]{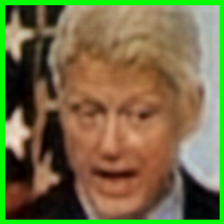}
        \includegraphics[width=0.095\textwidth]{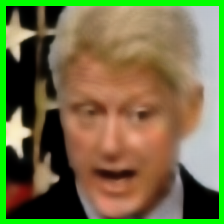}
        \includegraphics[width=0.095\textwidth]{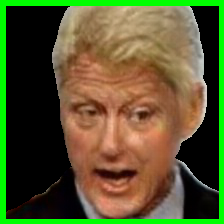}
        \includegraphics[width=0.095\textwidth]{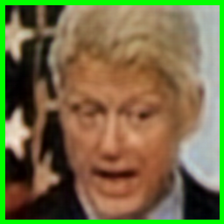}
        \includegraphics[width=0.095\textwidth]{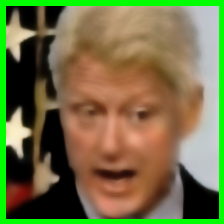}
        \includegraphics[width=0.095\textwidth]{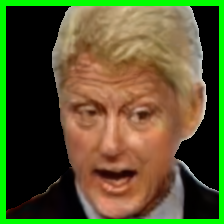}
        \includegraphics[width=0.095\textwidth]{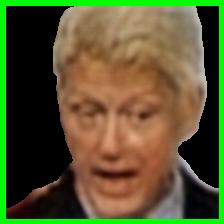}
        \includegraphics[width=0.095\textwidth]{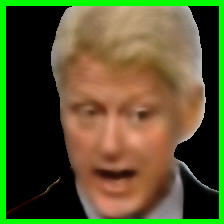}
\endminipage\\  \vspace{-2mm} 

\minipage{1\textwidth}
\minipage{0.09\textwidth}
\centering
\text{\scriptsize $0.661$}
\endminipage
\hfill
\minipage{0.09\textwidth}
\centering
\text{\scriptsize $0.630$}
\endminipage
\hfill
\minipage{0.09\textwidth}
\centering
\text{\scriptsize $0.696$}
\endminipage
\hfill
\minipage{0.09\textwidth}
\centering
\text{\scriptsize $1.000$}
\endminipage
\hfill
\minipage{0.09\textwidth}
\centering
\text{\scriptsize $0.638$}
\endminipage
\hfill
\minipage{0.09\textwidth}
\centering
\text{\scriptsize $0.685$}
\endminipage
\hfill
\minipage{0.09\textwidth}
\centering
\text{\scriptsize $0.795$}
\endminipage
\hfill
\minipage{0.09\textwidth}
\centering
\text{\scriptsize $0.742$}
\endminipage
\hfill
\minipage{0.09\textwidth}
\centering
\text{\scriptsize $0.963$}
\endminipage
\hfill
\minipage{0.09\textwidth}
\centering
\centering
\text{\scriptsize $0.769$}
\endminipage
\endminipage 
\endminipage
\hspace{1mm}
\minipage{0.015\textwidth}
\begin{turn}{270}
  \scriptsize{ \hspace{5mm} $k$--AAP \hspace{7mm} FGSM \hspace{5mm} FlowSAN--3 \hspace{3mm} FlowSAN--5
  \hspace{2.5mm} PrivacyNet\hspace{3mm}}
  \end{turn}
\endminipage
\caption{Examples of \frameworkname~reconstructions, where the goal is to recover gender information from privacy-enhanced images. The number below each image corresponds to the probability output of a gender classifier  -- probabilities between $0$--$0.5$ correspond to female and $0.5$--$1$ to male subjects. Note that the privacy enhanced images, $I_{pr}$, generate incorrect gender probabilities (indicated by the red frames) or significantly reduce the correct-gender prediction probability. PrivacyProber, on the other hand,  recovers the correct gender information in most cases -- indicated by the green frames. Depending on version of PrivacyProber images of different visual quality are generated.\label{fig:examples}\vspace{-0mm}}
\end{figure*}